\documentclass{article} 
\usepackage[final]{colm2026_conference}

\usepackage{microtype}
\usepackage{wrapfig}
\usepackage{hyperref}
\usepackage{url}
\usepackage{booktabs}
\usepackage{longtable}
\usepackage{graphicx}
\usepackage{subcaption}
\usepackage{float}
\usepackage{array}
\usepackage{ragged2e}
\usepackage{xltabular}
\usepackage[utf8]{inputenc}

\usepackage{lineno}

\definecolor{darkblue}{rgb}{0, 0, 0.5}
\hypersetup{colorlinks=true, citecolor=darkblue, linkcolor=darkblue, urlcolor=darkblue}

\usepackage{amsmath,amsfonts,bm}

\def\eqref#1{equation~\ref{#1}}

\def\1{\bm{1}}

\DeclareMathAlphabet{\mathsfit}{\encodingdefault}{\sfdefault}{m}{sl}
\SetMathAlphabet{\mathsfit}{bold}{\encodingdefault}{\sfdefault}{bx}{n}

\title{Uncovering the Computational Ingredients of Human-Like Representations in Large Language 
Models}

\author{
Zach Studdiford$^{1}$ \quad Timothy T. Rogers$^{1}$ \quad Kushin Mukherjee$^{*2}$ \quad Siddharth Suresh$^{*1}$ \\[0.5ex]
\textnormal{$^{1}$University of Wisconsin--Madison \quad $^{2}$Stanford University} \\[0.5ex]
\texttt{\{zstuddiford, ttrogers, siddharth.suresh\}@wisc.edu} \\
\texttt{kushinm@cs.stanford.edu}
}

\begin{document}

\ifcolmsubmission
\linenumbers
\fi

\maketitle

\begingroup
\renewcommand\thefootnote{}\footnote{$^*$Equal contribution. }%
\addtocounter{footnote}{-1} 
\endgroup 

\begin{abstract}

The human ability to translate diverse perceptual and linguistic inputs into structured behavior has been thought to rest on learning robust representations of concepts. The rapid advancement of transformer-based large language models (LLMs) has surfaced a diversity of computational ingredients relevant for model building — architectures, fine-tuning methods, and training datasets among others — yet it remains unclear which are most crucial for developing human-like conceptual representations. Further, most current benchmarks are ill-suited to measuring \textit{representational alignment}, making LLMs' scores on them unreliable for assessing whether they are progressing as cognitive models. We address these limitations by evaluating over 75 models on a triplet similarity task, a method well established in cognitive science for measuring conceptual representations, using concepts from the THINGS database. We find that instruction fine-tuning and larger attention head dimensionality are among the strongest predictors of human alignment, while activation function choice, multimodal pretraining, and parameter size have limited influence on alignment. 
Correlations between alignment scores and existing benchmark scores reveal that while some benchmarks (e.g., BigBenchHard) better capture representational alignment than others (e.g., MUSR), none fully accounts for the variance in human-model alignment, demonstrating their insufficiency. Taken together, our findings highlight key computational ingredients for advancing LLMs as models of human conceptual representation and address a key gap in LLM evaluation.

\end{abstract}
\section{Introduction}

The success of deep neural network models in domains ranging from perception \citep{yamins2014performance, yamins2016using}, to robotic manipulation \citep{finn2016unsupervised}, to language understanding \citep{tuckute2024language} has partly been attributed to their ability to learn powerful \textit{representations} that support translating diverse inputs into appropriate outputs.
The continued development of transformer-based large language models (LLMs) and their capabilities on naturalistic human tasks might imply that, through training on sufficiently large corpora and with the correct architectural ingredients, these models come to possess representations largely isomorphic to human mental representations. 
This implication is critical for cognitive science, which has long sought to characterize human conceptual representations and the operations performed over them that give rise to naturalistic behavior. Indeed, there is a rich tradition of using deep neural networks as \textit{computational cognitive models} — treating their learned representations as proxies for human representations, and mapping their evolution to the development of human conceptual knowledge \citep{jackson2022late, warstadt2025findings, rogers2004semantic, cichy2019deep, saxe2019mathematical, vong2024grounded}. Yet, while evidence across domains has revealed concordances between LLM representations and both human behavior and neural activity, it remains unclear which design decisions relating to building LLMs best predict strong human-model alignment — that is, which \textit{computational ingredients} (architecture, scale, instruction-tuning, activation functions, etc.) give rise to human-aligned conceptual representations across a large and diverse set of concepts.

Identifying these computational ingredients is critical for several reasons.
While frontier models have achieved remarkable performance on diverse benchmarks spanning scientific reasoning \citep{rein2024gpqa,suzgun2022challenging}, software engineering \citep{chen2021evaluating, jimenez2023swe}, and even graphics and humor understanding \citep{masry2022chartqa, methani2020plotqa, mukherjee2025encqa, zhou2025bridging, hessel2022androids} -- domains thought to require `human-like' thinking -- we lack thorough accounts explaining what properties of models account for these successes.
Recent work has shown that optimizing for benchmark accuracy alone is insufficient for building models that fail in human-like ways and that are generally aligned with humans \citep{fel2022harmonizing, ying2025benchmarking}. 
We argue that prioritizing representational alignment \citep{sucholutsky2023representational, sucholutsky2023alignment, muttenthaler2024aligning} offers a promising path forward. Measuring the similarity between model and human internal representations can provide a more unified account of model capabilities, guide the development of more robustly aligned systems \citep{collins2024building, muttenthaler2024aligning, peterson2018evaluating}, and accelerate cognitive science by yielding more faithful computational models of the human mind.



\vspace{-1em}
\section{Related work}


\paragraph{Quantifying Human and Model Representations using Large-Scale Concept Datasets}
Mapping the geometry of human conceptual knowledge has long been a central goal of cognitive science \citep{rogers2004semantic, shepard1980multidimensional, tversky1977features, rumelhart1986parallel}. Early attempts relied on explicit feature ratings and pairwise similarity judgments, which scaled poorly and thus limited their utility for capturing the vast inventory of human concepts \citep{rosch1975cognitive, nosofsky1984choice, shepard1980multidimensional}. Structured probabilistic models, while useful for explaining aspects of concept learning, have similarly been constrained by their reliance on specific relational structures, limiting scope and generalizability \citep{zhao2024model, wong2022identifying, kemp2008discovery}. Recent years have seen a resurgence of similarity-based approaches, driven by scalable algorithms that convert human judgments into expressive high-dimensional \textit{semantic embeddings} capable of capturing the latent dimensions organizing human semantic knowledge \citep{king2019similarity, mukherjee2025using, hebart2019things, suresh-etal-2023-conceptual, peterson2018evaluating, kriegeskorte2012inverse, cichy2019spatiotemporal, hebart2023things, muttenthaler2022vice, jamieson2015next, sievert2023efficiently}. Among these, triadic similarity judgment tasks — where participants select the odd item from a triplet — are a particularly powerful representational learning paradigm \citep{tamuz2011adaptively, wah2014similarity, kleindessner2014uniqueness}, with derived embeddings proving reliable predictors of both neural and behavioral patterns across a variety of semantic tasks \citep{mukherjee2025using, mukherjee2025drawings, hebart2023things, colon2023yours, suresh-etal-2023-conceptual, marjieh2022predicting, mukherjee2022color, mur2013human, suresh2024categories}.
 

A complementary advance that has recently galvanized research in representational alignment is the development of concept datasets that aim to capture the diversity of object concepts humans reason about \citep{mehrer2021ecologically, hebart2019things, giallanza2024integrated, suresh2025aienhancedsemanticfeaturenorms}.
Datasets like THINGS \citep{hebart2019things} and ECOSET \citep{mehrer2021ecologically} consist of concept sets and associated images along with rich psychological and neural metadata \citep{hebart2023things} that allow for comparisons between human and model representations along multiple levels of analysis --- behavioral, neural, and computational.


\vspace{-1 em}
\paragraph{Measurement of Representational Alignment}

Several complementary measures for assessing representational alignment have been developed over recent years \citep{kriegeskorte2008representational, sucholutsky2023representational,barbosa2025quantifying, oswal2016representational}. 
Core measurement techniques include Representational Similarity Analysis (RSA) using correlation between similarity matrices derived from behavioral, neural, or neural network activation data \citep{kriegeskorte2008representational, nili2014toolbox}, Procrustes analysis for finding optimal linear alignments \citep{schonemann1966generalized}, and Centered Kernel Alignment (CKA) for comparing representations invariant to linear transformations \citep{kornblith2019similarity, williams2021generalized}. 
Recent work has further developed more sensitive metrics including shape-based metrics \citep{williams2021generalized}, topological measures \citep{barannikov2021representation}, information-theoretic approaches \citep{bansal2021revisiting}, and regularized regression-based methods \citep{oswal2016representational} each of which iteratively address issues relating to model regularization, generalization, and error-bounds.
Measuring alignment is useful because models with higher human alignment show improved few-shot learning \citep{sucholutsky2023alignmenthumanrepresentationssupports, huh2024platonic}, better out-of-distribution generalization \citep{moschella2023relative, norelli2023asif}, enhanced adversarial robustness \citep{engstrom2019adversarial}, and more human-like error patterns \citep{geirhos2021partial, fel2022harmonizing}. These findings suggest that human-aligned models capture meaningful invariances that support robust intelligence.

\vspace{-1em}
\paragraph{Computational Ingredients Driving Alignment}

Understanding which design choices yield human-aligned representations in foundation models can help uncover general principles undergirding human intelligence. In vision, work has shown that optimizing models for categorization \citep{yamins2014performance} or contrastive objectives \citep{konkle2022self, zhuang2021unsupervised} produces representations closely aligned with human visual cortex, offering insight into the objective functions the brain may solve to construct visual representations. Extending this line of inquiry to over 75 models, \citet{muttenthaler2022human} found that training data and objectives strongly predicted human-model alignment, whereas architecture and scale had minimal impact — a finding that challenges scaling law accounts \citep{kaplan2020scaling, cherti2023reproducible} predicting that larger models should broadly perform better on most tasks.
Other key results regarding the importance of different computational ingredients include the finding that self-supervised contrastive methods align more closely with human vision than non-contrastive approaches \citep{chen2020simple, zbontar2021barlow, caron2020unsupervised}; multimodal image–text contrastive training improves alignment over vision-only training \citep{radford2021learning, jia2021scaling, pham2022combined}; training on diverse, naturalistic datasets such as \texttt{JFT-3B} improves alignment beyond ImageNet \citep{zhai2022scaling, dehghani2023scaling}; enforcing shape bias on models enhances alignment while texture bias impairs it \citep{geirhos2019imagenet, hermann2020origins}; and intermediate layers often show stronger perceptual alignment than final layers \citep{berardino2017eigen, kumar2022better}.
Similar approaches, dissecting the predictive power of different model building choices, have not been brought to bear in investigations into conceptual structure in LLMs- leaving unclear what the relative importance of different computational ingredients are when building foundation models of human semantic knowledge.


While early studies showed that contextualized embeddings from models like BERT and GPT-2 could predict human similarity judgments \citep{bommasani2020interpreting, grand2022semantic}, more recent work demonstrates that LLMs can achieve remarkably high alignment with human conceptual structure \citep{marjieh2024large, marjieh2024predicting, mukherjee2024largelanguagemodelsestimate, suresh2025aienhancedsemanticfeaturenorms, suresh2023semanticfeatureverificationflant5, mukherjee2023humanmachinecooperationsemanticfeature}, with instruction-tuned models showing particular advantages \citep{tong2024mass, gurnee2024language}.
Several studies have used triplet tasks specifically to probe LLM representations \citep{nam2025idealign, studdiford2025evaluating, rathi2024can, suresh-etal-2023-conceptual}, finding that larger models and those trained on more diverse data show better alignment. However, systematic analyses disentangling the effects of scale, architecture, training data, and objectives remain limited. Work on multimodal models suggests that vision-language training improves conceptual alignment beyond either modality alone \citep{yuksekgonul2023when, thrush2022winoground, qin2025vision}, though the precise mechanisms remain unclear.

\begin{figure}[t!]
    \centering
    \includegraphics[width=.8\linewidth]{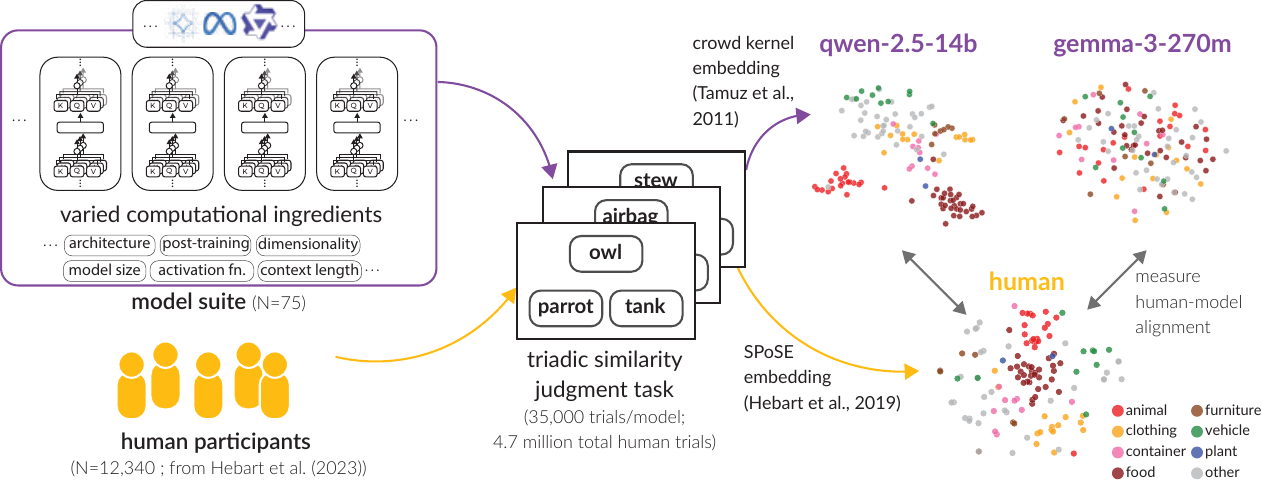}
    \caption{\textit{Method for estimating human-model alignment.} We collected 35k triplet similarity judgments from each of the 77 models in our suite. We computed semantic embeddings based on these judgments and compared the representational geometry of model embeddings to human embeddings derived from \citet{hebart2023things}.}
    \vspace{-1em}
    \label{fig:methods}
\end{figure}

\vspace{-1em}
\section{Methods}

\textbf{Summary of key contributions.} We leverage the open-weight model ecosystem to construct a suite of 75+ models spanning diverse computational ingredients. Using concepts and semantic embeddings from THINGS dataset \citep{hebart2019things} in conjunction with a triadic similarity judgment paradigm \citep{jamieson2015next, hebart2023things, sievert2023efficiently}, we obtain both human and model-specific conceptual embeddings using analogous methods to ensure model-fair comparisons \citep{firestone2020performance}. 
We then compare these embeddings to human data and apply statistical models to identify the ingredients most predictive of human-model alignment. 
Lastly, we relate alignment to performance on standard LLM benchmarks, highlighting points of divergence between representational alignment and task capabilities.

\paragraph{Concept Dataset}
To investigate how various LLM characteristics influence human alignment, we carefully curated a set of test concepts that would provide robust and interpretable results. 
Starting from the full set of 1,854 THINGS object concepts \citep{hebart2019things, hebart2023things}, we subsampled 128 concrete, real-world object concepts (e.g., ``lion'', ``banjo'', ``car'')  that spanned four main cognitive axes of variation: familiarity, artificiality, animacy, and size based on prior work \citep{mukherjee2023seva} \footnote{We empirically validated the consistency of this subsample with the full 1,854 concept set, finding negligible differences in the human-alignment of large and small models between embeddings computed from full and subsamples, see Appendix \ref{app:subsample_justification}.}.
This sampling strategy ensured that we had adequate semantic diversity in our concept dataset (representative of the full THINGS database) while also ensuring that we could generate semantic embeddings for a large suite of models in a tractable manner.
See Appendix~\ref{app:concepts} for the complete list of concepts with detailed descriptions of concept selection criteria and validation procedures.

\paragraph{Model Suite}

We evaluate a range of open-source models varying across several key computational ingredients: model architecture, primary activation functions used, pre-training modality, whether models were instruction fine-tuned or not,  scale (both in terms of model parameters and training tokens), context length, dimensionality of attention heads/MLPs/residual streams.
While in principle it is possible to offer even more granular analyses of computational ingredients, these factors constitute major architectural and training decisions when building foundation models, which we believe provide adequate resolution to investigate the drivers of human-model alignment.
We evaluated a set of 77 open-weight transformer models including models from popular families (Llama, Qwen, Gemma, etc.). 
The full set of models can be seen in Table \ref{tab:model_rankings_new}. Notably, we included models that varied in size from 100M parameters to 42B parameters, ranged in the degree of post-training (from base to instruction fine-tuned), and architecture-style (decoder-only, encoder-decoder, and state space models (SSMs) \citep{gu2022efficientlymodelinglongsequences}.

\paragraph{Triadic similarity judgment task}
The triadic similarity judgment task can be framed in two equivalent ways: (1) \textbf{anchored similarity judgments} where participants are presented with a triplet of items (images or words depicting different concepts) and select which of two option items is most similar to an anchored reference item, or \textbf{odd-one-out judgments} where they identify which of the three presented item is the odd-one-out. 
Both framings have been successfully employed to study conceptual representations in humans and models \citep{hebart2019things, hebart2023things, mukherjee2025using, colon2023yours, suresh-etal-2023-conceptual, studdiford2025evaluating, mukherjee2025drawings, muttenthaler2022vice}. 
In this work, we leverage existing human similarity ratings data \citep{hebart2023things} and corresponding embeddings, which were collected using the odd-one-out paradigm and embedded using the sparse positive similarity embedding (SPoSE) method, respectively. 
For model evaluations, we adopted the anchored similarity judgment paradigmn for computational efficiency, deriving embeddings using ordinal embedding techniques \citep{jamieson2015next, tamuz2011adaptively} following recent work \citep{suresh-etal-2023-conceptual, mukherjee2025drawings, colon2023yours}. 
Crucially, the ordinal embedding approach provides theoretical bounds on sample complexity \citep{jamieson2015next}, allowing us to determine the number of triplet comparisons needed for faithful representations a priori. While these approaches differ in framing and embedding computation, they yield comparable representational structures, as we demonstrate empirically in Appendix~\ref{app:rep-compare}. 
This methodological choice allows us to maximize the use of existing high-quality, validated human data while efficiently collecting model responses with known sampling requirements in a scalable manner, ensuring robust human-model comparisons across a large suite of models.

\label{subsec:embedding_estimation}

\paragraph{Human Semantic Embeddings}

Human semantic embeddings were obtained by \citet{hebart2023things} from the behavioral judgments of $N=12,340$ online participants in a triplet odd-one-out task. On each trial, participants were shown three items from the THINGS database and were prompted to select the most dissimilar object (\textit{"Which is the odd one out?"}). In total, 4.7 million behavioral triplet judgments were obtained for the full set of 1,854 concepts. 
To estimate an embedding space from the set of individual triplet judgments, \citet{hebart2023things} used the SPoSE algorithm \citep{hebart2023things}, which learns low-dimensional vectors for each object such that (1) pairs of objects judged as similar are placed closer in the embedding space, and (2) dimensions are constrained to be sparse in order to yield human-interpretable properties. The resulting space captured meaningful representational structure from the individual judgments of participants, revealing core dimensions along which human semantic representations are organized.

{\paragraph{Model Semantic Embeddings}

While much work on representational similarity extracts activation vectors from models and computes their similarity to human behavior-based representations via methods like RSA \citep{kriegeskorte2008representational}, we instead estimated model semantic embeddings from similarity judgments using a triadic comparison task akin to those used to estimate human embeddings. We did so for two principled reasons. First, although computations in transformer layers and attention heads have been linked to human language processing \citep{tuckute2024language}, no general framework exists for identifying which parts of a model carry signatures of human semantic knowledge across model families, making fair comparison of activations to human representations difficult. Second, the ability of modern LLMs to perform structured tasks via natural language prompting — combined with the relative simplicity of the triadic judgment task — means that administering the same test to both humans and models constitutes a species-fair \citep{firestone2020performance} comparison, ensuring that alignment discrepancies reflect genuine representational differences rather than artifacts of differing embedding methods.

For each model in our suite, we collected 35k triadic judgment responses.
Using the human semantic embeddings, we estimated that a rank 30 space explains over 95\% of the variance in embeddings (see \ref{app:dim-select}). 
Based on known bounds \citep{sievert2023efficiently}, we needed ($n\cdot d \cdot log_2(n)$\footnote{$n$ is the number of items and $d$ is the expected dimensionality of the space}) at least 26,900 judgments on random sets of triplets to estimate a robust embedding. 
Each triplet trial was randomly sampled from the ${128 \choose 3}$ possible triplet trials. We evaluated each model using its default huggingface sampling settings using a common system-level prompt --- \texttt{"You are a helpful assistant who gives responses to questions"}.
We used the following prompt for each trial for each model ---
\texttt{QUESTION: Which item is most similar to {item\_x}: {item\_y} or {item\_z}? Respond only with the item name.} If we were evaluating a base model (not instruction-tuned), we appended \texttt{Answer:} after the initial prompt. Instruction template formats were applied where appropriate and
each prompt was evaluated independently.

Using these similarity judgments, we fit an ordinal embedding algorithm \citep{tamuz2011adaptively, sievert2023efficiently} that organized the semantic embedding space such that items that were judged to be similar more often were pulled closer together in this space. 
We fit 30D embeddings based on the dimensionality of human semantic embeddings.
We also held out 20\% of the data during the embedding fitting process and evaluated the quality of the embeddings by monitoring the crowd-kernel loss \citep{tamuz2011adaptively} to ensure the embeddings were of high quality.


\vspace{-1em}
\section{Results and Discussion}
We first report how well different computational ingredients predicted human-model alignment, initially considering each ingredient separately via individual correlations or $t$-tests \citep{kim2015t}, then assessing relative contributions of each via mixed-effect regression models \citep{lindstrom1988newton}. Finally, we evaluate the relationship between human-model alignment and model performance on other key LLM benchmarks \citep{sprague2023musr, rein2024gpqa, srivastava2023imitationgamequantifyingextrapolating, zhou2024ifeval, hendrycks2020measuring}.
Our primary alignment measure was the variance in human semantic embeddings explained by model embeddings ($R^2$) after Procrustes alignment \citep{gower1975generalized}. Procrustes transforms find the optimal linear transformation (rotation, scaling, translation) between two vector spaces of equal dimensionality to minimize squared distances between corresponding points. Using human embedding variance (human SS) as the baseline, we compute $R^2 = 1 - \tfrac{\text{residual SSE}}{\text{human SSE}}$. Other alignment measures such as CKA \citep{kornblith2019similarity} and RSA\citep{kriegeskorte2008representational} were highly correlated with Procrustes $R^2$ (Appendix \ref{app:measure_corrs}). We also replicated the full suite of analyses reported below on a set of 213 abstract emotion words, details of which can be seen in Appendix Section \ref{app:emotions-replicate}.
\subsection{Contributions of Model Computational Ingredients to Alignment}

\textbf{Instruction tuning leads to greater alignment.}
Models that were instruction fine-tuned were significantly better aligned than those that were not ($t= 5.11, p < 0.001$; see Figure \ref{fig:t-scores}A).
Qualitatively, instruction fine-tuning clustered representations of semantic categories, such that within-category items were closer in representation space and between-category items were more distal (see Appendix \ref{fig:tsne_models_part1}). 
Thus, the same post-training techniques that enable models to be more adept at following prompt instructions also bring their representations closer into alignment with humans.

\begin{figure}[t!]
    \centering
    \includegraphics[width=.85\linewidth]{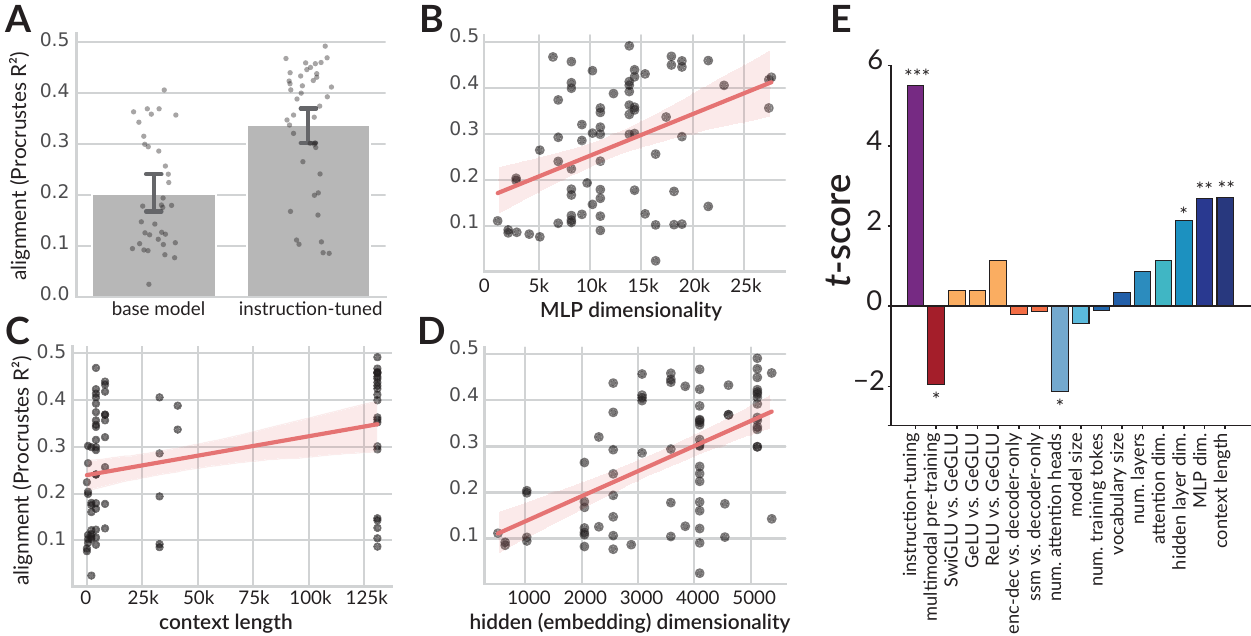}
    \caption{\textbf{Computational ingredients predictive of human model alignment}. \textbf{A.}, \textbf{B.}, \textbf{C.}, and \textbf{D.} show the relationship between instruction-tuning, MLP dimensionality, context length, and embedding dimensionality on Procrustes $R^2$. These were the four ingredients most predictive in the mixed-effects regression model. \textbf{E.} $t$-scores for each predictor, which highlight the relative contribution of each ingredient towards alignment when considered in a single statistical model.}
    \label{fig:t-scores}
    \vspace{-1em}
\end{figure}

\textbf{Model architecture dimensionality corresponds to greater alignment.}
We found a strong positive relationship between the dimensionality of individual model architecture components --- MLPs and attention heads --- and the degree of model-human alignment. Specifically, the number of per-layer attention heads ($r = 0.52, p < 0.001$), attention matrix dimensionality ($r = 0.30, p = 0.026$), embeddings dimensionality ($r = 0.54, p < 0.001$), and MLP dimensionality ($r = 0.40, p < 0.001$) each correlated positively with human representational alignment. 
Figure \ref{fig:t-scores}B and D show the individual correlations between MLP and embedding dimensionality and model-human Procrustes $R^2$ (the attention dimensionality was not significant in multiple regression models reported in subsequent sections). 
Thus, the greater expressivity granted by larger latent activation spaces and more attention heads predicts greater human alignment.

\textbf{Multi-modal pretraining has no effect on alignment.}
While much of human concept learning is inherently multimodal \citep{rogers2004semantic, patterson2007you}, there is limited work showing whether vision-language pretraining (the most prominent class of multimodal training) leads to more aligned semantic representations (cf. \cite{qin2025vision}).
Here, multimodal (image) pre-training had no independent effect on human-model alignment relative to text-only models ($t=0.34, p>0.05$; Appendix Figure \ref{app:architecture_rels}).

\textbf{Larger model sizes show greater human alignment.}
For visual representations, \citet{muttenthaler2022human} found that larger models need not yield more human-like representations. Neural scaling laws \citep{kaplan2020scaling}, however, might predict that larger and deeper models should be more performant, and perhaps more aligned. In accordance with this prediction,
human-model alignment scaled approximately linearly with both model parameter count  and the number of model layers ($r_{param}=0.45, r_{layers} =0.41, p<0.001$; Appendix Figure \ref{app:architecture_rels}).
Thus, larger scale predicts greater human-model alignment.

\textbf{Alignment increased with amount of training data.}
Models exposed to more tokens in pretraining generally exhibited more human-like representations ($r=0.33$, $p<0.01$; Appendix Figure \ref{app:architecture_rels}). 
Although model training data also likely varied in quality and composition--thus limiting strong conclusions about which properties matter most--our results indicate that the \textit{amount} of training data can predict alignment.

\textbf{Alignment scales with context length.}
The triadic similarity task does not require a long context length, but we nevertheless found that models with a greater maximum context-length also demonstrated greater human alignment ($r=0.35, p<0.01$; Figure \ref{fig:t-scores}). 
This finding could be linked to post-training regimes that unlock long-context reasoning, which consequently lead models to perform well on short-context tasks as well.

\textbf{Choice of activation function does not affect alignment.}
Activation functions have seen iterative development over the years with recent variants (e.g., SwiGLU \citep{shazeer2020glu}) being particularly well-suited to managing gradient flows during training and finetuning experiments leading to faster and stabler training regimes. But do choices in activation functions ground out in alignment?
We found no advantage for any particular function in increasing human alignment ($F(3, 64) = 1.36, p = .26$; Appendix Figure \ref{app:architecture_rels}).

\textbf{Larger vocabulary size does not increase alignment.}
One might expect that larger model vocabularies unlock greater expressivity for models, leading them to be more human-aligned. Against this prediction, we found no relationship between model vocabulary size and human-alignment ($r=0.10$; Appendix Figure \ref{app:architecture_rels}).


\textbf{Which computational ingredients have the greatest relative contribution towards human-model alignment?}
Thus far, we have investigated how different computational ingredients impact human-model alignment in isolation. 
In reality, many of these ingredients vary across models at the same time, leaving open the questions of (1) which ingredients are most important \textit{relative to others} and (2) which ingredients account for unique variance after others are considered.

To answer these questions, we fit a mixed-effects multiple regression model \citep{lindstrom1990nonlinear} predicting Procrustes $R^2$ from \textit{all} computational ingredients. Mixed-effects models combine the interpretability of OLS with random effects to capture item-level variance \citep{barr2013random}\footnote{We additionally check the collinearity of model predictors and find low to intermediate levels of collinearity across all predictors, see Appendix \ref{app:multicollinearity_tests}}.} 
We included random intercepts for model family (e.g., Llama, Qwen) to capture variance related to these families not grounded out in our set of ingredients, and fixed effects for all ingredients. The fitted model explained substantial variance in alignment ($R^2 = 0.78$; Fig. \ref{fig:t-scores}E). 
Instruction fine-tuning emerged as the strongest predictor ($\beta = 0.13$, $SE$ = 0.024, $p < 0.001$). Dimensionality measures --- MLP ($\beta = 0.049$, $SE$ = 0.018, $p < 0.01$), embedding/hidden layers ($\beta = 0.048$, $SE$ = 0.022, $p < 0.05$) and context length ($\beta = 0.042$, $SE$ = 0.015, $p < 0.01$) ---  also accounted for significant unique variance. In contrast to the independent analyses, attention dimensionality was not significant, and more attention heads predicted worse alignment after accounting for other factors ($\beta = -0.045$, $p < 0.05$).
Likewise, multimodal pretraining, which showed no independent effect on alignment, predicted reliably lower alignment after accounting for other factors ($\beta =-0.102$,$SE =0.052$, $p < 0.05$). Thus, current multimodal training regimes may actually limit rather than improve human-model alignment. No other ingredient explained significant unique variance when others were included in the model (see effect sizes in Figure \ref{fig:t-scores}E). 



\textbf{Predictors of human alignment are robust across different concept sets.} A potential concern is the possibility that our results are sensitive to experimenter decisions regarding the specific concepts tested. We addressed the possibility that these results are specific to the \texttt{THINGS} items evaluated, repeating the analyses above on a set of abstract \texttt{EMOTIONS} concepts \citep{shaver1987emotion} for our full model suite. 
We observe similar findings across both stimulus sets: individual correlations between
alignment and MLP dimensionality ($r=0.42$, $p<0.001$), context length ($r=0.31$,
$p<0.01$), and hidden layer dimensionality ($r=0.56$, $p<0.001$) are replicated in our
\texttt{EMOTIONS} stimuli, while the relative contributions of instruction tuning
(objects: $\beta=0.135$, $t=5.73$, $p<0.001$; emotions: $\beta=0.235$, $t=5.09$,
$p<0.001$) and number of attention heads (objects: $\beta=-0.048$, $t=-2.27$, $p<0.05$;
emotions: $\beta=-0.178$, $t=-2.76$, $p<0.01$) are reflected in both of the
mixed-effects models. Full results for the \texttt{EMOTIONS} stimuli are included in Appendix \ref{app:emotions-replicate}. We also find that human-model alignment is consistent when computed over both the subsample and complete set of \texttt{THINGS} concepts (Appendix \ref{app:subsample_justification}) and that randomly sampling different sets of LLMs produces results consistent with our main findings (Appendix \ref{app:model_subsample_pred_stability}). In sum, our findings show a fair degree of consistency across different kinds of concepts.

\textbf{Which concept dimensions drive human alignment?} While our analyses capture human-model alignment across the subset of THINGS concepts, it remains unclear whether certain kinds of concepts are more aligned than others. That is, do certain concept attributes predict more or less item-level human alignment? Leveraging attribute metadata included for each concept in THINGS, we first grouped each of the 128 concepts based on superordinate category labels (e.g \textit{furniture, clothing, food}) and computed the Procrustes $R^2$ within each category group. We found that alignment was significantly higher for inanimate objects (furniture, container) relative to animate objects (animals, food) (Appendix Figure \ref{fig:model_size_align}), and that these category-level differences were relatively consistent across models (Kendall's $\tau = 0.45$, $ps<$.05). Next, we asked whether certain item-level attributes (e.g familiarity, concreteness) predicted item-level differences in alignment (measured as the cosine distance between the model embedding and the ground-truth human embedding). 
We fit a mixed-effects regression predicting these item-level divergences from nine different attributes, and found that items judged as more concrete or typical of their category tended to be more aligned, while more polysemous concepts (words with multiple meanings) were less aligned (see Appendix Figure \ref{fig:dimension_alignment}).

\begin{wrapfigure}{r}{.45\textwidth}
    \centering
    \includegraphics[width=0.45\textwidth]{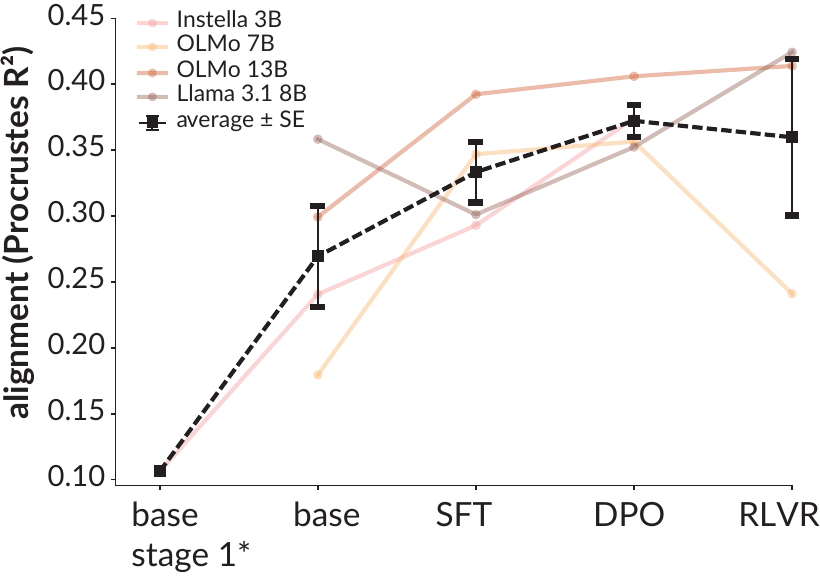}
    \caption{\textbf{Effects of post-training on alignment}. The black points show mean alignment (across models) at each post-training stage.*Instella only.}
    \label{fig:journey}
    \vspace{-1 em}
\end{wrapfigure}

\textbf{Effect of different post-training regimens on human alignment.} A small number of open-source model families expose checkpoints across various stages of pre-training, allowing for a direct comparison of different post-training methods on human-alignment. Figure \ref{fig:journey} shows changes in human alignment when SFT, DPO, and RLVR are applied in \texttt{OLMo}, \texttt{Instella}, and \texttt{Llama} models. While we caveat that the limited number of available checkpoints prevents a more systematic analysis, we qualitatively observe that SFT and DPO post-training stages led to improvements in alignment whereas the effect of RLVR was more muted.  


\subsection{Relationship Between Alignment and Established Benchmarks}



\begin{figure}
\centering
 \includegraphics[width=.9\textwidth]{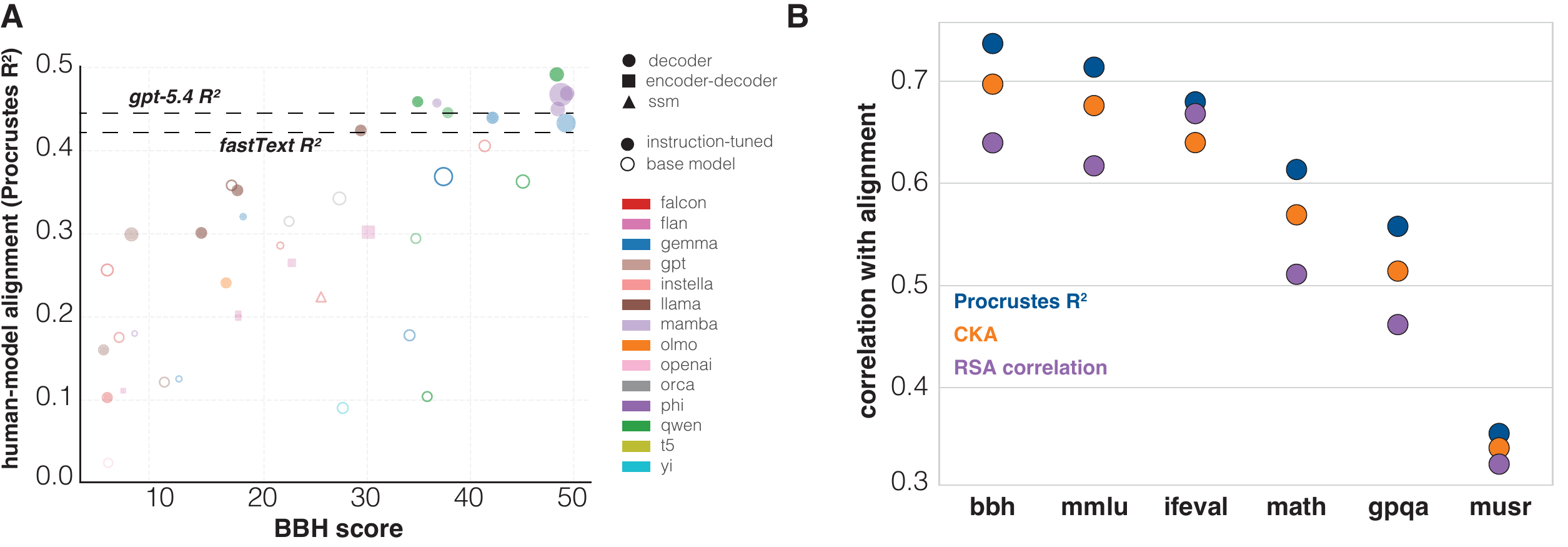}
  \caption{\textbf{Relationship between alignment and other LLM benchmarks}. \textbf{A.} \texttt{BigBenchHard} scores for each model in our suite vs. their Procrustes $R^2$ value w.r.t. human embeddings. \texttt{gpt-5.4} and \texttt{fastText} $R^2$ values indicated by dotted lines. \textbf{B.} Correlation between alignment and the six LLM benchmarks evaluated.}
  \label{fig:other-benchmarks}
\end{figure}

It is possible that representational alignment is already tracked by existing benchmarks. If so, it would provide a lens into what aspects of behavior most closely track representational alignment (e.g., grammaticality judgments). If not, then standard benchmarks overlook representational alignment as an axis of model competence.
To understand the relationship between alignment and performance on standard LLM benchmarks we estimated the correlation between alignment, measured using three metrics (Procrustes $R^2$, CKA, and RSA correlation), and model scores on five key benchmarks -- \texttt{MMLU \citep{hendrycks2020measuring}, IFEval \citep{zhou2023instruction}, BigBenchHard (BBH)\citep{suzgun2022challenging}, GPQA \citep{rein2023gpqa}} and \texttt{MUSR} \citep{sprague2024musr}. 
In addition to the open-source models included in our evaluation, we compute the Procrustes $R^2$ for \texttt{GPT-5.4} in order to obtain an approximate empirical upper-bound on human-model alignment in models with high benchmark scores.
We also included the Procrustes $R^2$ obtained from \texttt{FastText} embeddings of concepts to measure how well aligned concepts are between people and modern, static word embedding models. Despite a reported \texttt{BigBenchHard} score of $90.9$, we observe that the human-alignment of \texttt{GPT-5.4} is exceeded by open-source LLMs with significantly lower benchmark scores. Conversely, we observe notably high alignment in \texttt{FastText} embeddings despite this model being comparatively much smaller than any of the LLMs tested. Figure \ref{fig:other-benchmarks}A shows the relationship between alignment (Procrustes $R^2$) and scores on BBH, the benchmark most strongly correlated with alignment.
While models that performed well on BBH generally had higher alignment, we found some notable divergences especially for base models that scored lower on alignment than one might expect based on their BBH score.
Figure \ref{fig:other-benchmarks}B shows the correlations between alignment and all the established benchmarks. While all three alignment metrics were in agreement, we found that the overall correlations varied ($r_{min} = 0.36, p <0.05$; $r_{max} = 0.74, p<0.001$). We focus on Procrustes $R^2$ moving forward.

Performance on benchmarks that probe probe domain-general knowledge (BBH; $r=0.74, p<0.001$; MMLU; $r=0.71, p<0.001$) and instruction-following ability (IFEval; $r=0.68, p<0.001$) were more correlated with alignment than benchmarks testing more specific domain knowledge and reasoning ability (Math; $r=0.61, p<0.001$, MUSR;  $r=0.36, p<0.05$). 
This finding is sensible given that alignment with human semantic judgments requires representing semantic \textit{knowledge} broadly in human-like ways (captured by \texttt{BBH} and \texttt{MMLU}) and \textit{deploying} that knowledge in context-sensitive ways according to instructions (captured by \texttt{IFEval}).
We also found that the correlation between \texttt{BBH} and \texttt{IFEval} was significantly weaker than the correlation between these evaluations and our measure of representational alignment($r_{BBH,IFEval}=0.44, p<0.01$), suggesting our alignment measure captures separate competencies unique to each benchmark.


\section{Conclusion}


The rapid advancement of language models in scale and diversity, alongside their varied, often unpredictable, benchmark performance, leaves open a fundamental question: what does it take to build human-like models? A gold standard for "human-like" requires not only human-like task performance but also internal representations and computations that mirror that of humans. Here, we attempt to identify the computational ingredients most predictive of human-model representational alignment, leveraging large-scale human triadic similarity judgments as our method and semantic embeddings of the THINGS concepts as our target.

We carefully constructed a suite of 77 open-weight language models that varied in several key computational ingredients including model scale, training dataset size, dimensionality of model embeddings, and post-training regimes, among others.
By comparing human-model semantic embedding alignment as a function of these model characteristics, we arrived at several insights.

First, we found two convergent pieces of evidence that modern post-training techniques are central to human-model alignment: (1) instruction-finetuning was the strongest predictor of alignment even after controlling for variance explained by other ingredients (Figure \ref{fig:t-scores}) and (2) alignment tended to increase across different stages of post-training (Figure \ref{fig:journey}).
The importance of context length in predicting alignment further suggests that the methods that enable long context-reasoning might bring models into closer alignment with humans.
Second, we observed that architectural dimensionality (embedding layers, MLP) were relatively more important than overall model size and training corpus size in predicting alignment, especially in the mixed-effects model suggesting that representational capacity is key to developing models that align with human representations. 
Third, we found that while some existing benchmarks were correlated with our measure of alignment (e.g., \textit{MMLU, BBH}), no single benchmark captured all the variance in representational alignment. Notably, benchmarks emphasizing broad semantic knowledge and its flexible deployment showed the strongest correlations, consistent with theories of how humans represent semantic knowledge \citep{rogers2004semantic, saxe2019mathematical}.

\vspace{-1em}
\paragraph{Limitations}
To our knowledge, this is the first attempt to isolate the computational ingredients predictive of human–LLM conceptual alignment, though several limitations remain. First, while we evaluated many core ingredients, training dataset composition (e.g., text–code balance, language coverage) could not be assessed due to limited documentation across models; future work should clarify which dataset properties enable semantic alignment. Second, certain model classes — SSMs, multimodal models, and mixture-of-experts — were sparsely represented in our suite, limiting the reliability of conclusions about those architectures. Third, triadic similarity judgments, while well-validated and scalable, capture only one facet of semantic knowledge; future work should examine how models \textit{deploy} such knowledge in open-ended reasoning tasks \citep{wong2025modeling, mahowald2024dissociatinglanguagethoughtlarge}.
Fourth, here we only present correlational evidence. To establish true causal validity of our claims, future work should seek to leverage large-scale, open, community-drive model-training efforts (such as \href{https://marin.community/}{Marin}) actively train models while varying core computational ingredients.

\textbf{Acknowledgments.} We thank members of the Knowledge and Concepts Lab and Lupyan Lab for helpful feedback. We also thank the Center for High Throughput Computing for providing the compute resources for this project. 

\textbf{Code and materials:} \href{https://github.com/Knowledge-and-Concepts-Lab/llm-alignment-benchmark}{https://github.com/Knowledge-and-Concepts-Lab/llm-alignment-benchmark}.

\bibliography{colm2026_conference}
\bibliographystyle{colm2026_conference}


\newpage
\appendix
\section{Appendix}

\subsection{Odd-One-Out Judgments $\to$ SPoSE Estimation (Human Semantic Embeddings)}
\label{sec:spose}

To establish a ground-truth human semantic space, we employ the SPoSE (Sparse Positive Similarity Embedding) framework applied to odd-one-out behavioral data from the THINGS dataset \citep{hebart2023things}.

\textit{SPoSE Algorithm.} Let $\mathcal{X}=\{1,\dots,n\}$ index $n$ items and let $X\in\mathbb{R}_{\ge0}^{n\times d}$ be a nonnegative, low-dimensional embedding with rows $x_i^\top\in\mathbb{R}_{\ge0}^d$. We define pairwise similarities using the inner product $s_{ij}=x_i^\top x_j$, yielding similarity matrix $S=X X^\top$. In an odd-one-out trial with unordered triplet $\{i,j,k\}$, participants select the \emph{most similar pair} among $\{(i,j),(i,k),(j,k)\}$. SPoSE models this choice behavior with a multinomial logit over pairwise similarities:
\begin{equation}
\Pr\!\big[(i,j)\text{ chosen}\mid \{i,j,k\},X\big]
=\frac{\exp(s_{ij})}{\exp(s_{ij})+\exp(s_{ik})+\exp(s_{jk})}.
\label{eq:spose-softmax}
\end{equation}

\textit{Estimation and Implementation.} The SPoSE estimate $\hat X$ is obtained via a sparsity-promoting MAP program that encourages both nonnegativity and interpretability:
\begin{equation}
\hat X~\in~\arg\max_{X\ge 0}\Big\{
\underbrace{\sum_{t=1}^{T}\log p\big(y_t\mid \{i_t,j_t,k_t\},X\big)}_{\text{log-likelihood}}
~-\;\lambda \|X\|_{1}\Big\},
\label{eq:spose-objective}
\end{equation}
where $y_t$ is the chosen pair on trial $t$, and $\lambda>0$ controls sparsity.\footnote{The nonnegativity constraint and elementwise $\ell_1$ penalty are equivalent to an exponential prior on factors and empirically yield interpretable semantic dimensions.} We use the publicly released SPoSE embeddings as our human reference space.

\textit{Dimensionality Selection.} To determine the appropriate embedding dimension, we apply PCA to the released SPoSE matrix and retain the smallest $d$ that explains at least $95\%$ of the variance (Appendix~\ref{app:dim-select}). For our corpus of $n=128$ concepts, this yields $d=\hat{d}$ dimensions (\emph{30}), which we use as the target dimensionality for model embeddings. Additionally, we validate this dimensionality empirically by running models in the range $\hat{d}=20...30$ and observe relative stability in embeddings loss for all $\hat{d}$) (see Appendix \ref{app:dimension_compare}).

\subsection{Anchored Similarity Judgments $\to$ Ordinal Embedding (Model Semantic Embeddings)}
\label{sec:ordinal-emb}

To obtain comparable model-based semantic embeddings, we apply an ordinal embedding algorithm \citep{tamuz2011adaptively, sievert2023efficiently} to model similarity judgments, creating semantic spaces where frequently co-judged similar items are positioned closer together.

\textit{Embedding Algorithm.} Each triplet $\{i,j,k\}$ from model judgments yields an ordinal constraint: ``$i$ is closer to $j$ than to $k$.'' Let $\{x_i^\star\}_{i=1}^n\subset\mathbb{R}^d$ denote the unknown true point locations we seek to estimate, with corresponding (squared) Euclidean distance matrix $D^\star$. We model the probability of observing each ordinal constraint using a standard noisy triplet model with monotone link function:
\begin{equation}
\Pr\big[y_{(i;j,k)}=1\big]~=~f\!\big(D^\star_{ik}-D^\star_{ij}\big), \qquad f(0)=\tfrac12.
\label{eq:link}
\end{equation}

\textit{Estimation and Implementation.} The ordinal embedding algorithm minimizes crowd-kernel triplet loss \citep{tamuz2011adaptively} by optimizing Euclidean distances between item pairs. We minimize an empirical surrogate of the negative log-likelihood using a low-rank Gram matrix parameterization. To ensure reliable estimates, we reserve 20\% of our triplet data for validation purposes and monitor crowd-kernel loss throughout the fitting procedure.

\textit{Sample Complexity Considerations.} When the true embedding has rank $d$ and triplet judgments are sampled approximately uniformly at random, finite-sample theory provides the out-of-sample prediction error bound:
\begin{equation}
\mathcal{E}_{\mathrm{pred}}~=~\tilde O\!\left(\sqrt{\frac{d\,n\log n}{|S|}}\right),
\qquad
\Rightarrow\quad
|S|~=~\tilde \Theta\!\big(d\,n\log n\big).
\label{eq:pred-error}
\end{equation}
Thus, $\tilde \Theta(nd\log n)$ triplet judgments suffice for accurate prediction of new comparisons, and at least $\Omega(d\,n\log n)$ ordinal comparisons are information-theoretically necessary \citep{jainjamiesonnowak}. Guided by this theory and setting the model embedding dimension to match the human-derived $\hat{d}$, we collect
\begin{equation}
N_{\mathrm{triplets}}~=~c\,d\,n\log n
\end{equation}
triplet judgments per model, where $c$ is chosen to balance statistical efficiency with computational constraints.

\vspace{0.5em}
\noindent This parallel approach provides a principled framework for human-model semantic comparison. Both methods yield $d$-dimensional embeddings where geometric relationships reflect semantic similarities, with the SPoSE framework extracting interpretable human semantic structure and ordinal embedding converting model judgments into geometrically comparable representations with theoretically grounded sample complexity $\tilde \Theta(d\,n\log n)$.

\subsection{Effect of post-training techniques on model alignment}

\begin{table}[htbp]
\centering
\caption{Model Alignment Across Training Stages}
\label{tab:alignment_stages}
\begin{tabular}{llr}
\toprule
Model & Training Stage & Alignment (R²) \\
\midrule
Instella 3B & Base Stage 1 & 0.106 \\
Instella 3B & Base & 0.241 \\
Instella 3B & SFT & 0.293 \\
Instella 3B & DPO & 0.374 \\
Llama 3.1 8B & Base & 0.358 \\
Llama 3.1 8B & SFT & 0.301 \\
Llama 3.1 8B & DPO & 0.352 \\
Llama 3.1 8B & RLVR & 0.424 \\
OLMo 13B & Base & 0.299 \\
OLMo 13B & SFT & 0.392 \\
OLMo 13B & DPO & 0.406 \\
OLMo 13B & RLVR & 0.414 \\
OLMo 7B & Base & 0.179 \\
OLMo 7B & SFT & 0.347 \\
OLMo 7B & DPO & 0.357 \\
OLMo 7B & RLVR & 0.241 \\
\midrule
Average & Base Stage 1 & 0.106 \\
Average & Base & 0.269 \\
Average & SFT & 0.333 \\
Average & DPO & 0.372 \\
Average & RLVR & 0.360 \\
\bottomrule
\end{tabular}
\end{table}

\subsection{Object Concepts}
\label{app:concepts}

The following table presents the 128 concrete object concepts from the THINGS dataset used in our experiments, along with their semantic categories and definitions.

\begin{xltabular}{\textwidth}{@{}
  >{\RaggedRight\arraybackslash}p{0.16\textwidth}
  >{\RaggedRight\arraybackslash}p{0.13\textwidth}
  >{\RaggedRight\arraybackslash}X@{}}
  \caption{Complete list of object concepts with definitions.}
\label{tab:concepts} \\
\toprule
\textbf{Concept} &  \textbf{Category}   & \textbf{Definition} \\
\midrule
\endfirsthead
\multicolumn{3}{c}{{\bfseries Table \thetable\ continued from previous page}} \\
\toprule
\textbf{Concept} &  \textbf{Category}  & \textbf{Definition} \\
\midrule
\endhead
\midrule
\multicolumn{3}{r}{\footnotesize\emph{continued on next page}} \\
\endfoot
\bottomrule
\endlastfoot

badger & animal & sturdy carnivorous burrowing mammal with strong claws; widely distributed in the northern hemisphere \\
bear & animal & massive plantigrade carnivorous or omnivorous mammals with long shaggy coats and strong claws \\
butterfly & animal & diurnal insect typically having a slender body with knobbed antennae and broad colorful wings \\
chipmunk & animal & a burrowing ground squirrel of western America and Asia; has cheek pouches and a light and dark stripe running down the body \\
cow & animal & female of domestic cattle: "`moo-cow' is a child's term" \\
crow & animal & black birds having a raucous call \\
fly & animal & two-winged insects characterized by active flight \\
giraffe & animal & tallest living quadruped; having a spotted coat and small horns and very long neck and legs; of savannahs of tropical Africa \\
grasshopper & animal & terrestrial plant-eating insect with hind legs adapted for leaping \\
hyena & animal & doglike nocturnal mammal of Africa and southern Asia that feeds chiefly on carrion \\
iguana & animal & large herbivorous tropical American arboreal lizards with a spiny crest along the back; used as human food in Central America and South America \\
lion & animal & large gregarious predatory feline of Africa and India having a tawny coat with a shaggy mane in the male \\
mosquito & animal & two-winged insect whose female has a long proboscis to pierce the skin and suck the blood of humans and animals \\
mouse & animal & any of numerous small rodents typically resembling diminutive rats having pointed snouts and small ears on elongated bodies with slender usually hairless tails \\ \\
owl & animal & nocturnal bird of prey with hawk-like beak and claws and large head with front-facing eyes \\
parrot & animal & usually brightly colored zygodactyl tropical birds with short hooked beaks and the ability to mimic sounds \\
puffin & animal & any of two genera of northern seabirds having short necks and brightly colored compressed bills \\
snail & animal & freshwater or marine or terrestrial gastropod mollusk usually having an external enclosing spiral shell \\
snake & animal & limbless scaly elongate reptile; some are venomous \\
tarantula & animal & large hairy tropical spider with fangs that can inflict painful but not highly venomous bites \\

bathrobe & clothing & a loose-fitting robe of towelling; worn after a bath or swim \\
beanie & clothing & a small skullcap; formerly worn by schoolboys and college freshmen \\
bow & clothing & a knot with two loops and loose ends; used to tie shoelaces \\
bracelet & clothing & jewelry worn around the wrist for decoration \\
button & clothing & a round fastener sewn to shirts and coats etc to fit through buttonholes \\
chaps & clothing & (usually in the plural) leather leggings without a seat; joined by a belt; often have flared outer flaps; worn over trousers by cowboys to protect their legs \\
hat & clothing & headdress that protects the head from bad weather; has shaped crown and usually a brim \\
jeans & clothing & (usually plural) close-fitting trousers of heavy denim for manual work or casual wear \\
kilt & clothing & a knee-length pleated tartan skirt worn by men as part of the traditional dress in the Highlands of northern Scotland \\
kimono & clothing & a loose robe; imitated from robes originally worn by Japanese \\
kneepad & clothing & protective garment consisting of a pad worn by football or baseball or hockey players \\
tuxedo & clothing & semiformal evening dress for men \\
uniform & clothing & clothing of distinctive design worn by members of a particular group as a means of identification \\
veil & clothing & a garment that covers the head and face \\
visor & clothing & a brim that projects to the front to shade the eyes \\

bag & container & a flexible container with a single opening \\
cooker & container & a utensil for cooking \\
doily & container & a small round piece of linen placed under a dish or bowl \\
fishbowl & container & a transparent bowl in which small fish are kept \\
honeypot & container & South African shrub whose flowers when open are cup-shaped resembling artichokes \\
thermos & container & vacuum flask that preserves temperature of hot or cold drinks \\
vase & container & an open jar of glass or porcelain used as an ornament or to hold flowers \\

appetizer & food & food or drink to stimulate the appetite (usually served before a meal or as the first course) \\
applesauce & food & puree of stewed apples usually sweetened and spiced \\
baklava & food & rich Middle Eastern cake made of thin layers of flaky pastry filled with nuts and honey \\
beer & food & a general name for alcoholic beverages made by fermenting a cereal (or mixture of cereals) flavored with hops \\
bok choy & food & Asiatic plant grown for its cluster of edible white stalks with dark green leaves \\
bread & food & food made from dough of flour or meal and usually raised with yeast or baking powder and then baked \\
crepe & food & small very thin pancake \\
cupcake & food & small cake baked in a muffin tin \\
dessert & food & a dish served as the last course of a meal \\
dough & food & a flour mixture stiff enough to knead or roll \\
enchilada & food & tortilla with meat filling baked in tomato sauce seasoned with chili \\
grapefruit & food & large yellow fruit with somewhat acid juicy pulp; usual serving consists of a half \\
gravy & food & a sauce made by adding stock, flour, or other ingredients to the juice and fat that drips from cooking meats \\
hummus & food & a thick spread made from mashed chickpeas, tahini, lemon juice and garlic; used especially as a dip for pita; originated in the Middle East \\
leek & food & plant having a large slender white bulb and flat overlapping dark green leaves; used in cooking; believed derived from the wild Allium ampeloprasum \\
mango & food & large oval tropical fruit having smooth skin, juicy aromatic pulp, and a large hairy seed \\
margarita & food & a cocktail made of tequila and triple sec with lime and lemon juice \\
mashed potato & food & potato that has been peeled and boiled and then mashed \\
milkshake & food & frothy drink of milk and flavoring and sometimes fruit or ice cream \\
oatmeal & food & porridge made of rolled oats \\
parfait & food & layers of ice cream and syrup and whipped cream \\
pumpkin & food & usually large pulpy deep-yellow round fruit of the squash family maturing in late summer or early autumn \\
quesadilla & food & a tortilla that is filled with cheese and heated \\
quiche & food & a tart filled with rich unsweetened custard; often contains other ingredients (as cheese or ham or seafood or vegetables) \\
ravioli & food & small circular or square cases of dough with savory fillings \\
sea urchin & food & shallow-water echinoderms having soft bodies enclosed in thin spiny globular shells \\
souffle & food & light fluffy dish of egg yolks and stiffly beaten egg whites mixed with e.g. cheese or fish or fruit \\
stew & food & food prepared by stewing especially meat or fish with vegetables \\
tortellini & food & small ring-shaped stuffed pasta \\
waffle & food & pancake batter baked in a waffle iron \\
wrap & food & a sandwich in which the filling is rolled up in a soft tortilla \\

bassinet & furniture & a basket (usually hooded) used as a baby's bed \\
bathtub & furniture & a relatively large open container that you fill with water and use to wash the body \\
beanbag & furniture & a small cloth bag filled with dried beans; thrown in games \\
bed & furniture & a piece of furniture that provides a place to sleep \\
bench & furniture & a long seat for more than one person \\
coaster & furniture & a covering (plate or mat) that protects the surface of a table (i.e., from the condensation on a cold glass or bottle) \\
computer screen & furniture & a screen used to display the output of a computer to the user \\
cot & furniture & a small bed that folds up for storage or transport \\
couch & furniture & an upholstered seat for more than one person \\
crib & furniture & baby bed with high sides made of slats \\

album & other & a book of blank pages with pockets or envelopes; for organizing photographs or stamp collections etc \\
backscratcher & other & a long-handled scratcher for scratching your back \\
banjo & other & a stringed instrument of the guitar family that has long neck and circular body \\
baseball & other & a ball used in playing baseball \\
baseball glove & other & the handwear used by fielders in playing baseball \\
bassoon & other & a double-reed instrument; the tenor of the oboe family \\
beachball & other & large and light ball; for play at the seaside \\
blower & other & a device that produces a current of air \\
bongo & other & a small drum; played with the hands \\
cannonball & other & a solid projectile that in former times was fired from a cannon \\
canvas & other & an oil painting on canvas fabric \\
chainsaw & other & portable power saw; teeth linked to form an endless chain \\
cymbal & other & a percussion instrument consisting of a concave brass disk; makes a loud crashing sound when hit with a drumstick or when two are struck together \\
doorknob & other & a knob used to release the catch when opening a door (often called `doorhandle' in Great Britain) \\
doorknocker & other & a device (usually metal and ornamental) attached by a hinge to a door \\
extinguisher & other & a manually operated device for extinguishing small fires \\
fire alarm & other & a device that makes a loud sound to warn people when there is a fire \\
guitar & other & a stringed instrument usually having six strings; played by strumming or plucking \\
hatchet & other & a small ax with a short handle used with one hand (usually to chop wood) \\
hula hoop & other & a large hoop spun around the body by gyrating the hips, for play or exercise. \\
knitting needle & other & needle consisting of a slender rod with pointed ends; usually used in pairs \\
knob & other & a round handle \\
lawnmower & other & garden tool for mowing grass on lawns \\
pinball & other & a game played on a sloping board; the object is to propel marbles against pins or into pockets \\
pocket watch & other & a watch that is carried in a small watch pocket \\
shuffleboard & other & a game in which players use long sticks to shove wooden disks onto the scoring area marked on a smooth surface \\
skeleton & other & the hard structure (bones and cartilages) that provides a frame for the body of an animal \\
swab & other & implement consisting of a small piece of cotton that is used to apply medication or cleanse a wound or obtain a specimen of a secretion \\
tennis ball & other & ball about the size of a fist used in playing tennis \\
toilet paper & other & a soft thin absorbent paper for use in toilets \\
treadmill & other & an exercise device consisting of an endless belt on which a person can walk or jog without changing place \\
trowel & other & a small hand tool with a handle and flat metal blade; used for scooping or spreading plaster or similar materials \\
volleyball & other & an inflated ball used in playing volleyball \\

flower & plant & a plant cultivated for its blooms or blossoms \\
gourd & plant & any of numerous inedible fruits with hard rinds \\

airbag & vehicle & a safety restraint in an automobile; the bag inflates on collision and prevents the driver or passenger from being thrown forward \\
carriage & vehicle & a vehicle with wheels drawn by one or more horses \\
exhaust pipe & vehicle & a pipe through which burned gases travel from the exhaust manifold to the muffler \\
hot-air balloon & vehicle & balloon for travel through the air in a basket suspended below a large bag of heated air \\
humvee & vehicle & a high mobility, multipurpose, military vehicle with four-wheel drive \\
missile & vehicle & a rocket carrying a warhead of conventional or nuclear explosives; may be ballistic or directed by remote control \\
odometer & vehicle & a meter that shows mileage traversed \\
taillight & vehicle & lamp (usually red) mounted at the rear of a motor vehicle \\
tank & vehicle & an enclosed armored military vehicle; has a cannon and moves on caterpillar treads \\
taxi & vehicle & a car driven by a person whose job is to take passengers where they want to go in exchange for money \\
\end{xltabular}


\subsection{Model Suite}

\begin{longtable}{@{\extracolsep{\fill}}p{0.35\textwidth}p{0.15\textwidth}p{0.15\textwidth}p{0.15\textwidth}p{0.12\textwidth}@{}}
\caption{List of all the models used}\\
\toprule
\textbf{Model} & \textbf{Attention Type} & \textbf{Training Tokens} & \textbf{\#Heads} & \textbf{Hidden Dim} \\
\midrule
\endfirsthead

\multicolumn{5}{c}%
{{\bfseries Table \thetable\ continued from previous page}} \\
\toprule
\textbf{Model} & \textbf{Attention Type} & \textbf{Training Tokens} & \textbf{\#Heads} & \textbf{Hidden Dim} \\
\midrule
\endhead

\midrule
\multicolumn{5}{r}{{Continued on next page}} \\
\endfoot

\bottomrule
\endlastfoot

\textbf{Falcon3-3B-Base}~\cite{Falcon3} & GQA & 100B & 12 (4 KV) & 3072 \\
\textbf{tiiuae/falcon-7b}~\cite{Falcon3} & MQA & 1.5T & 71 & 4544 \\
\textbf{tiiuae/falcon-7b-instruct}~\cite{Falcon3} & MQA & {} & 71 & 4544 \\
\textbf{Falcon3-10B-Base}~\cite{Falcon3} & GQA & 2T & 12 (4 KV) & 3072 \\
\textbf{tiiuae/falcon-11b}~\cite{Falcon3} & GQA & 5T & 32 (8 KV) & 4096 \\[1ex]

\textbf{flan-t5-small}~\cite{Chung2022} & Multi-Head & {} & 8 & 512 \\
\textbf{flan-t5-large}~\cite{Chung2022} & Multi-Head & {} & 16 & 1024 \\
\textbf{flan-t5-xl}~\cite{Chung2022} & Multi-Head & {} & 32 & 1024 \\
\textbf{flan-t5-xxl}~\cite{Chung2022} & Multi-Head & {} & 128 & 1024 \\[1ex]

\textbf{gemma-3-270m}~\cite{Gemma2024} & GQA & {} & 8 (shared KV) & 2048 \\
\textbf{gemma-3-270m-it}~\cite{Gemma2024} & GQA & {} & 8 (shared KV) & 2048 \\
\textbf{gemma-2-2b}~\cite{Gemma2024} & MQA & {} & 8 (shared KV) & 2048 \\
\textbf{gemma-2-2b-it}~\cite{Gemma2024} & MQA & {} & 8 (shared KV) & 2048 \\
\textbf{gemma-2-9b}~\cite{Gemma2024} & MHA & {} & 16 & 4096 \\
\textbf{gemma-2-9b-it}~\cite{Gemma2024} & MHA & {} & 16 & 4096 \\
\textbf{gemma-2-27b}~\cite{Gemma2024} & MHA & {} & 16 & 6144 \\
\textbf{gemma-2-27b-it}~\cite{Gemma2024} & MHA & {} & 16 & 6144 \\
\textbf{gemma-3-4b-it}~\cite{Gemma2024} & GQA & {} & 12 (4 KV) & 3072 \\
\textbf{gemma-3-4b-pt}~\cite{Gemma2024} & GQA & {} & 12 (4 KV) & 3072 \\
\textbf{gemma-3-12b-it}~\cite{Gemma2024} & GQA & {} & 16 (4 KV) & 4096 \\
\textbf{gemma-3-12b-pt}~\cite{Gemma2024} & GQA & {} & 16 (4 KV) & 4096 \\
\textbf{gemma-3-27b}~\cite{Gemma2024} & GQA & {} & 16 (4 KV) & 5120 \\
\textbf{gemma-3-27b-it}~\cite{Gemma2024} & GQA & {} & 16 (4 KV) & 5120 \\[1ex]

\textbf{GPT-OSS-20B}~\cite{Black2022} & Multi-Head & $\sim$300B & 64 & 6144 \\
\textbf{amd/Instella-3B}~\cite{Liu2025} & Multi-Head & 4.15T & 32 & 2560 \\
\textbf{amd/Instella-3B-Instruct}~\cite{Liu2025} & Multi-Head & {} & 32 & 2560 \\
\textbf{Llama-3.1-8B-Instruct}~\cite{Meta2024} & MHA & {} & 32 & 4096 \\[1ex]

\textbf{state-spaces/mamba-1.4b-hf}~\cite{gu2023mamba} & (SSM) & {} & {} & {} \\[1ex]

\textbf{allenai/OLMo-2-1124-7B}~\cite{Groeneveld2024} & MHA & {} & 32 & 4096 \\
\textbf{allenai/OLMo-2-1124-7B-Instruct}~\cite{Groeneveld2024} & MHA & {} & 32 & 4096 \\
\textbf{allenai/OLMo-2-1124-13B-Instruct}~\cite{Groeneveld2024} & MHA & {} & 40 & 5120 \\
\textbf{OLMo-2-1124-7B-DPO}~\cite{Groeneveld2024} & MHA & {} & 32 & 4096 \\
\textbf{OLMo-2-1124-7B-SFT}~\cite{Groeneveld2024} & MHA & {} & 32 & 4096 \\
\textbf{AMD-OLMo-1B}~\cite{Groeneveld2024} & MHA & {} & 16 & 2048 \\
\textbf{AMD-OLMo-1B-SFT-DPO}~\cite{Groeneveld2024} & MHA & {} & 16 & 2048 \\[1ex]

\textbf{GPT-J-6B}~\cite{EleutherAI2021} & Multi-Head & $\sim$300B & 16 & 4096 \\[1ex]

\textbf{orca-2-7b}~\cite{Mukherjee2023} & Multi-Head & {} & 32 & 4096 \\
\textbf{orca-2-13b}~\cite{Mukherjee2023} & Multi-Head & {} & 40 & 5120 \\[1ex]

\textbf{phi-3.5-mini-instruct}~\cite{Gunasekar2023} & Multi-Head & {} & 32 & 2048 \\
\textbf{phi-3-mini-128k-instruct}~\cite{Gunasekar2023} & Multi-Head & {} & 32 & 2048 \\
\textbf{phi-3-medium-128k-instruct}~\cite{Gunasekar2023} & Multi-Head & {} & 40 & 5120 \\
\textbf{Phi-3-medium-4k-instruct}~\cite{Gunasekar2023} & Multi-Head & {} & 40 & 5120 \\
\textbf{phi-1\_5}~\cite{Gunasekar2023} & Multi-Head & 7B & 32 & 2048 \\[1ex]

\textbf{qwen-14b}~\cite{Bai2023} & GQA & 3T & 40 (10 KV) & 5120 \\
\textbf{qwen-14b-chat}~\cite{Bai2023} & GQA & {} & 40 (10 KV) & 5120 \\
\textbf{qwen-2-7b}~\cite{Bai2023} & GQA & {} & 32 (8 KV) & 4096 \\
\textbf{qwen-2-7b-instruct}~\cite{Bai2023} & GQA & {} & 32 (8 KV) & 4096 \\
\textbf{qwen-2.5-7b}~\cite{Bai2023} & GQA & {} & 32 (8 KV) & 4096 \\
\textbf{qwen-2.5-7b-instruct}~\cite{Bai2023} & GQA & {} & 32 (8 KV) & 4096 \\
\textbf{qwen-2.5-14b}~\cite{Bai2023} & GQA & {} & 40 (10 KV) & 5120 \\
\textbf{qwen-2.5-14b-instruct}~\cite{Bai2023} & GQA & {} & 40 (10 KV) & 5120 \\[1ex]

\textbf{t5-3b}~\cite{Raffel2020} & Multi-Head & $\sim$1T & 32 & 1024 \\
\textbf{t5-11b}~\cite{Raffel2020} & Multi-Head & $\sim$1T & 128 & 1024 \\[1ex]

\textbf{yi-9B}~\cite{01AI2024} & Multi-Head & 3.1T & 32 & 6144 \\
\textbf{yi-9B (coder)}~\cite{01AI2024} & Multi-Head & 0.8T & 32 & 6144 \\
\end{longtable}

\subsection{Details for SPoSE (Odd-One-Out) Estimation}
\label{app:spose-details}

Given $S=X X^\top$ with $X\ge 0$, the triplet likelihood is the three-way softmax in \eqref{eq:spose-softmax}. The MAP program \eqref{eq:spose-objective} is optimized by minibatch stochastic gradients; $\lambda$ is tuned by held-out triplets. The released THINGS-data SPoSE embeddings are derived from large-scale odd-one-out judgments and approach the behavioral noise ceiling in predicting left-out triplets, despite far fewer than the $O(n^3)$ triplets required to fully determine a dense similarity matrix \citep{hebart2023things}.

\subsection{Ordinal Embedding: Risk Bounds and Recovery}
\label{app:ordinal-bounds}

Let $G^\star$ be the centered Gram matrix of $\{x_i^\star\}\subset\mathbb{R}^d$ and let $\mathcal{L}(G)$ denote the expected logistic triplet loss induced by \eqref{eq:link}. If $G^\star$ has rank $d$ and $|S|$ triplets are drawn uniformly at random, then with probability $1-\delta$,
\begin{equation}
\mathcal{L}(\hat G)-\mathcal{L}(G^\star)~\le~
C_1\sqrt{\frac{d\,n\log n}{|S|}}+C_2\sqrt{\frac{\log(1/\delta)}{|S|}},
\label{eq:risk-bound}
\end{equation}
for universal constants $C_1,C_2$ depending on the loss Lipschitz constant \citep{jainjamiesonnowak}. Moreover, writing $C^\star$ for the component of the EDM orthogonal to the linear operator's kernel, one can recover the distance structure with
\begin{equation}
\frac{1}{n^2}\,\|\hat C-C^\star\|_F^2~=~\tilde O\!\left(\frac{d\,n\log n}{|S|}\right),
\label{eq:dnlogn}
\end{equation}
which implies \eqref{eq:dnlogn} samples suffice up to logarithmic factors; $\Omega(d\,n\log n)$ lower bounds are known \citep{Jamieson2011_activeMDS}.

\subsection{Dimensionality Selection from Human SPoSE}
\label{app:dim-select}

We compute PCA on the released SPoSE embedding matrix (items $\times$ dimensions), retain the top $d$ components explaining at least $95\%$ variance, and set $d=\hat d$ for model-side ordinal embeddings. In our corpus ($n=128$), this yields $\hat d=\,$\emph{29}, and therefore $N_{\mathrm{triplets}}=c\,\hat d\,n\log n=\,$\emph{31,288} triplets per model where $c=4$. We increase $N_{\mathrm{triplets}}$ to \textbf{35,000} to provide a margin of error.

\subsection{Category and item-level alignment of THINGS concepts}

\textbf{Superordinate category-level alignment}

We first compute mean human-model Procrustes $R^2$ grouped by superordinate category label (\textit{furniture, container, vehicle, clothing, food, animal, other}) and model size (binned by parameter size $0B...1B$, $1B...3B$, $3B...10B$, and $108B+$). Results are shown in the figure below:

\begin{figure}[htbp]
    \centering
    \includegraphics[width=0.65\textwidth]{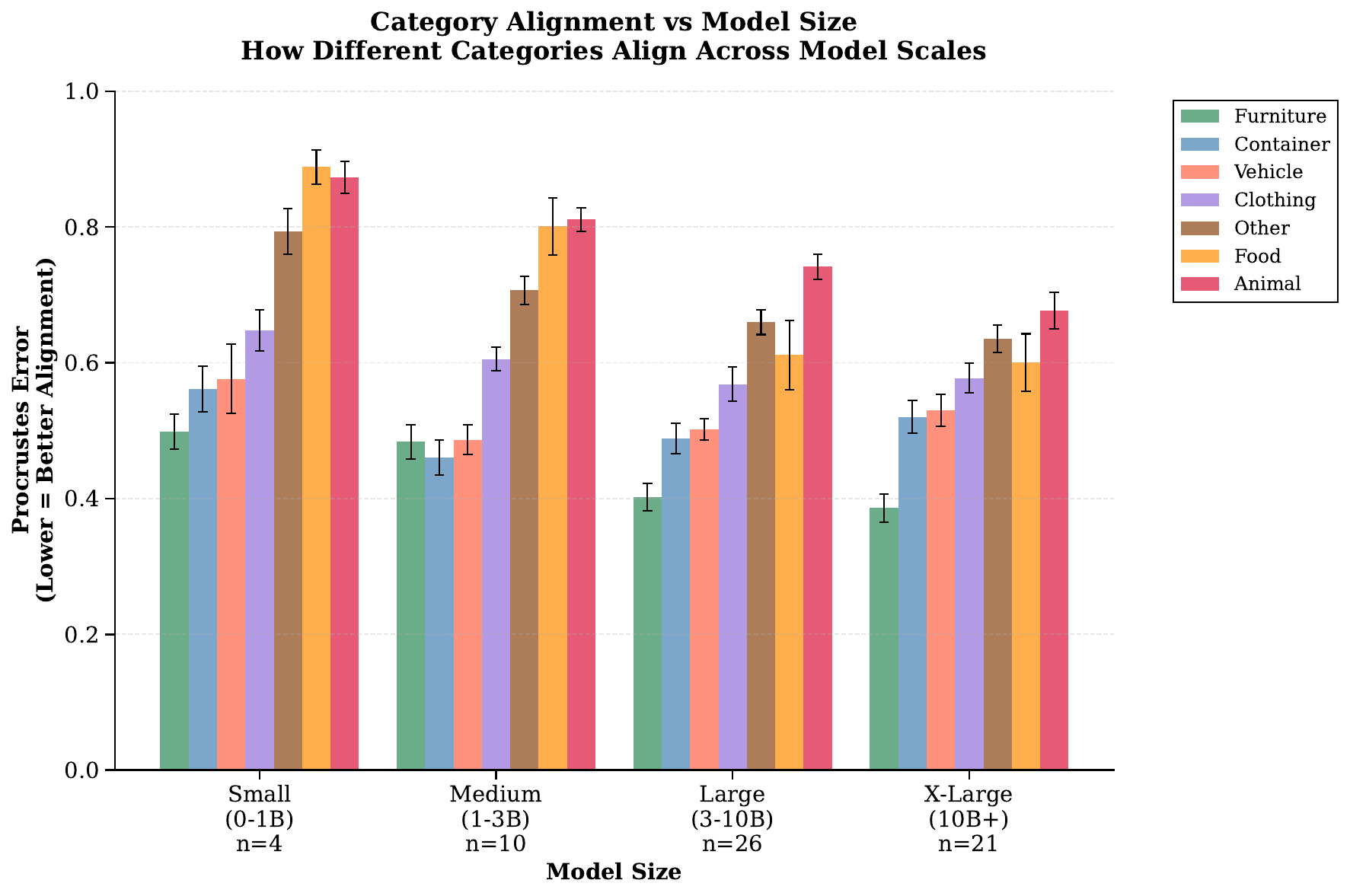}
    \caption{Alignment by superordinate category labels, grouped by model parameter size. Y-axis values indicate Procrustes error.}
    \label{fig:model_size_align}
\end{figure}

\textbf{Concept-level alignment}
We also test whether different concept attributes predict embeddings (cosine) distance between human and model concept representations. 
We use the following THINGS attribute dimensions as predictors of human-model alignment:
\begin{itemize}
    \item Typicality
    \item Familiarity
    \item Memorability
    \item Concreteness
    \item Frequency rank
    \item Frequency (COCA)
    \item Frequency (BNC)
    \item Dispersion
    \item Polysemy
\end{itemize}

We fit a mixed-effects regression with model as a random effect and each attribute dimension given as a fixed effect, defined as:

\begin{equation}
    d_{\text{cosine}_{ij}} = \beta_0 + \sum_{k=1}^{9} \beta_k X_{ki} + u_j + \epsilon_{ij}
\end{equation}

where $d_{\text{cosine}_{ij}}$ is the cosine distance between human and model embeddings for item $i$ and model $j$, $\beta_0$ is the intercept, $X_{ki}$ is the value of attribute $k$ for item $i$, $u_j$ is the random intercept for model $j$, and $\epsilon_{ij}$ is the residual error.

The t-values for all fixed-effects in this regression are included below.

\begin{figure}[htbp]
    \centering
    \includegraphics[width=0.85\textwidth]{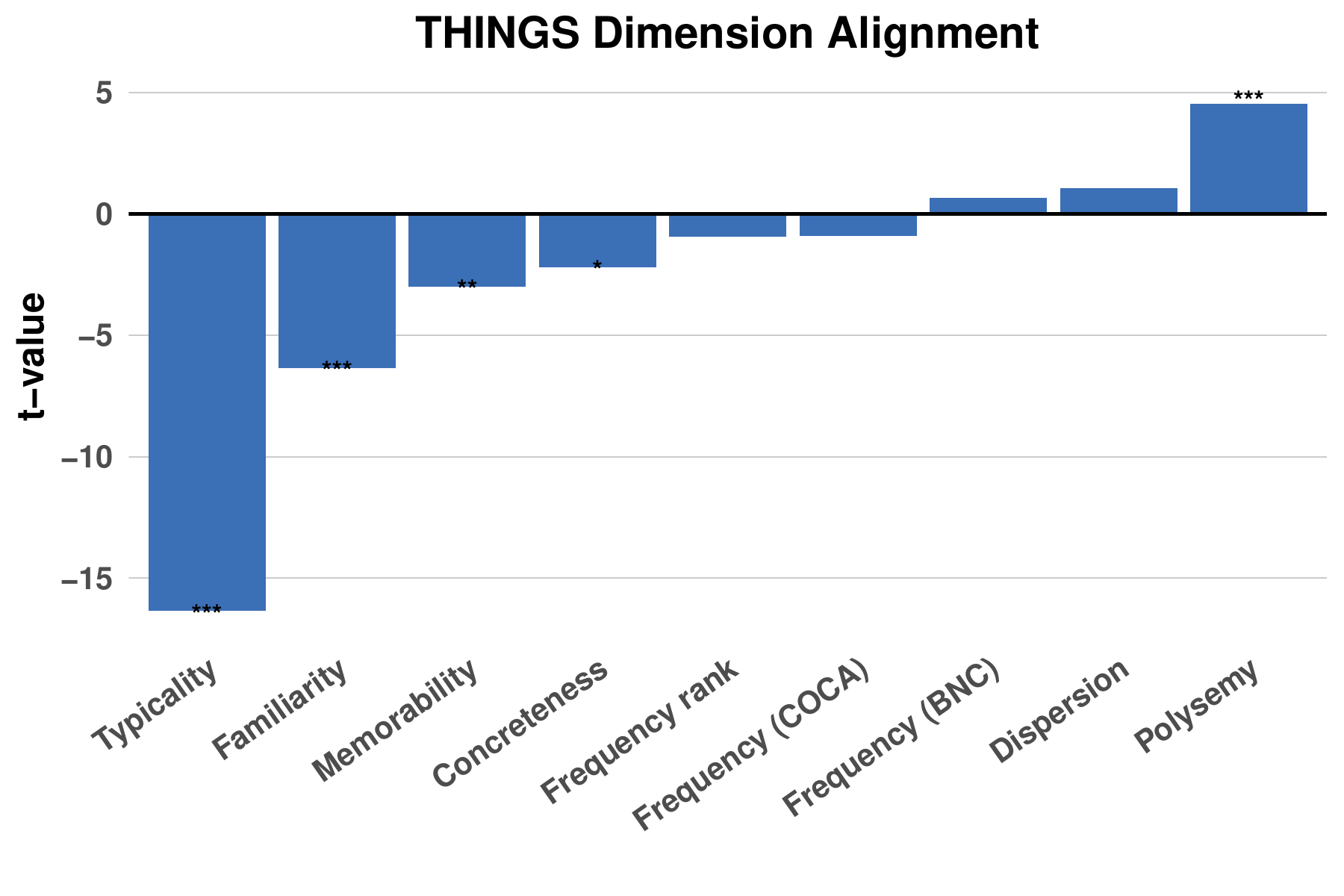}
    \caption{t-values for all THINGS attributes predicting cosine distance between human and model embeddings for each item.}
    \label{fig:dimension_alignment}
\end{figure}

\newpage

\subsection{A case study: the effects of post-training paradigms on alignment}
\label{app:case_study}
Modern LLM systems undergo various stages of post-training including \textit{supervised fine-tuning} (SFT) \citep{ouyang2022training}, \textit{direct preference optimization} DPO \citep{rafailov2023direct}, \textit{reinforcement learning from human feedback} (RLHF) \citep{ouyang2022training}, and \textit{instruction tuning} (IT) \citep{wei2021finetuned}. 
It remains unclear what impact these post-training steps have on human-model alignment. 
Progress on this front has been limited due to a lack of open-weight models that have been checkpointed at the various possible stages.
\texttt{OLMo} \citep{Groeneveld2024} is a notable exception since it is available at four stages of post-training, including the base model (no post-training), SFT, DPO, and IT and is available at two model sizes (7B and 13B).
\texttt{Instella} \citep{Liu2025} and \texttt{Llama-3.1} \citep{Meta2024} also have similar checkpointed model variants available with Instella having an additional stage of pretraining checkpointed.

\newpage

\subsection{Test loss and accuracy comparisons for different embeddings dimensionality}
\label{app:dimension_compare}

We re-compute embeddings for small, medium and large number parameter models and compare downstream performance of embeddings fit from those models with dimensions 20...30. We find that embeddings are largely similar in test loss and accuracy across different numbers of dimensions. 

\begin{figure}[htbp]
    \centering
    \includegraphics[width=1\textwidth]{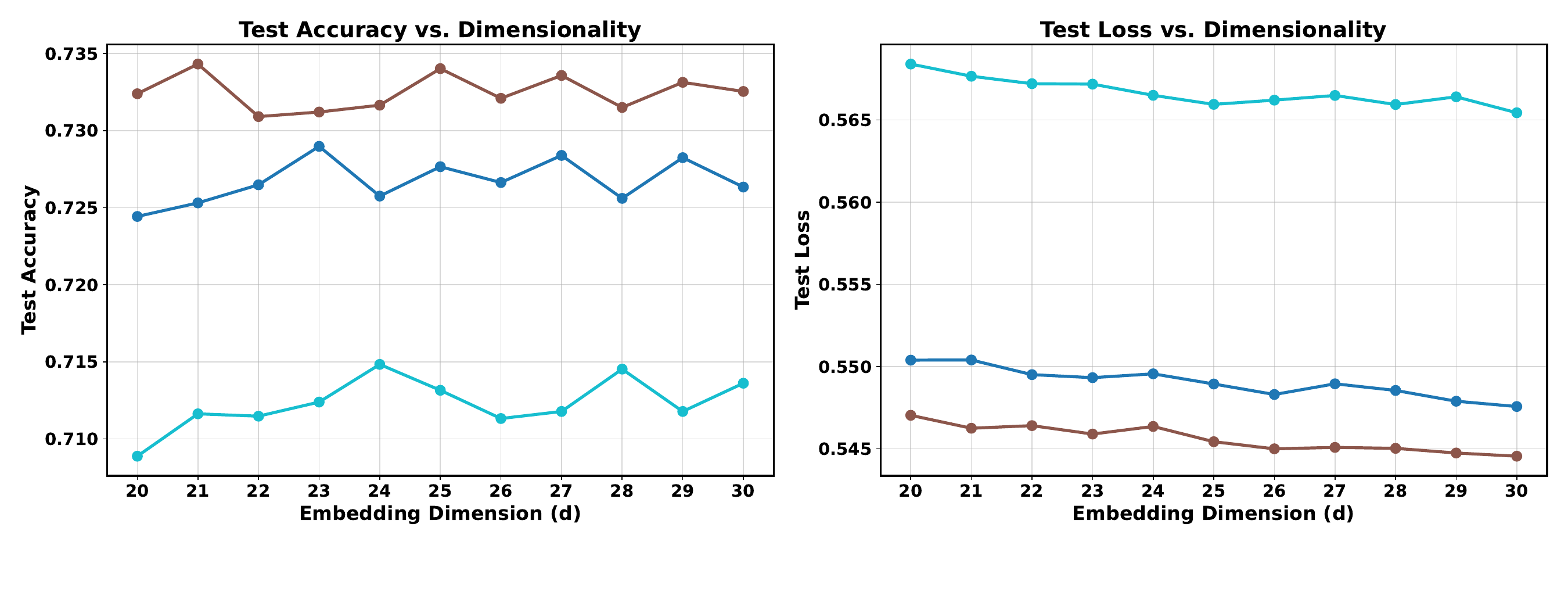}
    \caption{\textbf{Embeddings dimensionality comparisons}. Brown, blue, and aqua lines indicate test accuracy/loss for \texttt{gemma-3-27b-it}, \texttt{gemma-3-12b-it}, and \texttt{gemma-3-4b-it} respectively.}
    \label{fig:dimension_compare_fig}
\end{figure}

\subsection{Differences in SPoSE and crowd-kernel computed representations}
\label{app:rep-compare}

As human embeddings were computed using SPoSE and model embeddings were computed using the crowd-kernel algorithm, it is necessary to ensure that our human-model alignment results are not confounded by differences in implementation between SPoSE and crowd-kernel methods. To this end, we re-compute embeddings for 77 concepts where we have anchored human similarity judgments using both SPoSE and crowd-kernel algorithms, and then compute the Procrustes $R^2$ between the resulting embedding spaces. We observe extremely high alignment between embeddings computed from SPoSE and crowd-kernel ($R^2$=0.99) suggesting that differences due to embeddings algorithm are negligible. Comparisons of RSMs for embeddings computed with SPoSE and crowd-kernel are given below.

\newpage
\subsection{Human-Model alignment for abstract concepts}
\label{app:emotions-replicate}

\begin{figure}[ht!]
    \centering
    \includegraphics[width=1\textwidth]{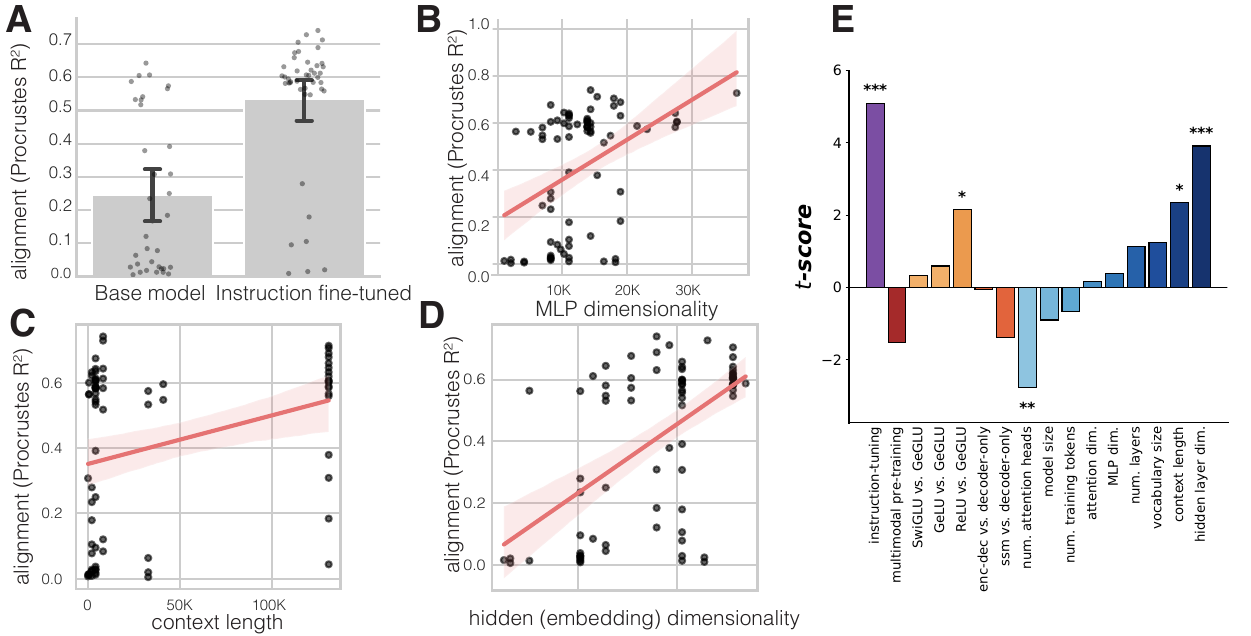}
    \caption{Factors contributing to human-model alignment for abstract emotions concepts.}
    \label{fig:emotion_results}
\end{figure}

In order to test whether our analyses generalized to non-concrete concepts, we replicated the exact same procedure described in the main text and obtained human and model semantic embeddings for a set of 213 emotion words derived from \cite{shaver1987emotion}. These words included terms like dread, love, alienation, rage, among others. In general, as can be seen in Figure \ref{fig:emotion_results}, we found thar larger context lengths, MLP and embedding dimensionality were strong predictors of alignment for these concepts as well. We also replicated the core finding that instruction-tuned models were also more aligned. The full set of $t$-scores for all predictors can be seen in Figure \ref{fig:emotion_results}E.

\clearpage
\label{app:measure_corrs}
\subsection{Measure correlations}
\begin{figure}[htbp]
    \centering
    \includegraphics[width=0.80\textwidth]{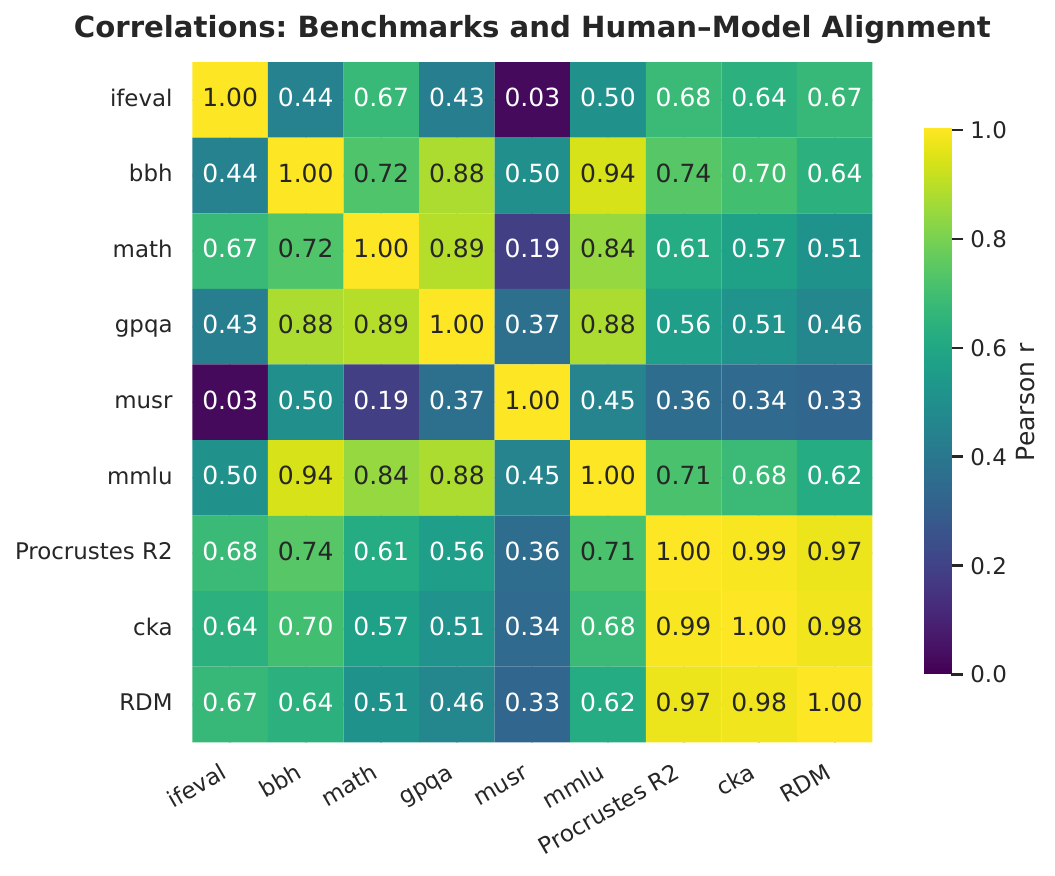}
    \caption{Correlations between measures of human-model alignment and model benchmarks, for models where benchmark scores are publicly are available. (Procrustes $R^2$, CKA, RSM)}
    \label{fig:human_embeddings_heatmap}
\end{figure}

\clearpage

\label{app:subsample_justification}
\subsection{Consistency in human-alignment for full and subsampled concept embeddings}

\begin{figure}[htbp]
    \centering
    \includegraphics[width=0.60\textwidth]{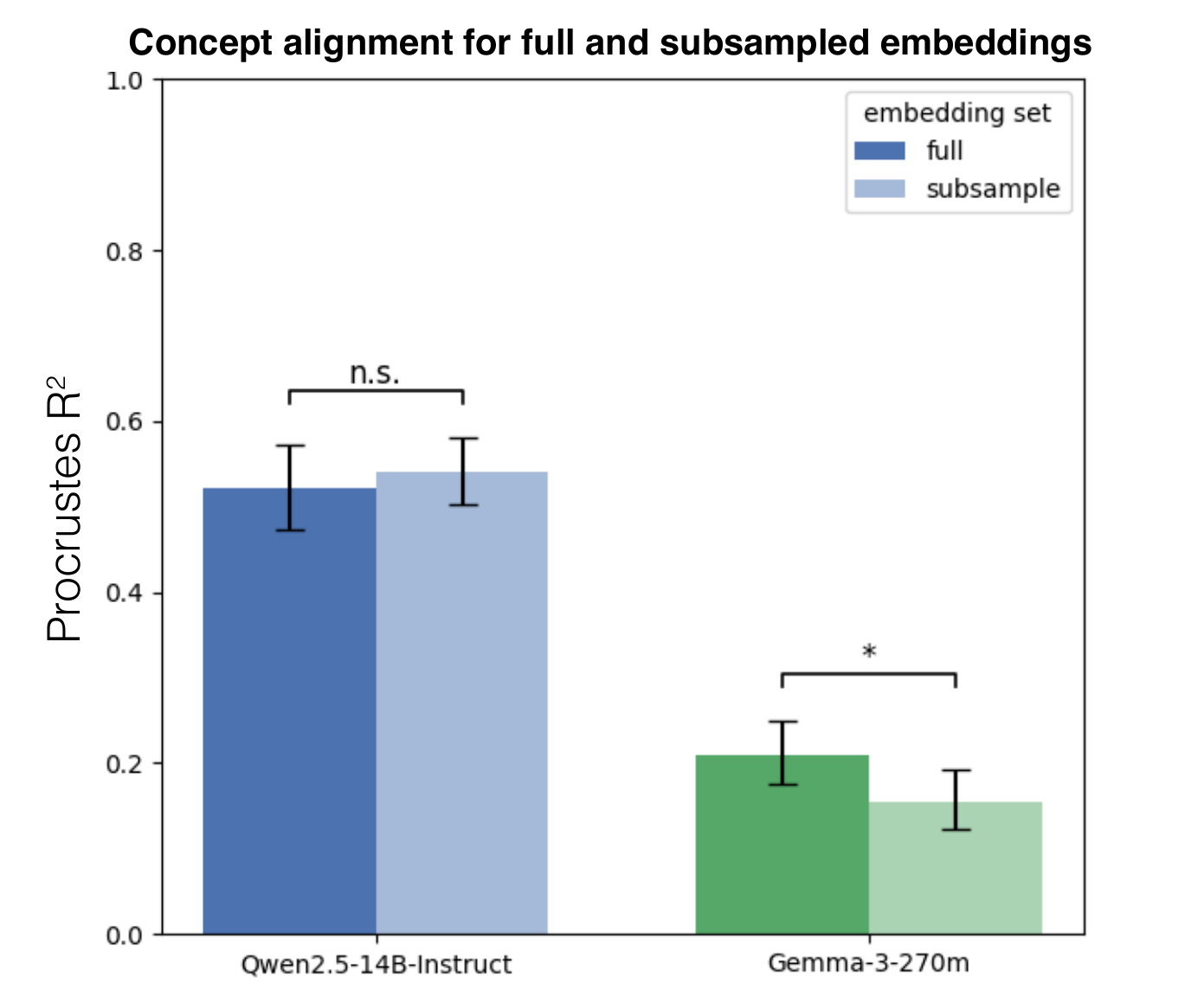}
    \caption{95\% bootstrap CIs for human-alignment of the 128 subsampled concepts, when computed from the subsampled or full embedding space.}
    \label{fig:subsample_align}
\end{figure}

While our procedure for subsampling embeddings (using $n$=128 concepts stratified across key object dimensions) is consistent with that of \citet{mukherjee2023seva}, it is an empirical question as to whether we would observe the same results if our embeddings were instead sampled using judgments from the full $n=1,824$ item space. To this end, we obtained 1.2 million triplets on all 1,854 items for the most aligned model (\texttt{qwen-2.5-14b}) and one of the least aligned models (\texttt{gemma-3-250m-it}) and tested for meaningful differences in the human alignment of the subsampled item embeddings when these embeddings were computed only in the 128 item subsample space, or taken from the full 1824 item space. We computed bootstrapped 95\% confidence intervals to test whether the human predictiveness of the subsample was meaningfully different from the full sample, for both \texttt{qwen-2.5-14b} and \texttt{gemma-3-250m-it.}  Figure \ref{fig:subsample_align} shows results for t-tests on differences of Procrustes alignment for \texttt{qwen-2.5-14b} and \texttt{gemma-3-250m-it.} We observe no meaningfull differences in alignment for the larger model \texttt{qwen-2.5-14b} ($p>0.05$), and small differences in alignment for the smallest model in our evaluation \texttt{gemma-3-250m-it} ($p<0.05$).

\clearpage

\label{app:architecture_rels}
\subsection{Relationships between computational ingredient properties and human-model alignment}
\begin{figure}[htbp]
    \centering
    \includegraphics[width=0.80\textwidth]{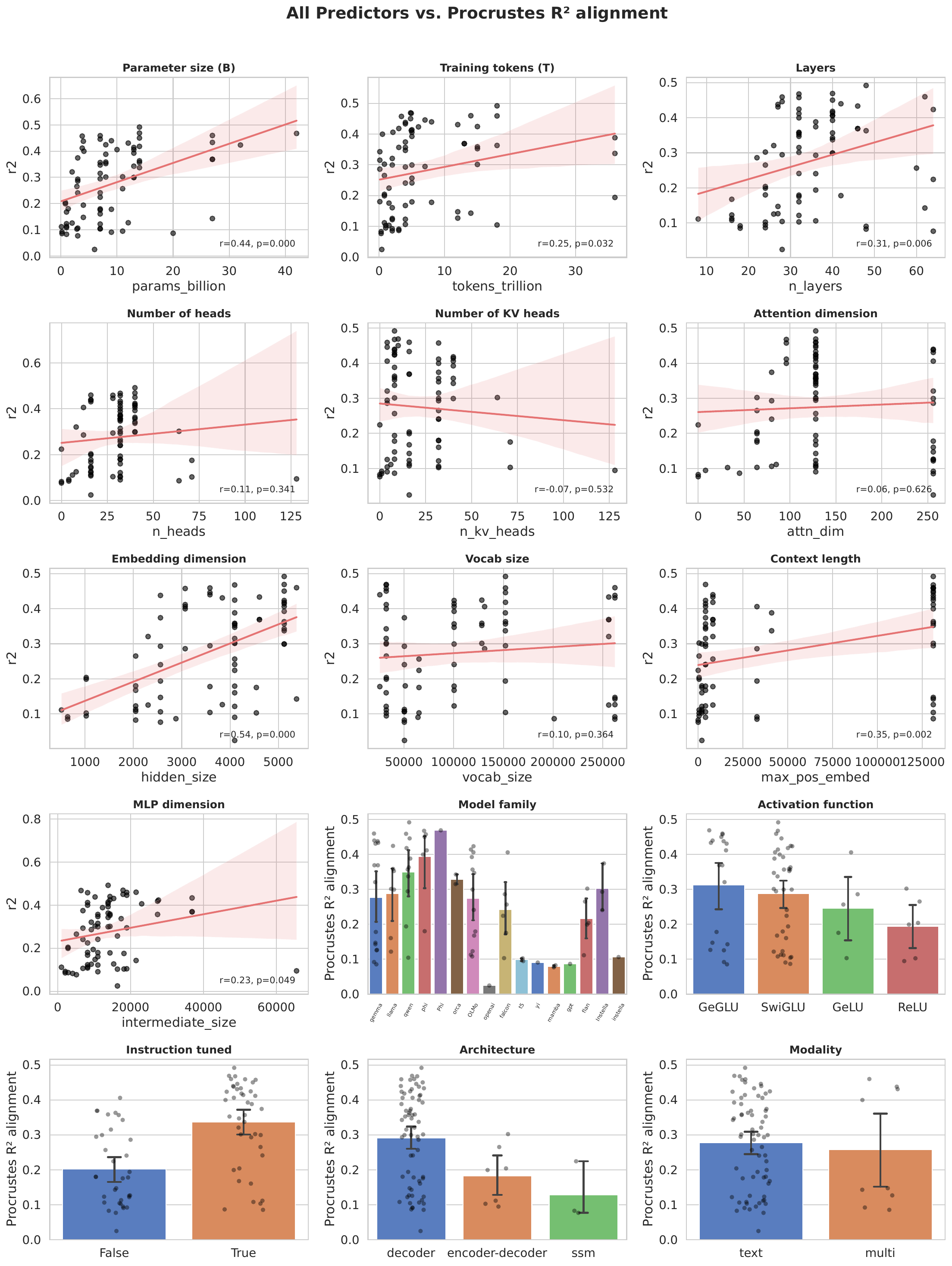}
    \caption{Categorical and continuous between all computational ingredients and human-model alignment. All significant effects in the full mixed linear-model are reported in the main text, Figure \ref{fig:t-scores}.}
    \label{fig:human_embeddings}
\end{figure}

\clearpage

\label{app:multicollinearity_tests}
\subsection{Testing for multicollinearity in predictors of human-model alignment}

\begin{figure}[htbp]
    \centering
    \includegraphics[width=0.6\textwidth]{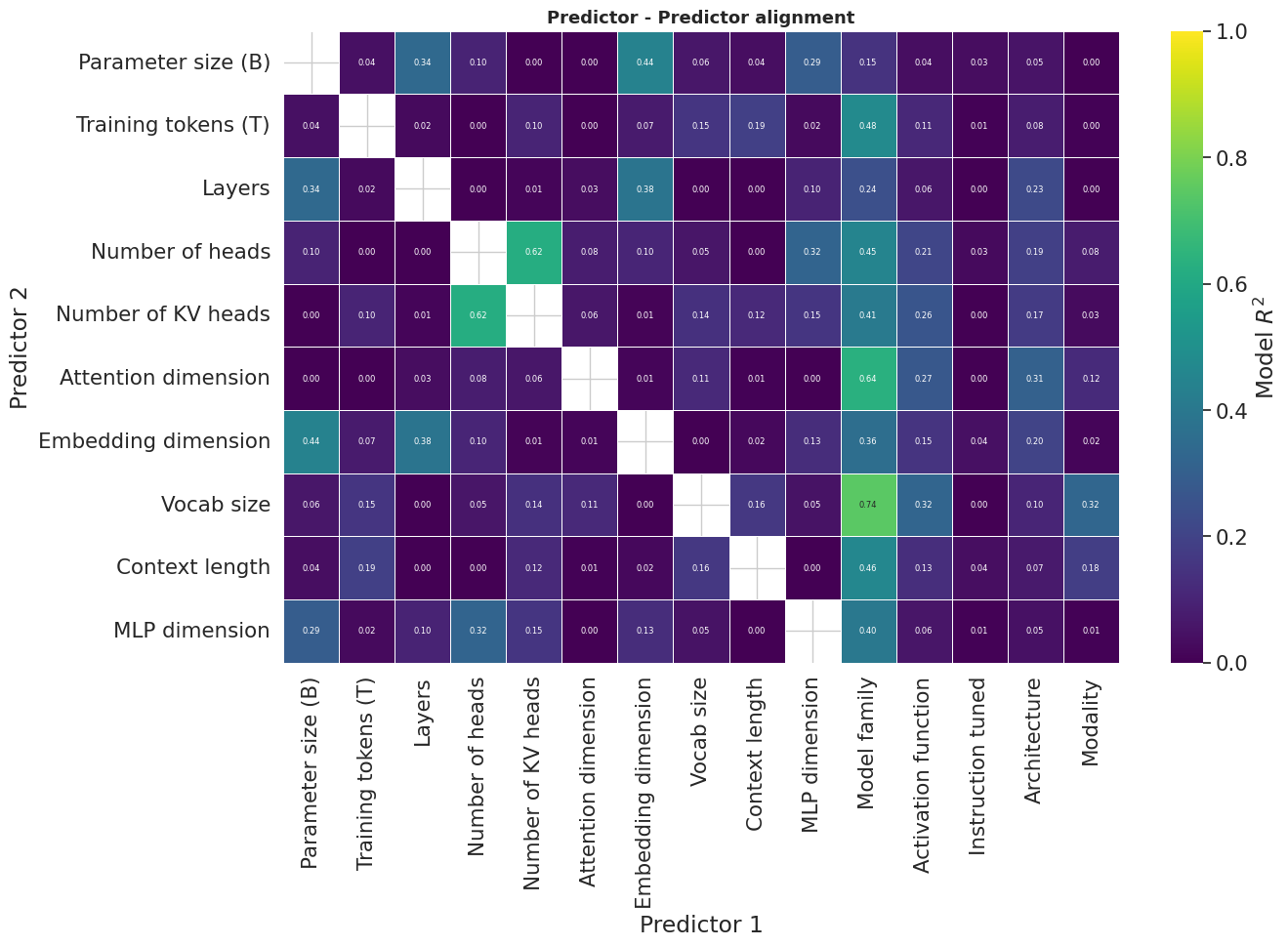}
    \caption{$R^2$ between individual predictors used in our mixed-effects linear models. For each unique predictor-predictor pair, a linear regression was fit predicting predictor 2 from the values of predictor 1.}
    \label{fig:pred_pred_cors}
\end{figure}

Mixed-effects linear models produce unbiased parameter estimates even in cases where level-1 predictors --in our case, computational ingredients-- are collinear \citep{shieh2003effect}. However, collinearity among these computational ingredients could present difficulties in future analyses if different statistical models were to be used. We measure the GVIF (generalized variance inflation factor, see \citet{fox1992generalized}) values of our fixed effects in both the THINGS and EMOTIONS regressions, and find that none of the significant predictors we report on exceed even an intermediate threshold of multicollinearity. Figure \ref{fig:vif_test} shows GVIF values of parameter estimates for both \texttt{EMOTIONS} and \texttt{THINGS} mixed-effects models. In both cases, we find that GVIF values do not exceed thresholds conventionally agreed upon as indicating high covariance. Additionally, variance explained ($R^2$) between individual predictors is relatively low for most predictor-predictor relationships, as shown qualitatively by Figure \ref{fig:pred_pred_cors}.

\begin{figure}[htbp]
    \centering
    \includegraphics[width=1\textwidth]{plots/gvif_emotions.pdf}
    \caption{GVIF values for mixed-effects linear models fit from \texttt{THINGS} (top) and \texttt{EMOTIONS} concepts (bottom). Dashed lines indicate GVIF values commonly used to indicate \textit{intermediate} and \textit{high} levels of multicollinearity among predictors.}
    \label{fig:vif_test}
\end{figure}

\clearpage

\label{app:model_subsample_pred_stability}
\subsection{Stability of predictors for different subsamples of models}

\begin{figure}[htbp]
    \centering
    \includegraphics[width=0.65\textwidth]{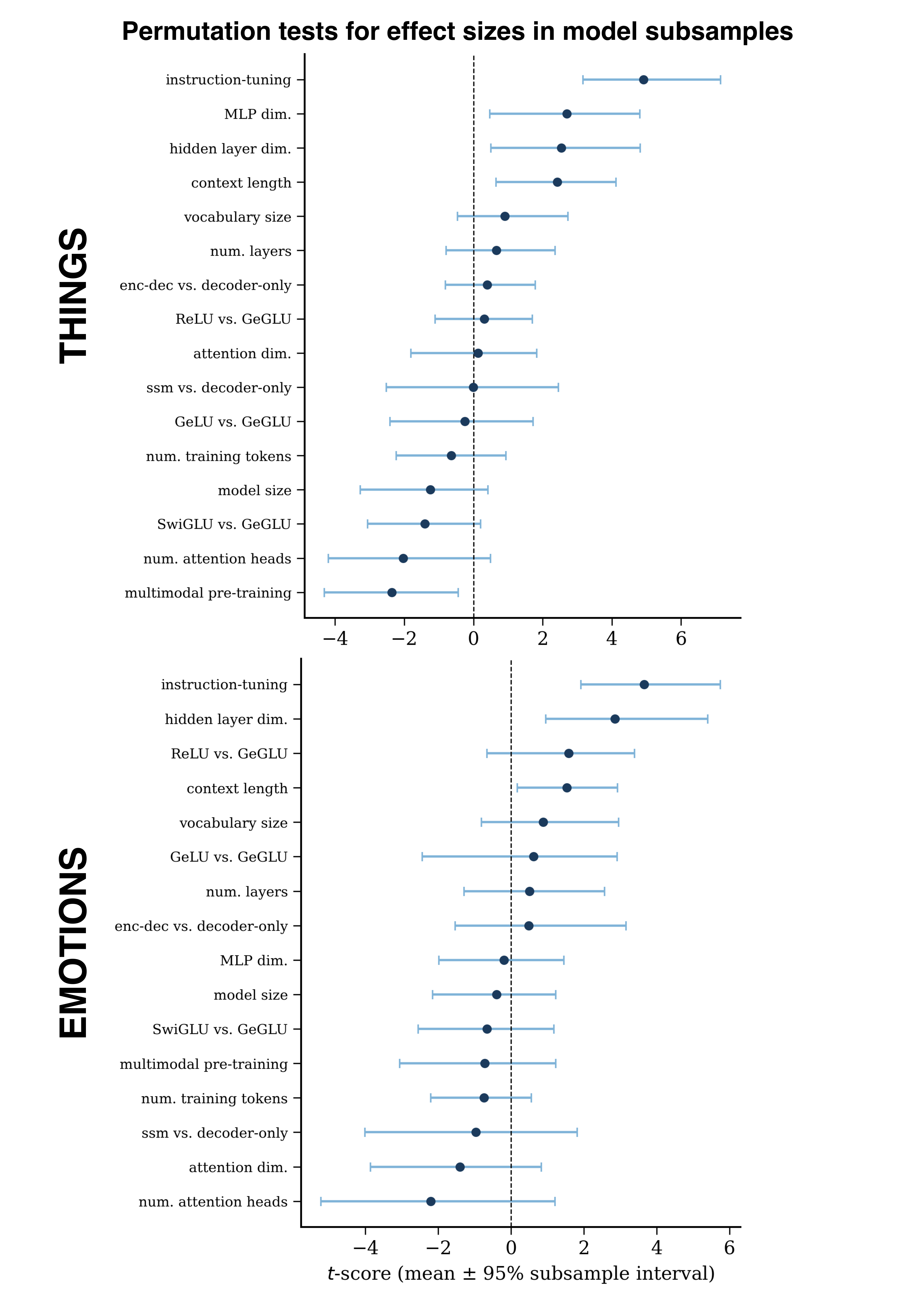}
    \caption{t-values estimated from permutation tests for THINGS and EMOTIONS data, with 95\% CIs.}
    \label{fig:permutation_model_test}
\end{figure}

Would we have observed different computational ingredients predictive of human-model alignment if we had tested a different sample of open source models? We conduct a permutation test to determine whether our observed effects were sensitive to the given models measured. We randomly selected subsamples of $n=50$ models across $k=10,000$ iterations, and then used the distributions of observed effects to obtain 95\% CIs for each predictor. We find that--even on smaller random subsamples of our models--the effects are commensurate with those we report in the original mixed effects models for \texttt{THINGS} and \texttt{EMOTIONS}. This indicates that significantly smaller subsamples of various open source models would converge to similar findings. Figure \ref{fig:permutation_model_test} shows t-values estimated from these permutation tests for \texttt{THINGS} and \texttt{EMOTIONS} data.

\clearpage

\clearpage
\subsection{$t$-SNE Visualizations}


\begin{table}[h!]
\centering
\caption{Complete ranking of models by R² score}
\label{tab:model_rankings_new}
\resizebox{\textwidth}{!}{%
\begin{tabular}{clc|clc}
\hline
Rank & Model Name & R² Score & Rank & Model Name & R² Score \\
\hline
1 & qwen25-14b-instruct & 0.492 & 39 & qwen2-7b & 0.294 \\
2 & Phi-3-medium-4k-instruct & 0.469 & 40 & Instella-3B-SFT & 0.293 \\
3 & Phi-3.5-MoE-instruct & 0.468 & 41 & Falcon3-3B-Base & 0.286 \\
4 & gemma-3-27b-it & 0.460 & 42 & flan-t5-xl & 0.265 \\
5 & qwen25-7b-instruct & 0.459 & 43 & falcon-11b & 0.257 \\
6 & Phi-3.5-mini-instruct & 0.457 & 44 & OLMo-2-1124-7B-Instruct & 0.241 \\
7 & phi-3-medium-128k-instruct & 0.450 & 45 & Instella-3B & 0.241 \\
8 & qwen2-7b-instruct & 0.446 & 46 & Falcon3-Mamba-7B-Base & 0.224 \\
9 & gemma-2-9b-it & 0.440 & 47 & flan-t5-large & 0.199 \\
10 & gemma-3-4b-it & 0.438 & 48 & Qwen3-4B-Base & 0.194 \\
11 & gemma-2-27b-it & 0.433 & 49 & phi-1\_5 & 0.180 \\
12 & gemma-3-12b-it & 0.431 & 50 & OLMo-2-1124-7B & 0.179 \\
13 & Llama-3.1-8B-Instruct & 0.424 & 51 & gemma-2-9b & 0.178 \\
14 & OLMo-2-0325-32B-Instruct & 0.423 & 52 & falcon-7b & 0.175 \\
15 & qwen-14b-chat & 0.418 & 53 & OLMo-2-0425-1B-SFT & 0.168 \\
16 & OLMo-2-1124-13B-Instruct & 0.414 & 54 & llama2-7b-chat & 0.160 \\
17 & phi-3-mini-128k-instruct & 0.412 & 55 & gemma-3-4b-pt & 0.147 \\
18 & OLMo-2-1124-13B-DPO & 0.406 & 56 & gemma-3-27b-pt & 0.143 \\
19 & Falcon3-10B-Base & 0.406 & 57 & gemma-3-12b-pt & 0.127 \\
20 & Phi-3.5-vision-instruct & 0.400 & 58 & gemma-2-2b & 0.126 \\
21 & OLMo-2-1124-13B-SFT & 0.392 & 59 & OLMo-2-0425-1B & 0.123 \\
22 & qwen3-8b & 0.388 & 60 & llama2-7b & 0.122 \\
23 & Instella-3B-Instruct & 0.374 & 61 & AMD-OLMo-1B & 0.113 \\
24 & gemma-2-27b & 0.369 & 62 & flan-t5-small & 0.111 \\
25 & qwen25-14b & 0.363 & 63 & AMD-OLMo-1B-SFT-DPO & 0.108 \\
26 & Llama-3.1-Tulu-3-8B & 0.358 & 64 & Instella-3B-Stage1 & 0.106 \\
27 & qwen-14b & 0.357 & 65 & qwen25-7b & 0.104 \\
28 & OLMo-2-1124-7B-DPO & 0.357 & 66 & falcon-7b-instruct & 0.103 \\
29 & Llama-3.1-Tulu-3-8B-DPO & 0.352 & 67 & t5-3b & 0.103 \\
30 & OLMo-2-1124-7B-SFT & 0.347 & 68 & t5-11b & 0.095 \\
31 & orca-2-13b & 0.343 & 69 & gemma-3-270m & 0.092 \\
32 & qwen3-14b & 0.337 & 70 & Yi-9B & 0.091 \\
33 & gemma-2-2b-it & 0.321 & 71 & gpt-oss-20b & 0.087 \\
34 & orca-2-7b & 0.315 & 72 & gemma-3-270m-it & 0.085 \\
35 & flan-t5-xxl & 0.302 & 73 & mamba-1.4b-hf & 0.083 \\
36 & Llama-3.1-Tulu-3-8B-SFT & 0.301 & 74 & mamba-2.8b-hf & 0.077 \\
37 & llama2-13b-chat & 0.299 & 75 & gpt-j-6b & 0.025 \\
38 & OLMo-2-1124-13B & 0.299 & & & \\
\hline
\end{tabular}%
}
\end{table}

\begin{figure}[htbp]
    \centering
    \caption{t-SNE visualizations of model embeddings (Part 1). R² scores are in parentheses.}
    \label{fig:tsne_models_part1}
    \begin{tabular}{@{}c@{\hspace{2mm}}c@{\hspace{2mm}}c@{}}
        \includegraphics[width=0.31\textwidth]{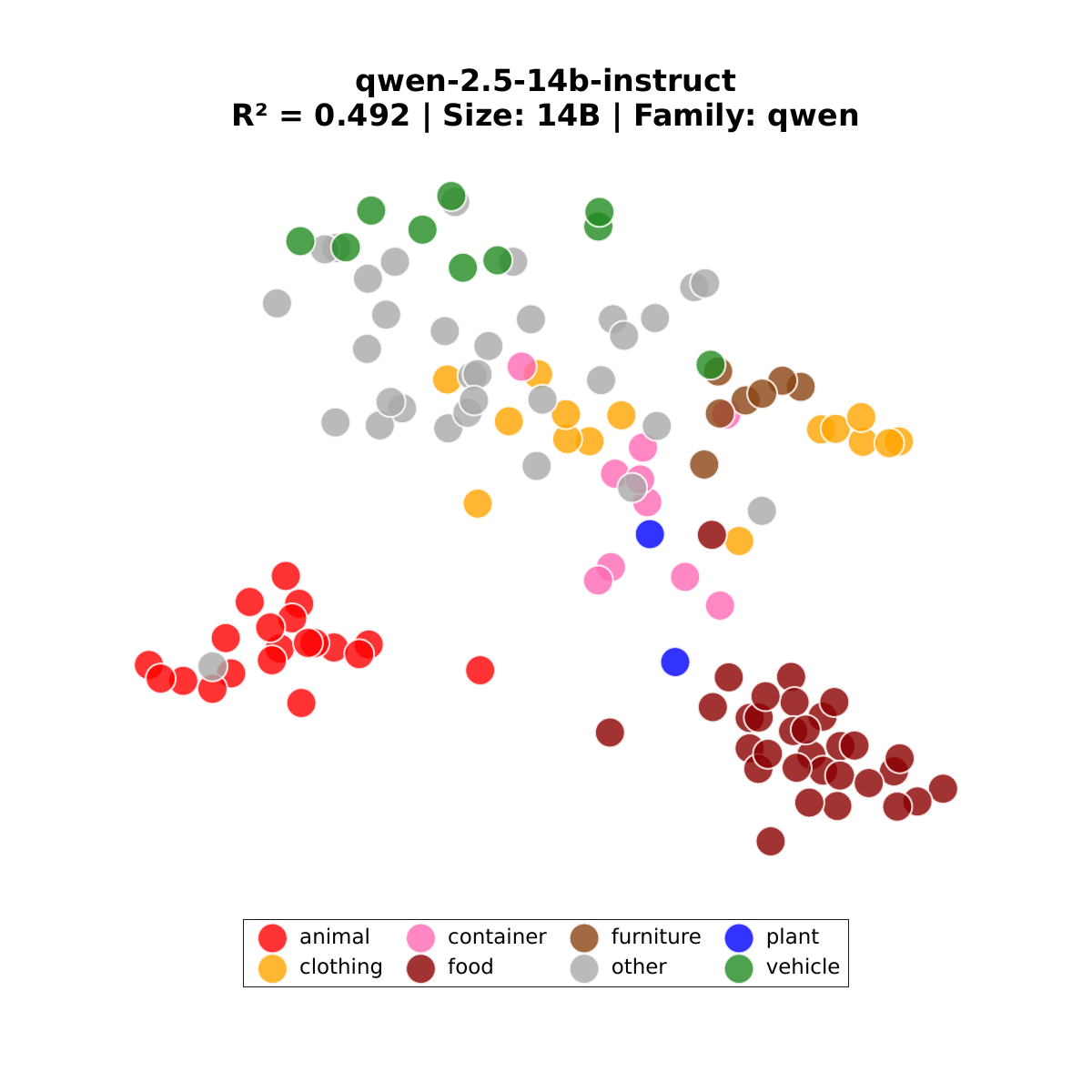} &
        \includegraphics[width=0.31\textwidth]{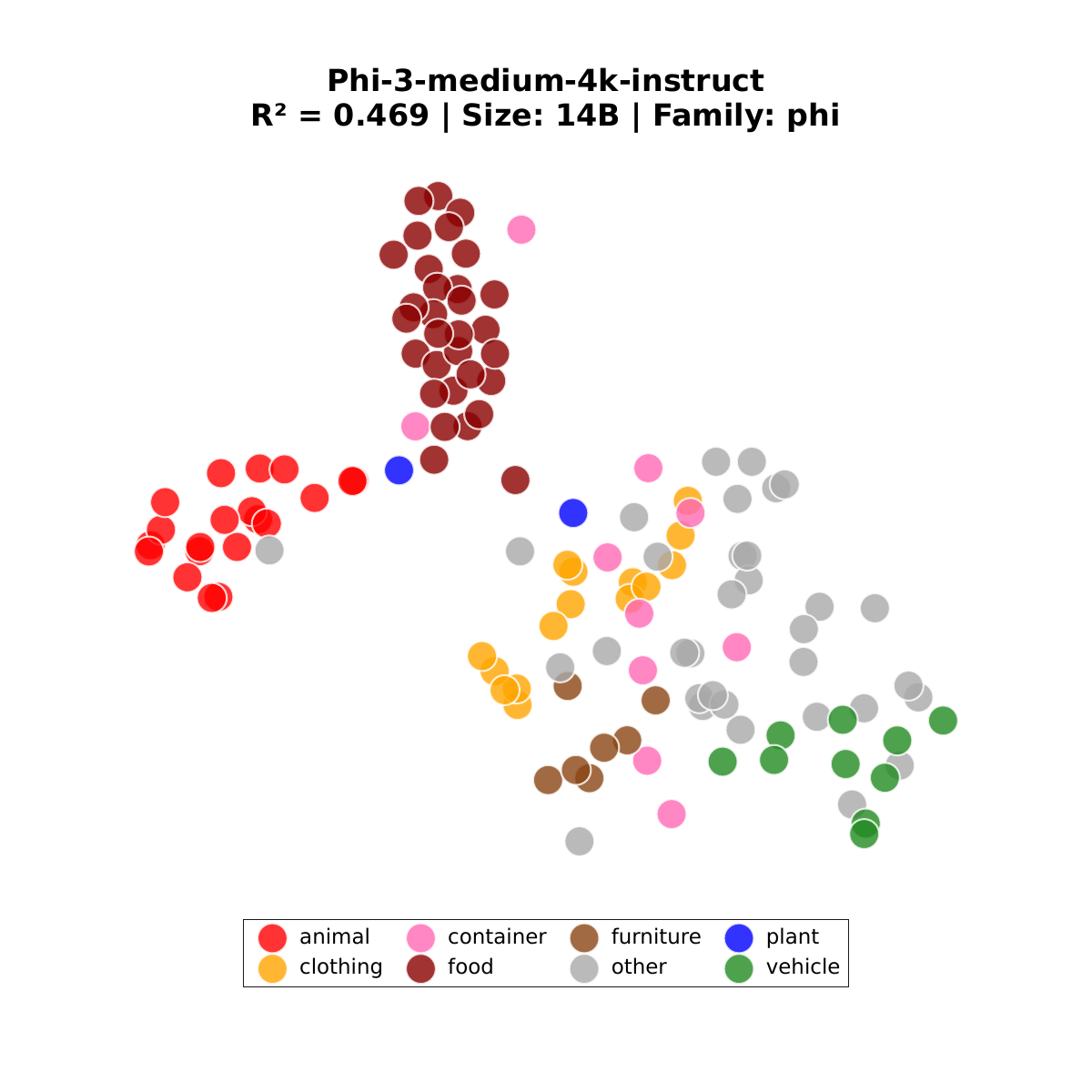} &
        \includegraphics[width=0.31\textwidth]{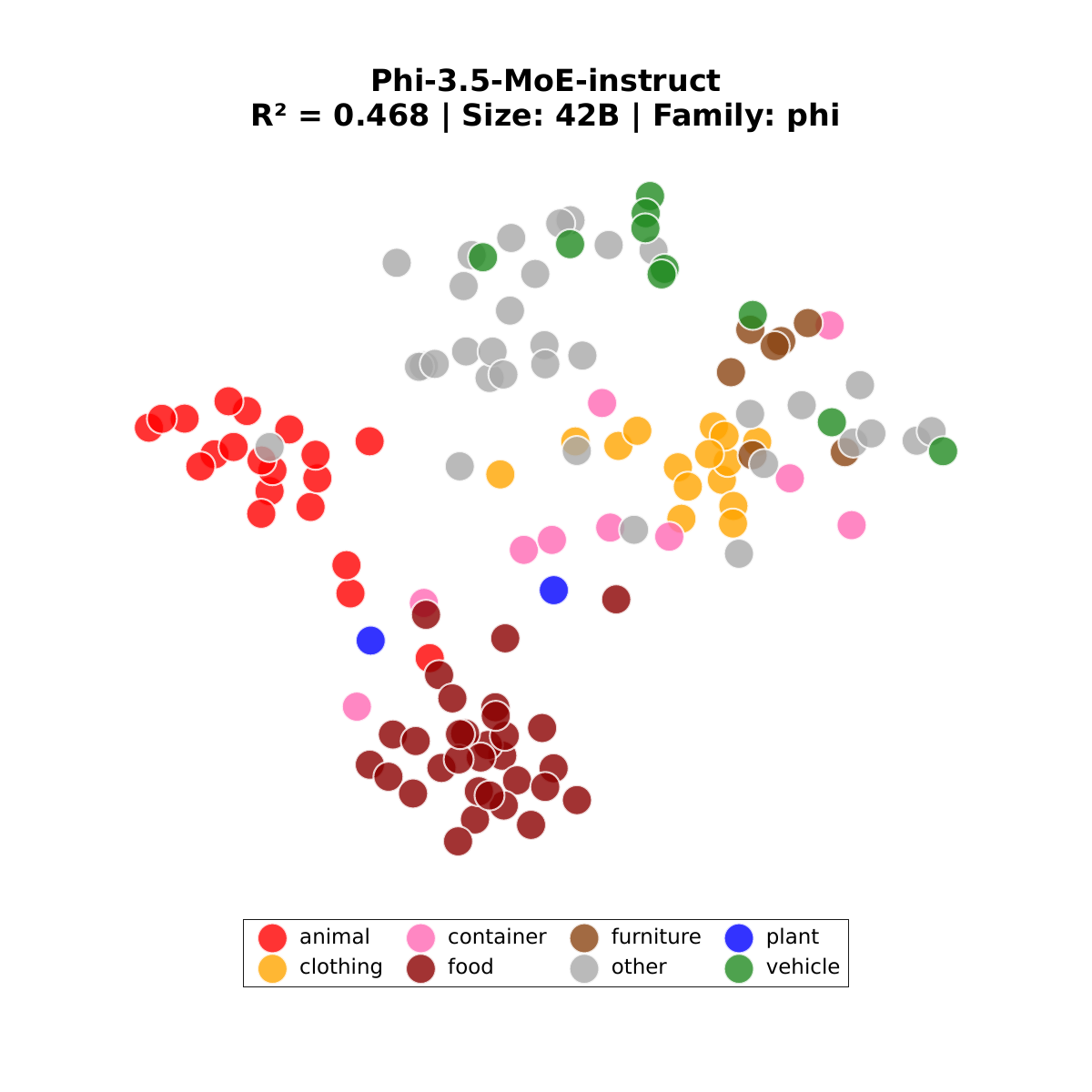} \\
        {\small (1) Qwen2.5-14B-Inst (0.492)} & {\small (2) Phi-3-Med-4k (0.469)} & {\small (3) Phi-3.5-MoE (0.468)} \\[3mm]
        
        \includegraphics[width=0.31\textwidth]{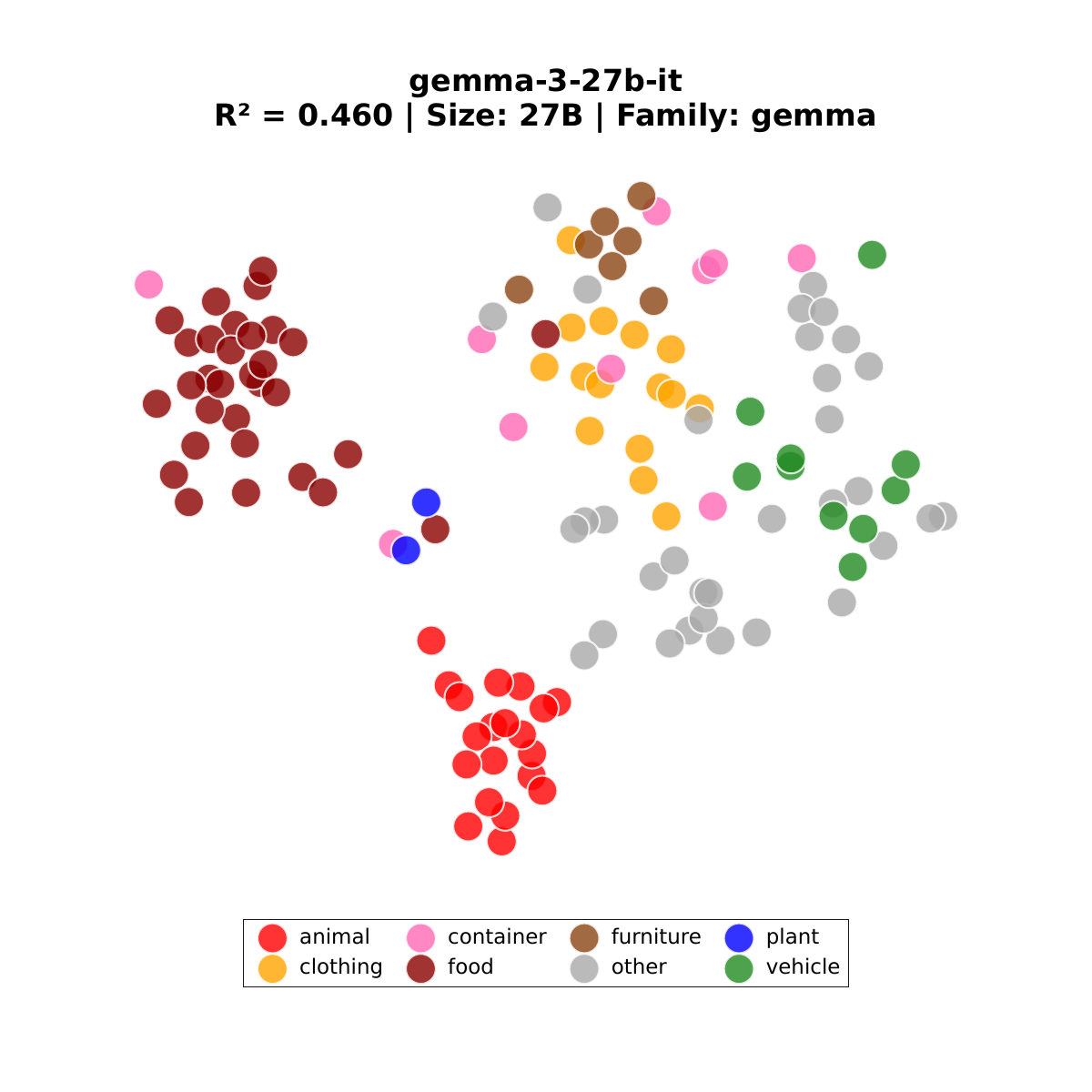} &
        \includegraphics[width=0.31\textwidth]{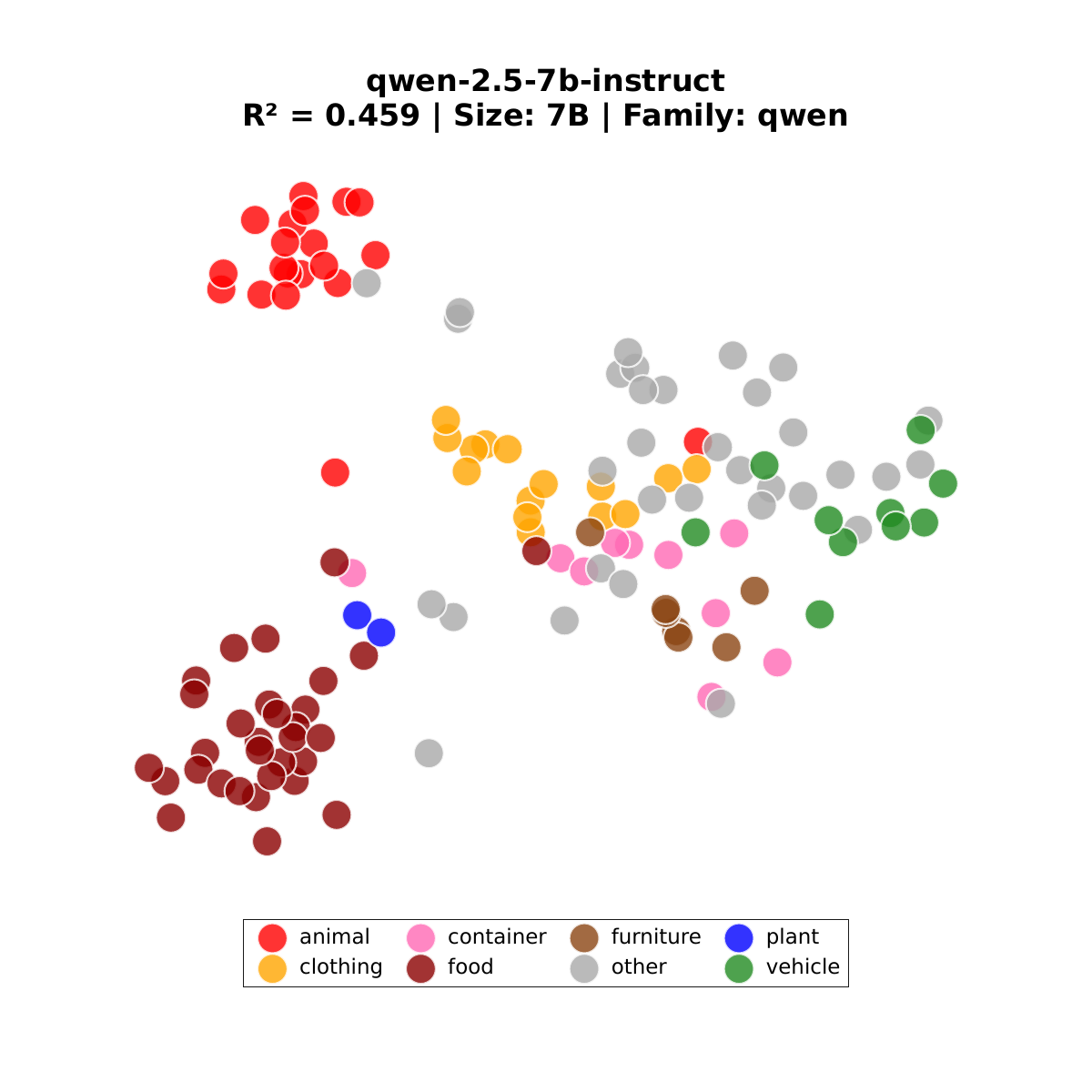} &
        \includegraphics[width=0.31\textwidth]{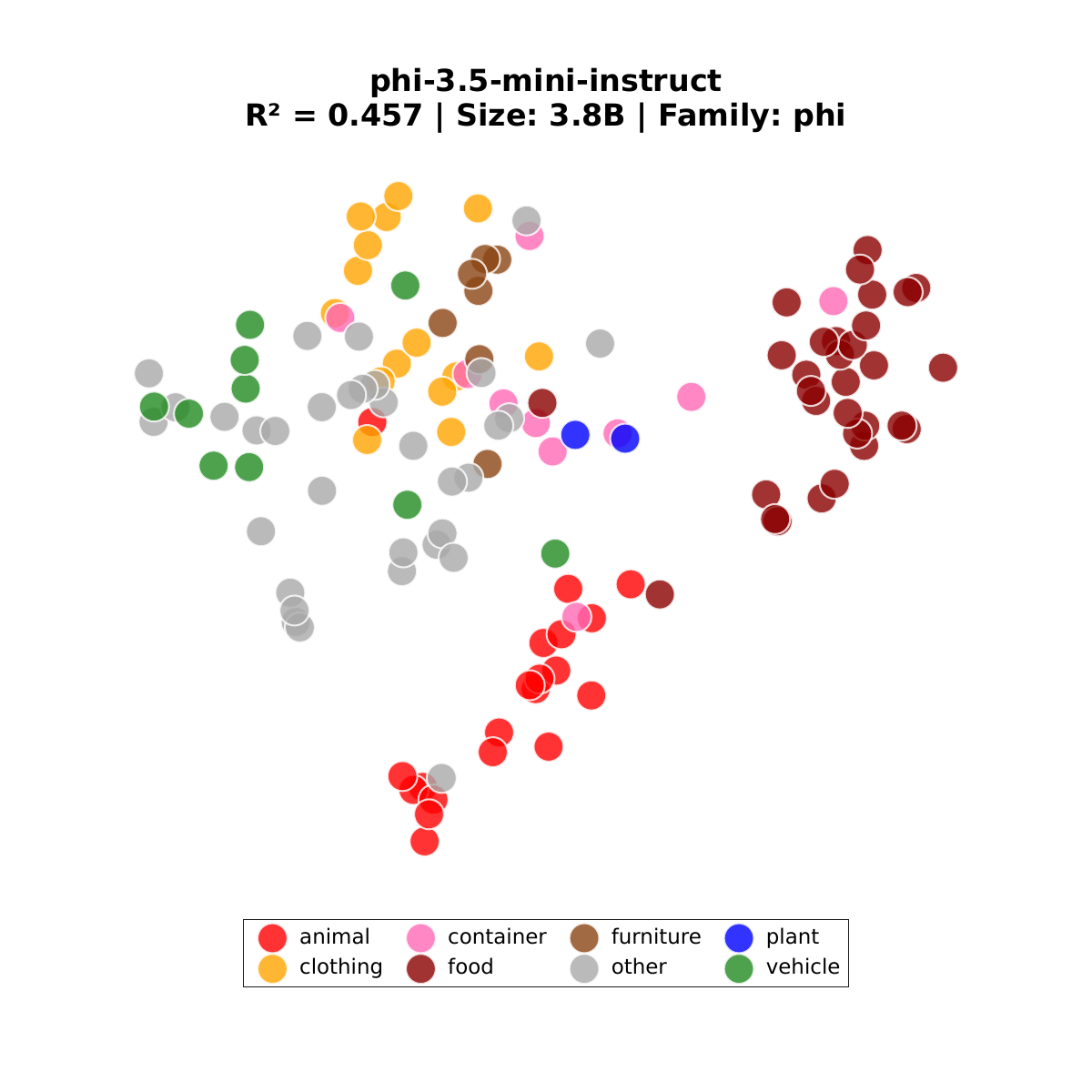} \\
        {\small (4) Gemma-3-27B-IT (0.460)} & {\small (5) Qwen2.5-7B-Inst (0.459)} & {\small (6) Phi-3.5-Mini (0.457)} \\[3mm]
        
        \includegraphics[width=0.31\textwidth]{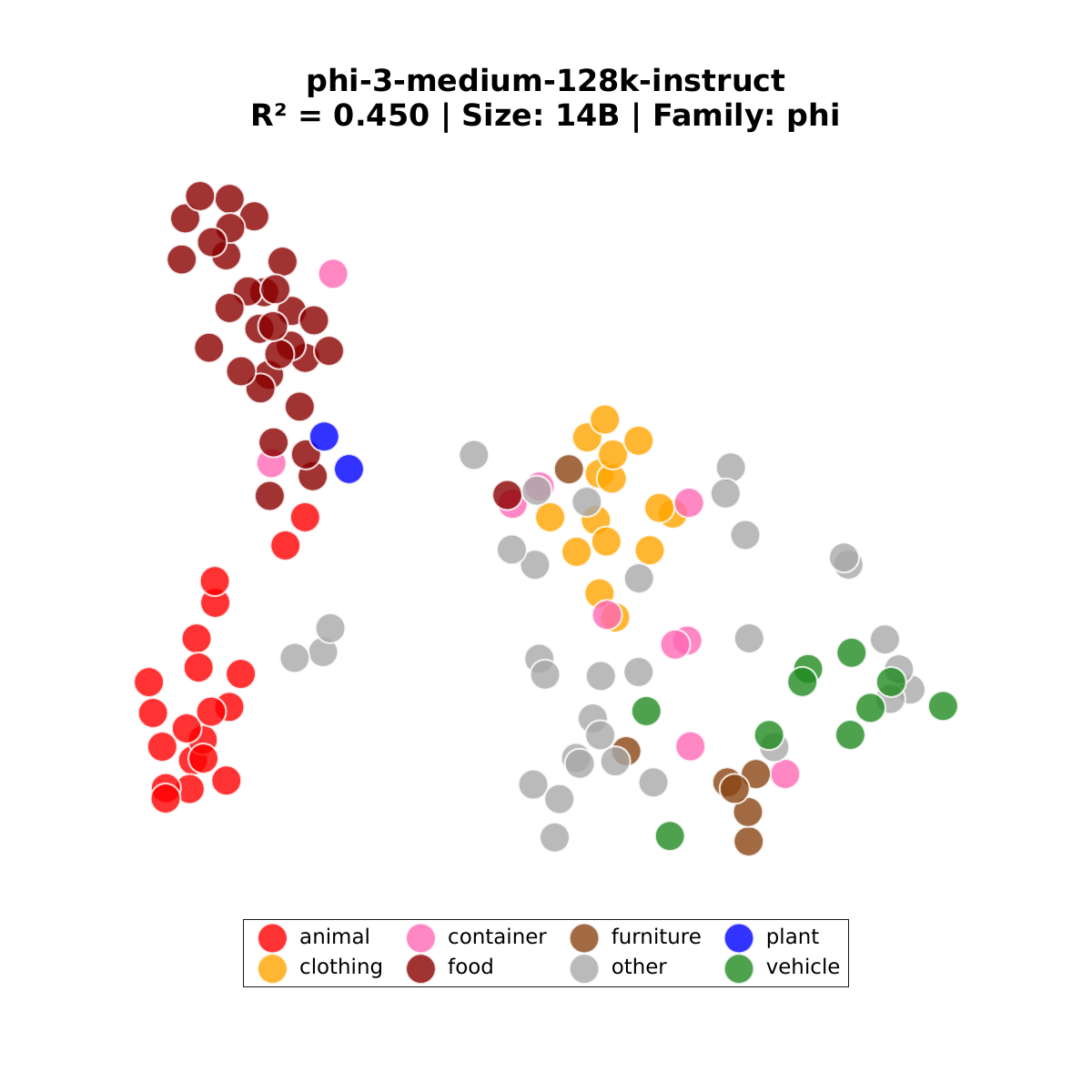} &
        \includegraphics[width=0.31\textwidth]{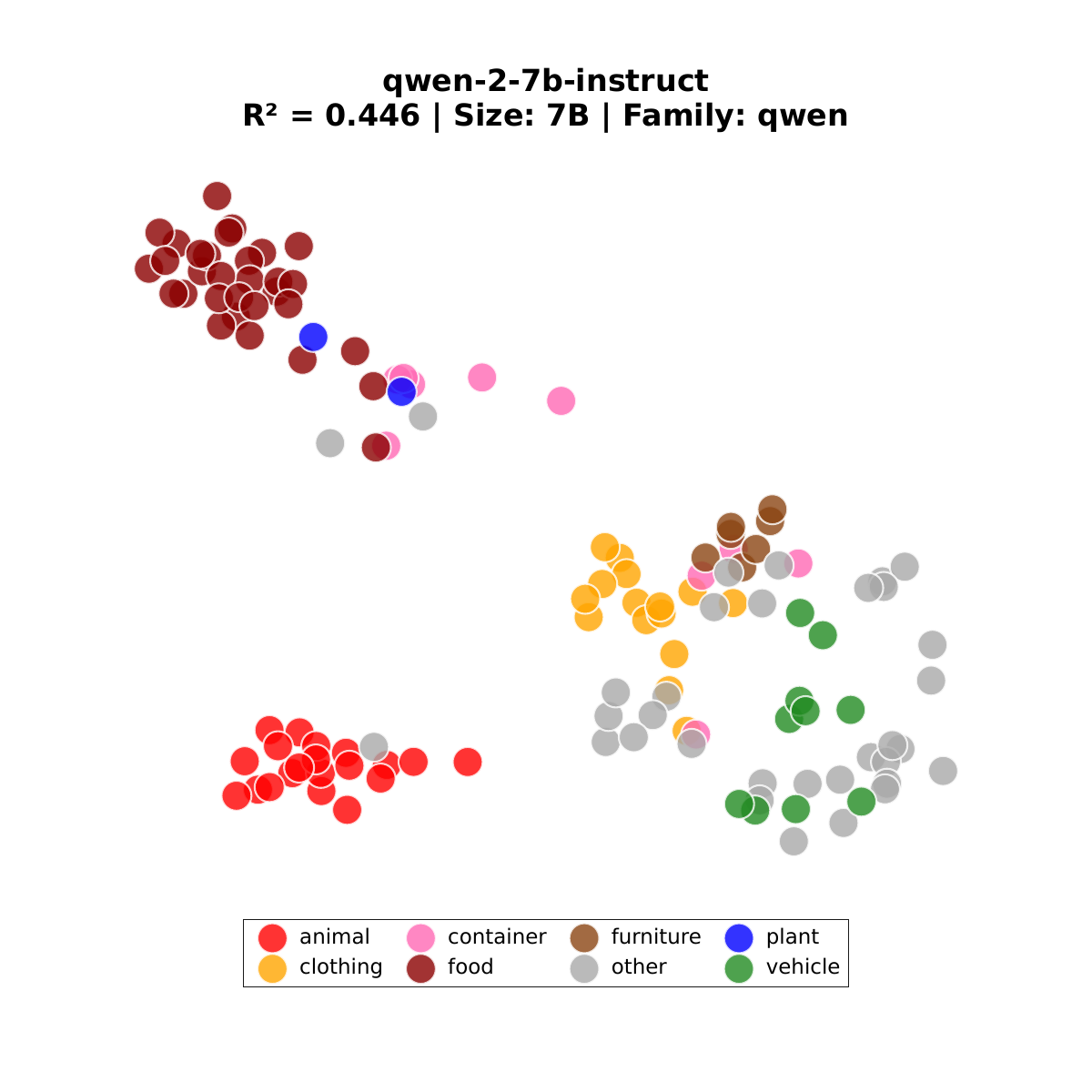} &
        \includegraphics[width=0.31\textwidth]{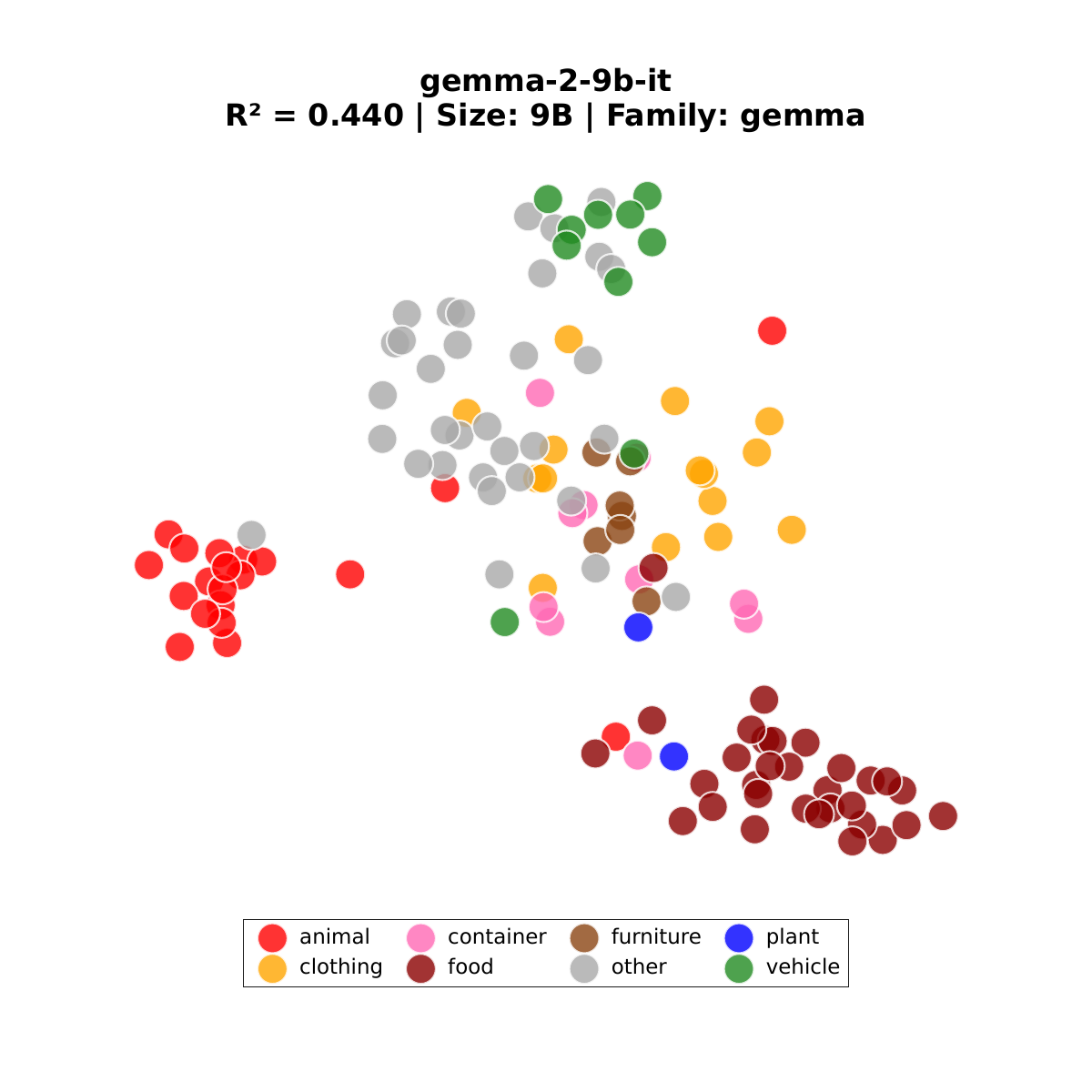} \\
        {\small (7) Phi-3-Med-128k (0.450)} & {\small (8) Qwen2-7B-Inst (0.446)} & {\small (9) Gemma-2-9B-IT (0.440)} \\[3mm]
        
        \includegraphics[width=0.31\textwidth]{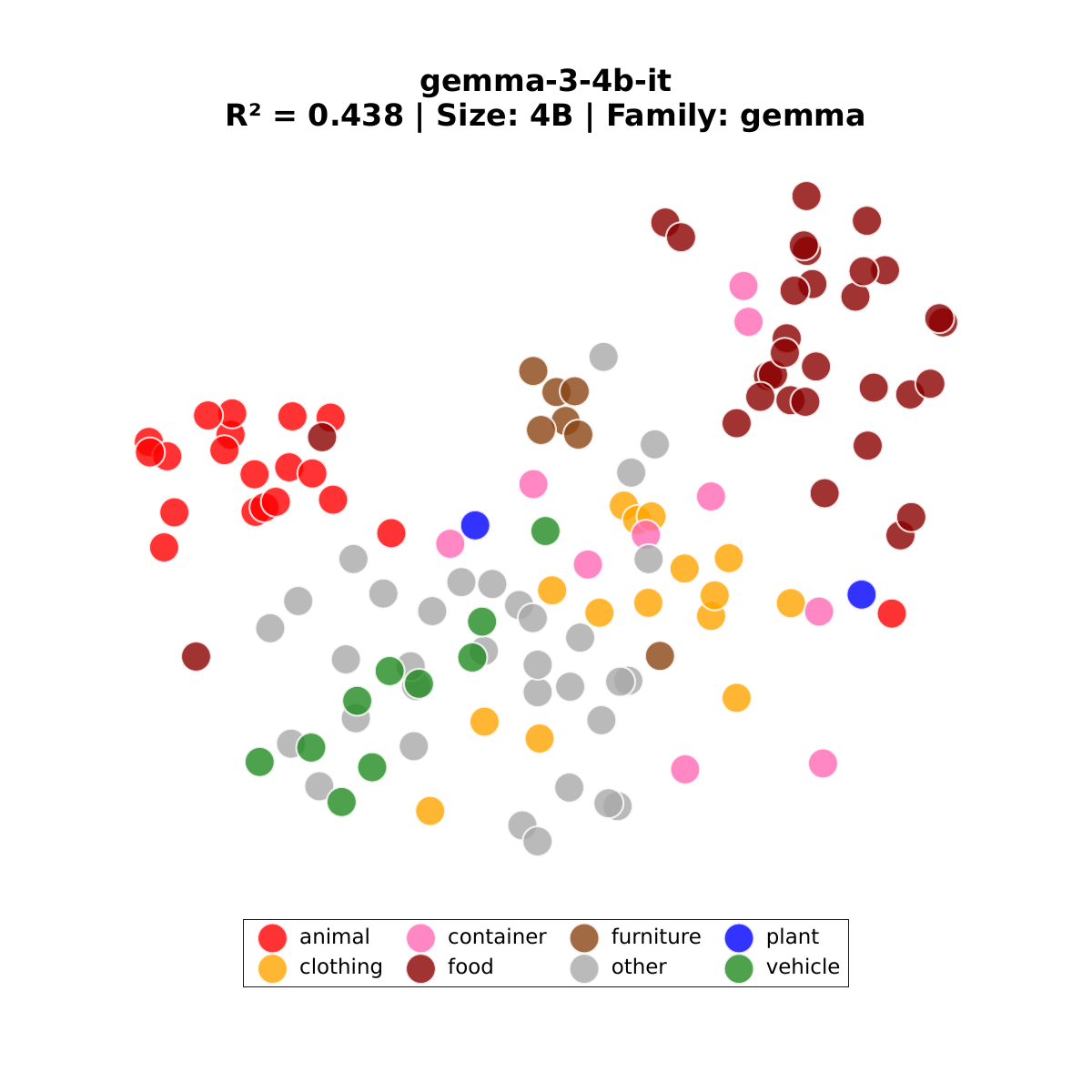} &
        \includegraphics[width=0.31\textwidth]{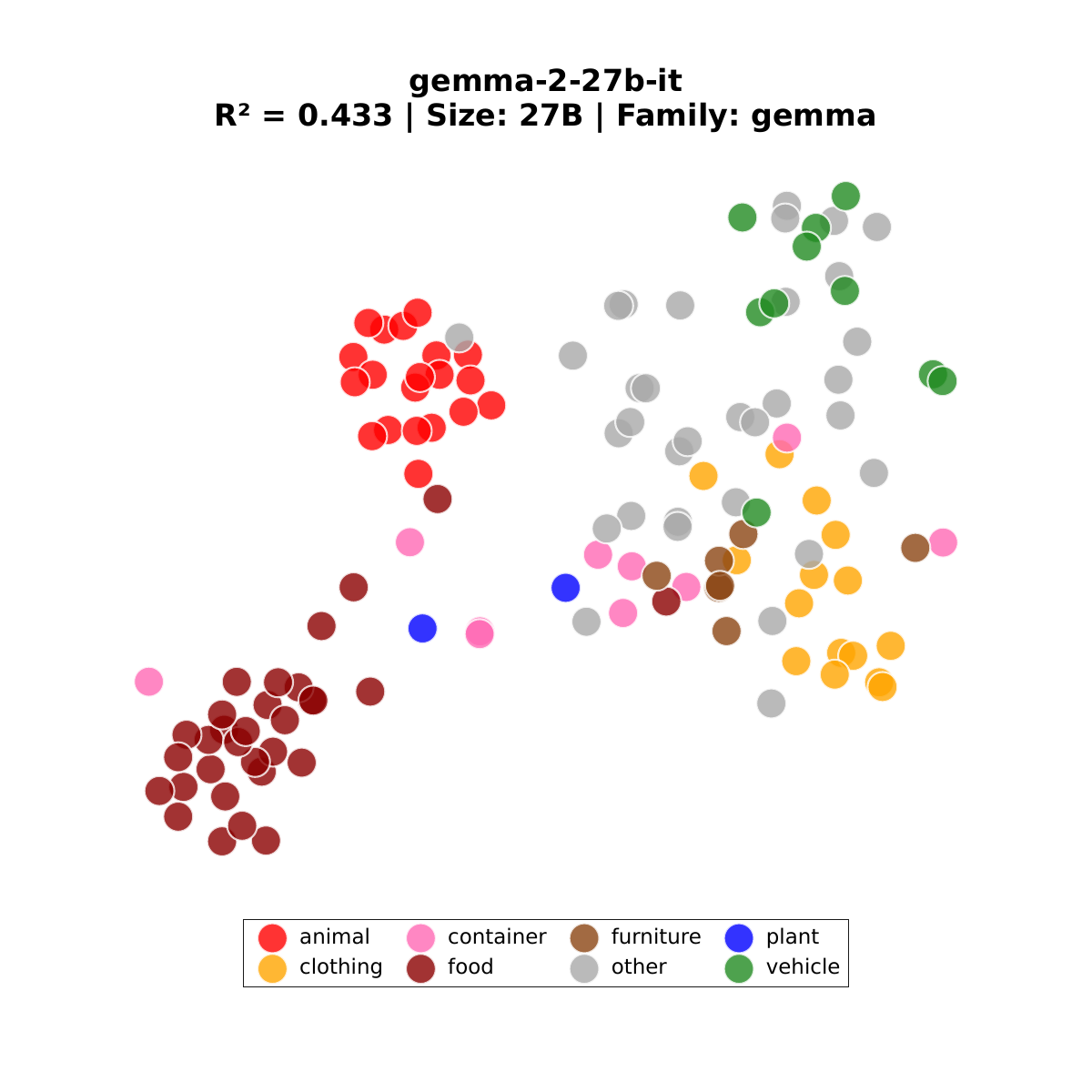} &
        \includegraphics[width=0.31\textwidth]{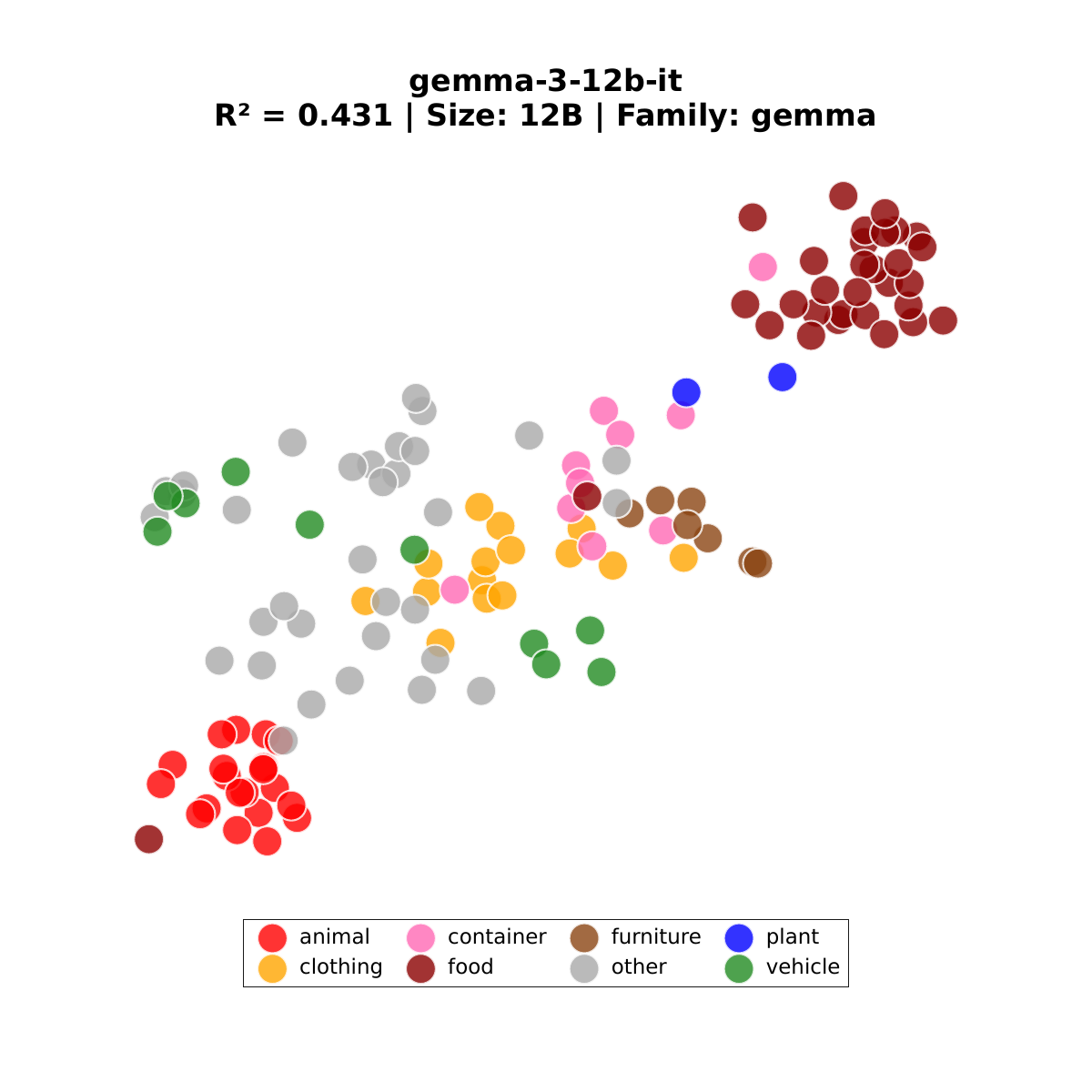} \\
        {\small (10) Gemma-3-4B-IT (0.438)} & {\small (11) Gemma-2-27B-IT (0.433)} & {\small (12) Gemma-3-12B-IT (0.431)} \\
    \end{tabular}
\end{figure}

\begin{figure}[htbp]
    \centering
    \caption{t-SNE visualizations of model embeddings (Part 2). R² scores are in parentheses.}
    \label{fig:tsne_models_part2}
    \begin{tabular}{@{}c@{\hspace{2mm}}c@{\hspace{2mm}}c@{}}
        \includegraphics[width=0.31\textwidth]{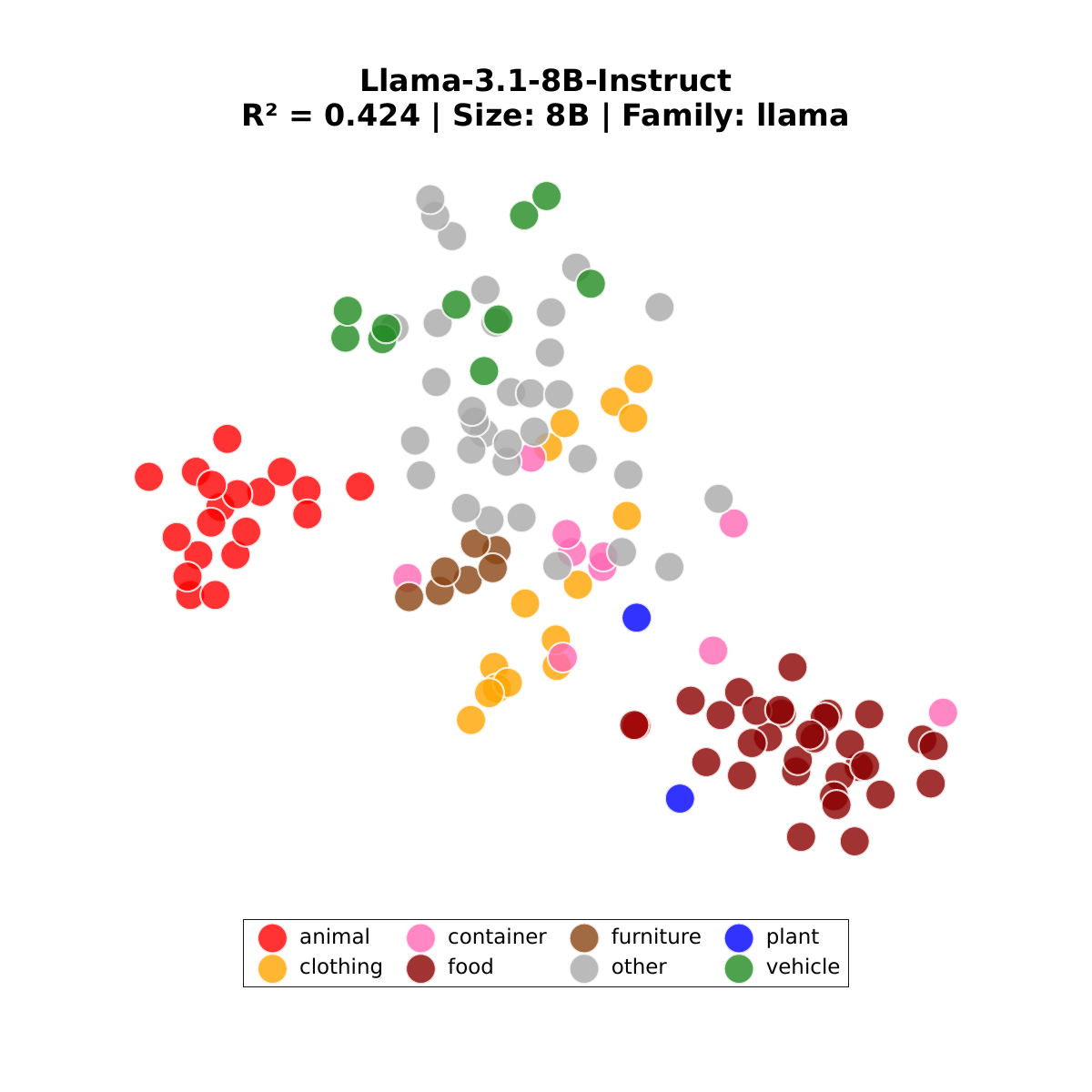} &
        \includegraphics[width=0.31\textwidth]{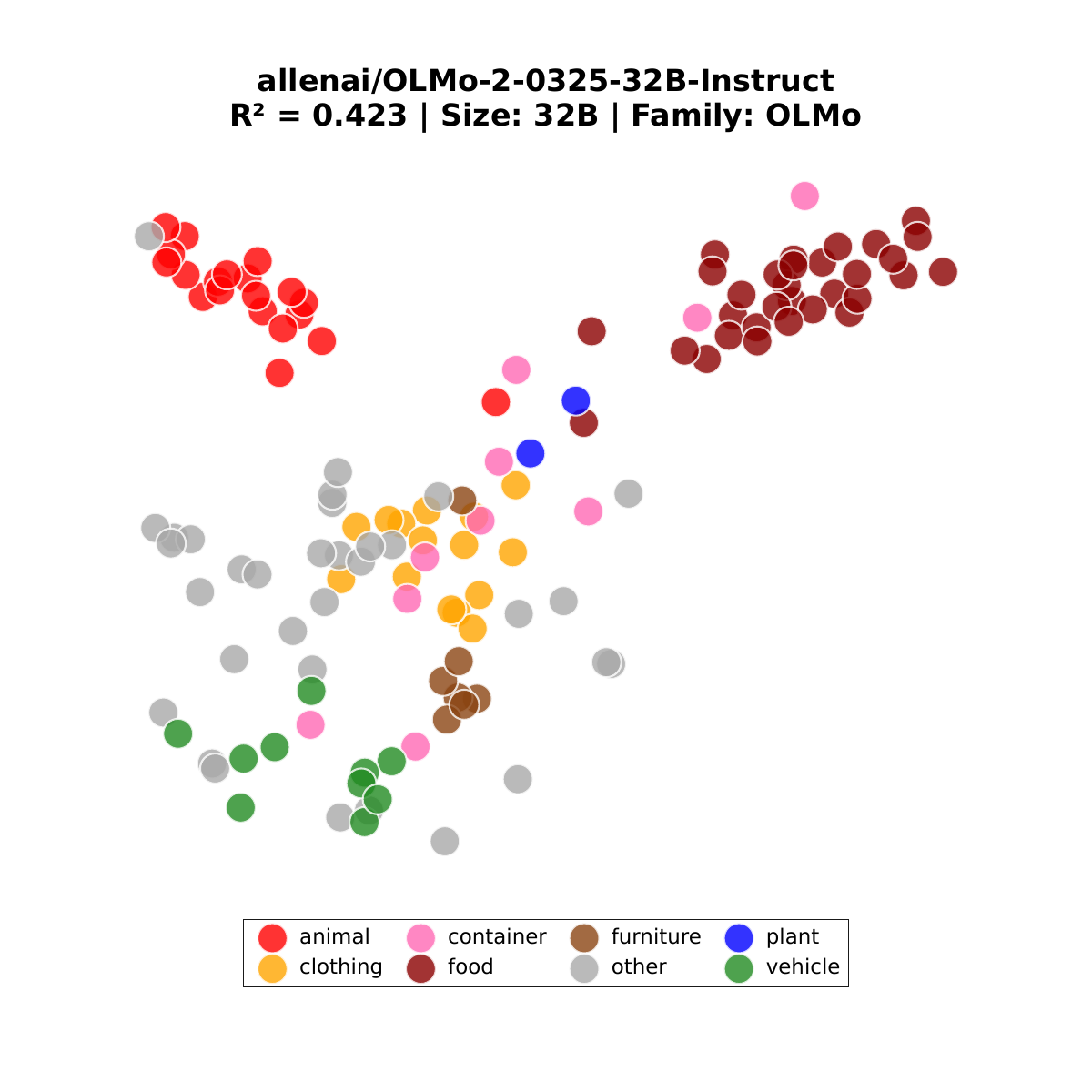} &
        \includegraphics[width=0.31\textwidth]{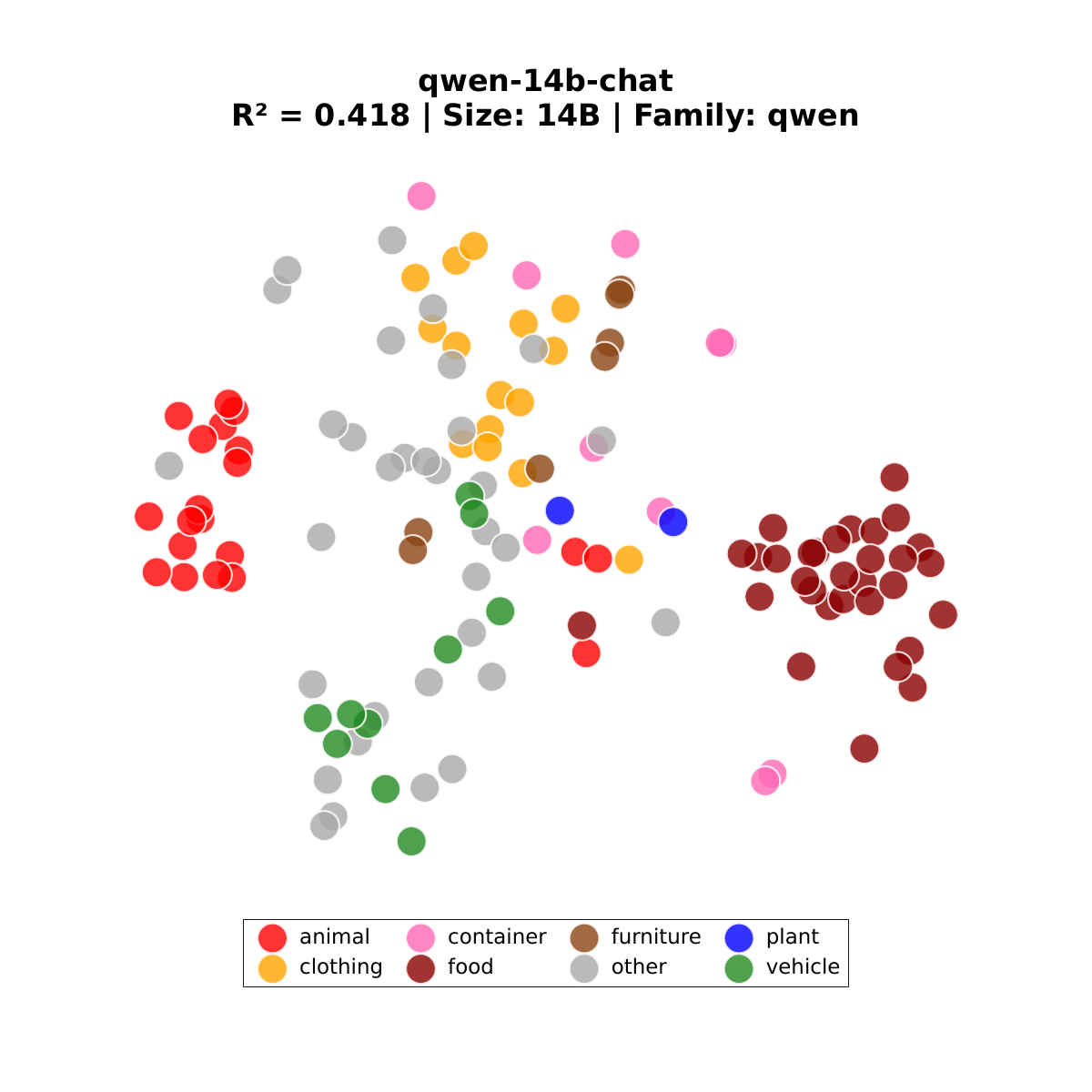} \\
        {\small (13) Llama-3.1-8B (0.424)} & {\small (14) OLMo-2-32B (0.423)} & {\small (15) Qwen-14B-Chat (0.418)} \\[3mm]
        
        \includegraphics[width=0.31\textwidth]{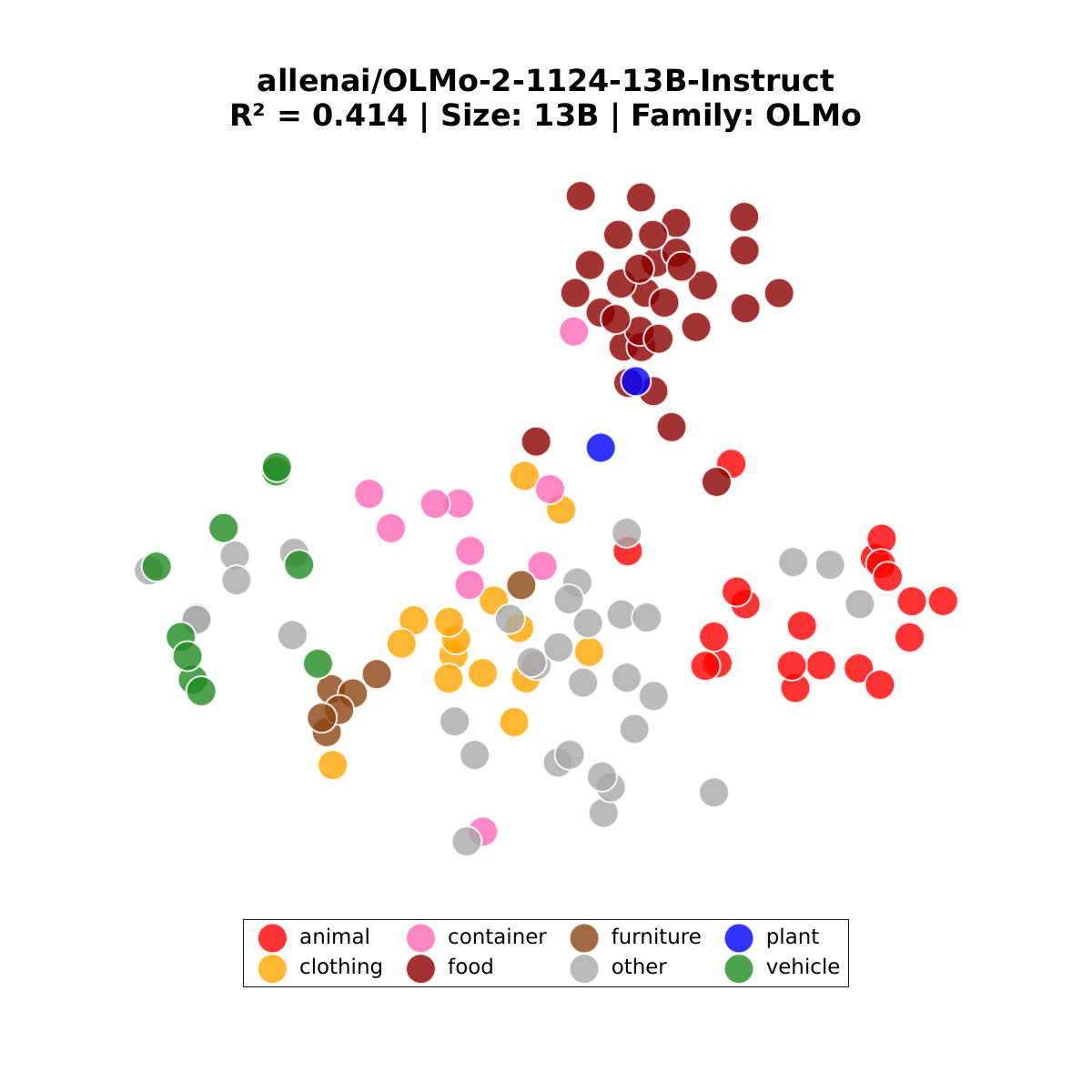} &
        \includegraphics[width=0.31\textwidth]{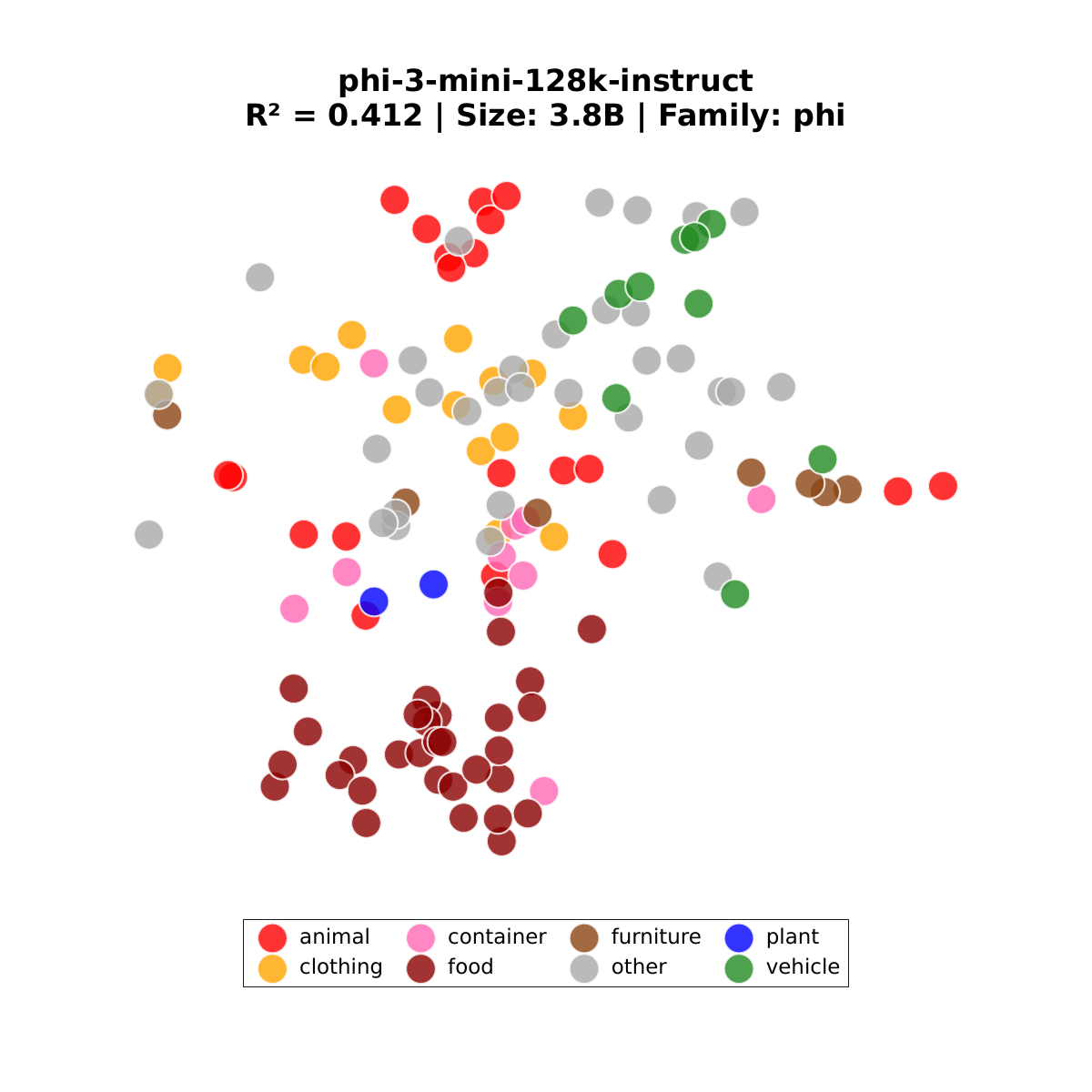} &
        \includegraphics[width=0.31\textwidth]{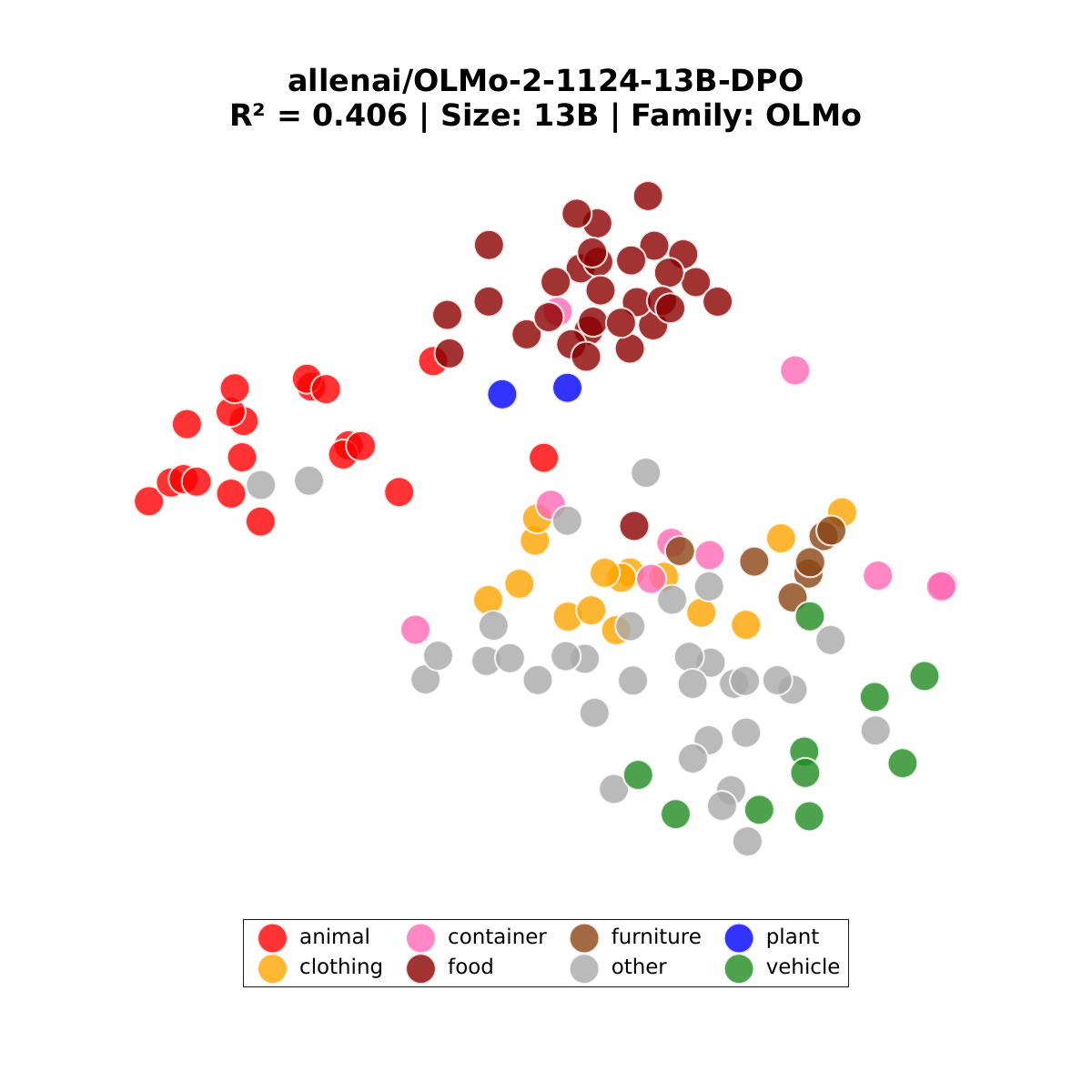} \\
        {\small (16) OLMo-2-13B-Inst (0.414)} & {\small (17) Phi-3-Mini-128k (0.412)} & {\small (18) OLMo-2-13B-DPO (0.406)} \\[3mm]
        
        \includegraphics[width=0.31\textwidth]{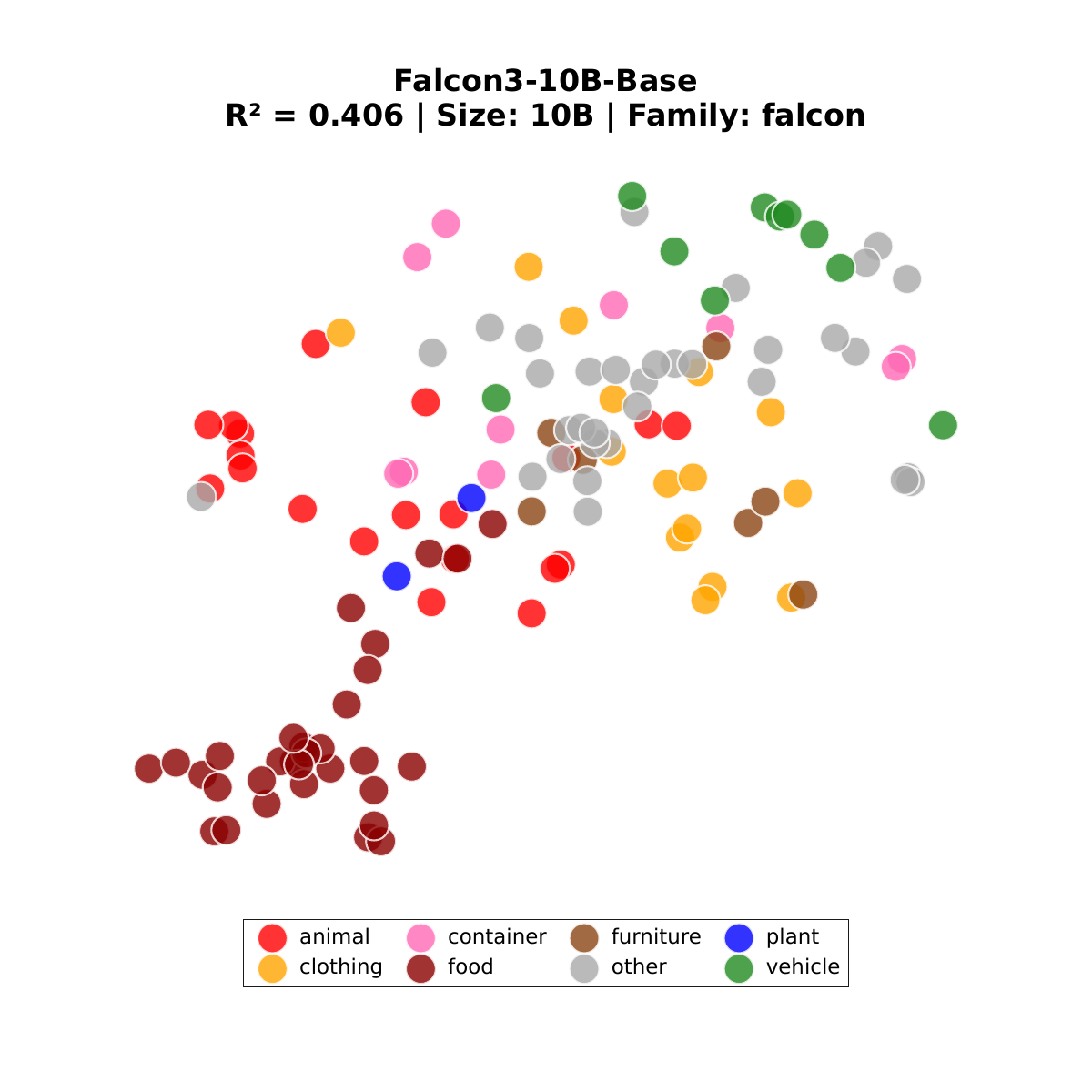} &
        \includegraphics[width=0.31\textwidth]{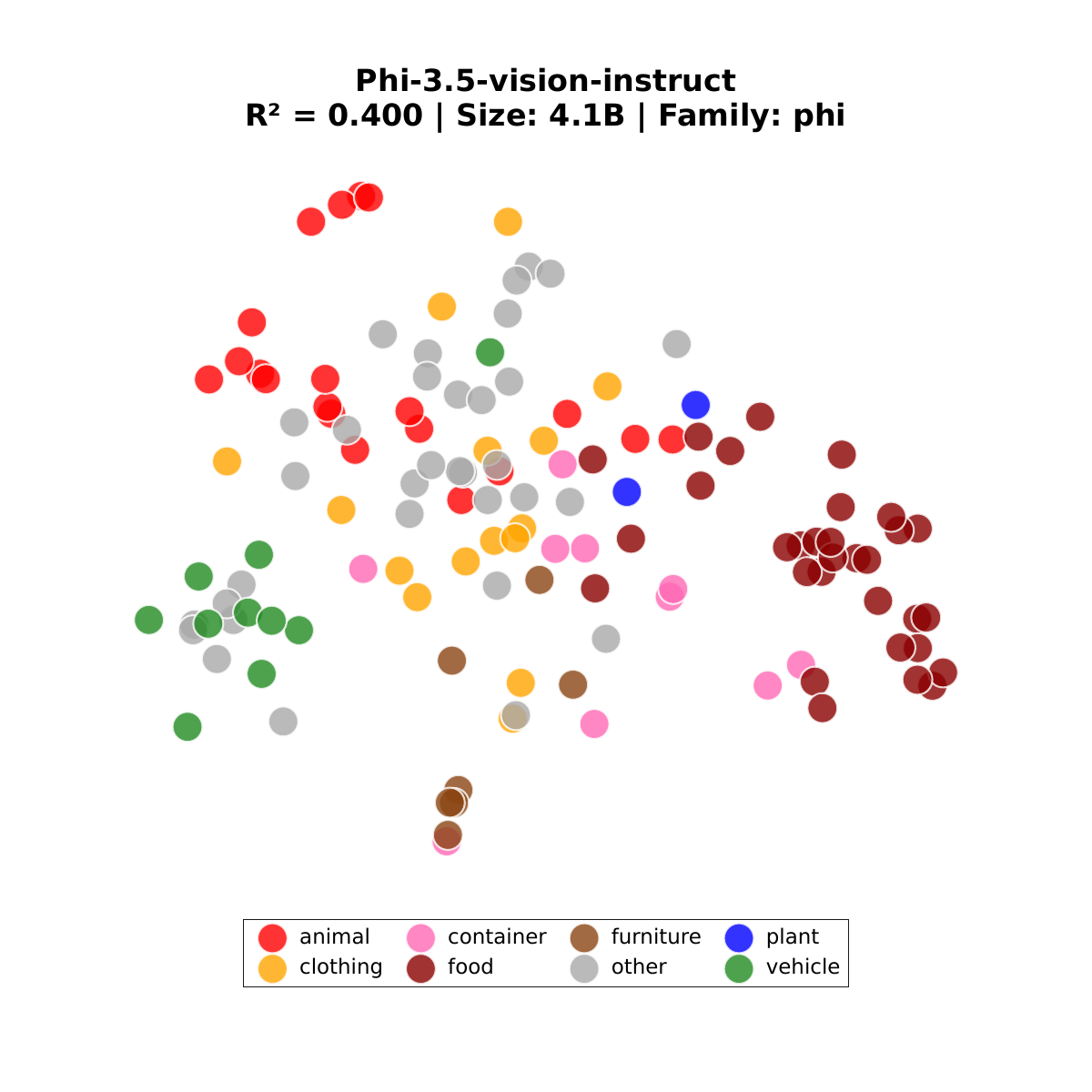} &
        \includegraphics[width=0.31\textwidth]{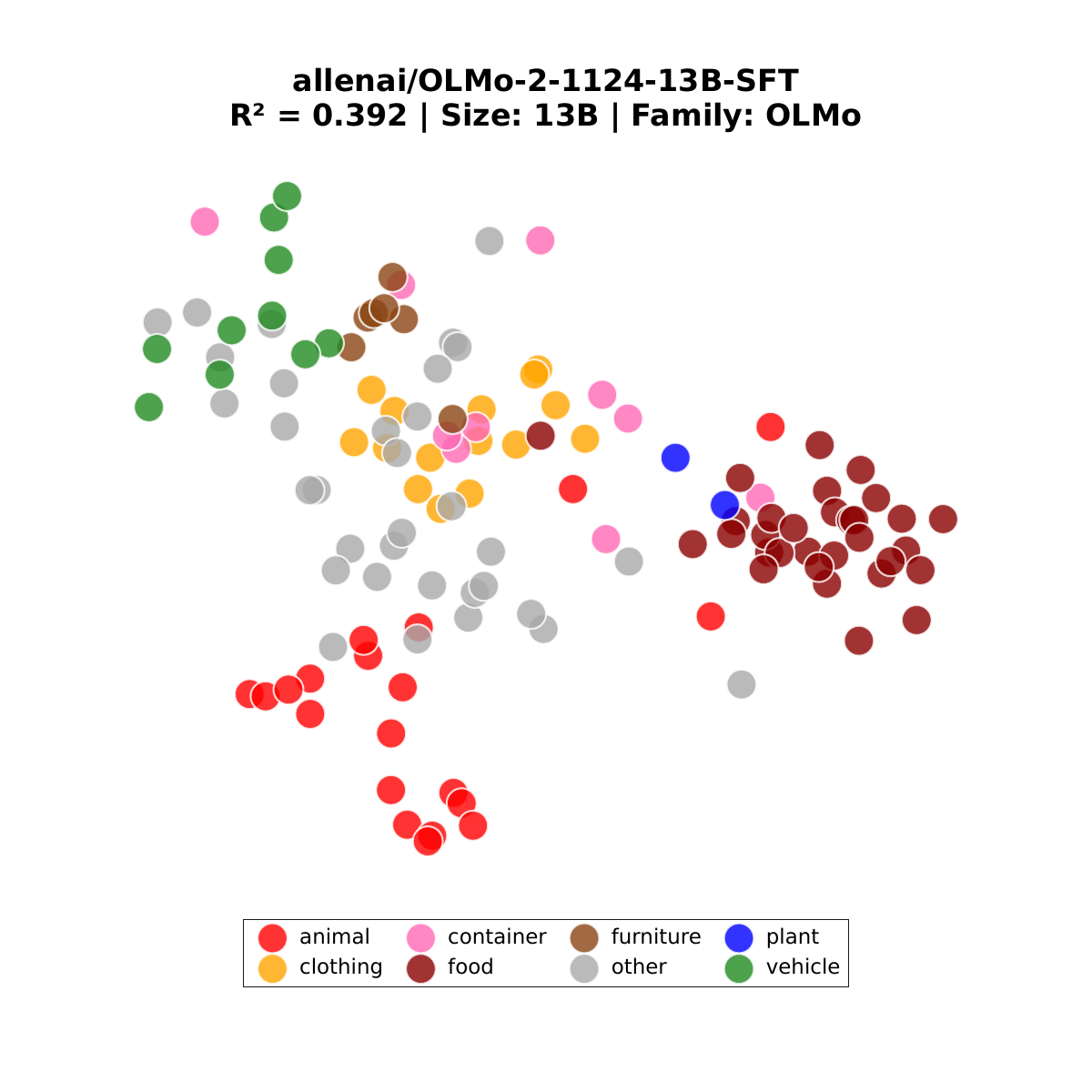} \\
        {\small (19) Falcon3-10B (0.406)} & {\small (20) Phi-3.5-Vision (0.400)} & {\small (21) OLMo-2-13B-SFT (0.392)} \\[3mm]
        
        \includegraphics[width=0.31\textwidth]{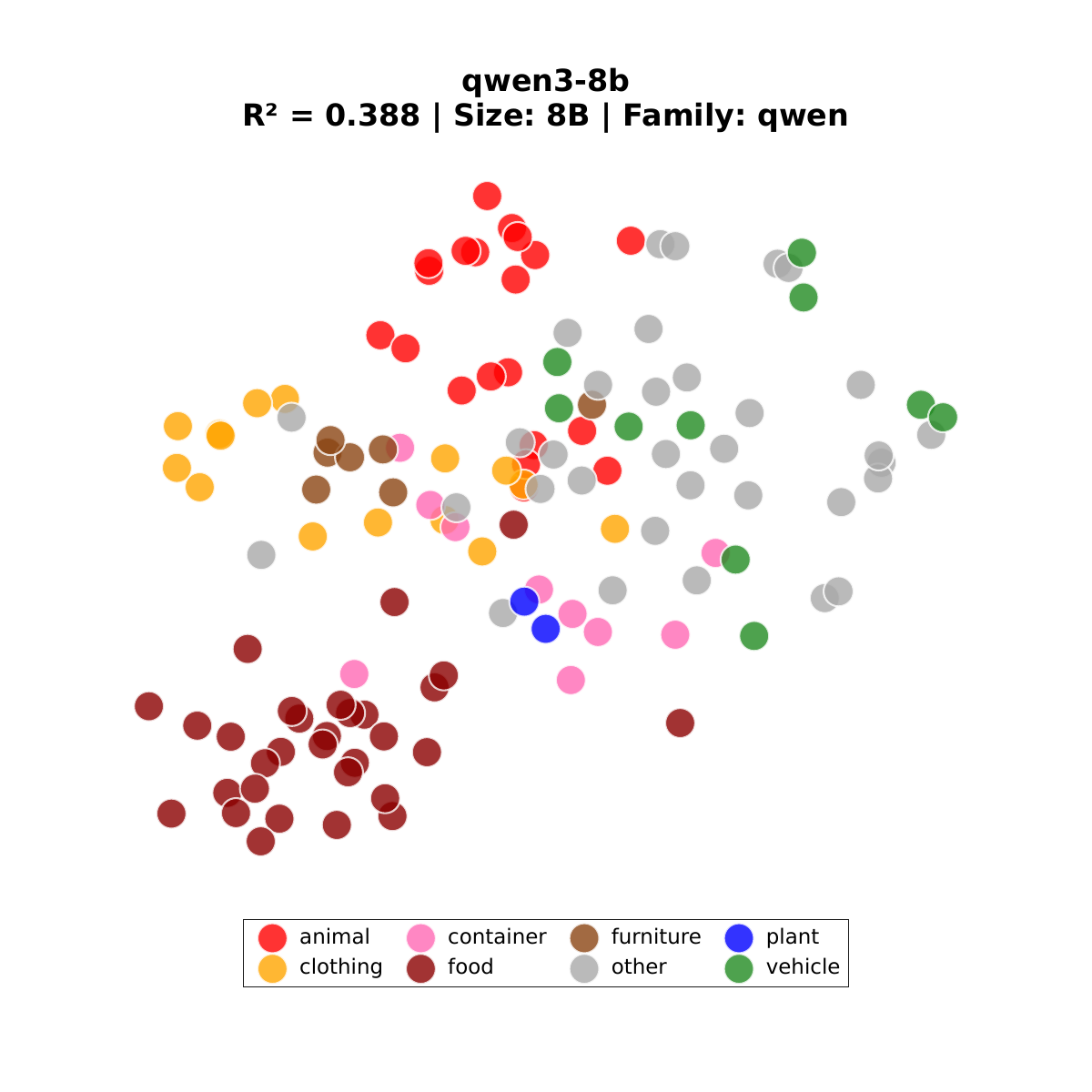} &
        \includegraphics[width=0.31\textwidth]{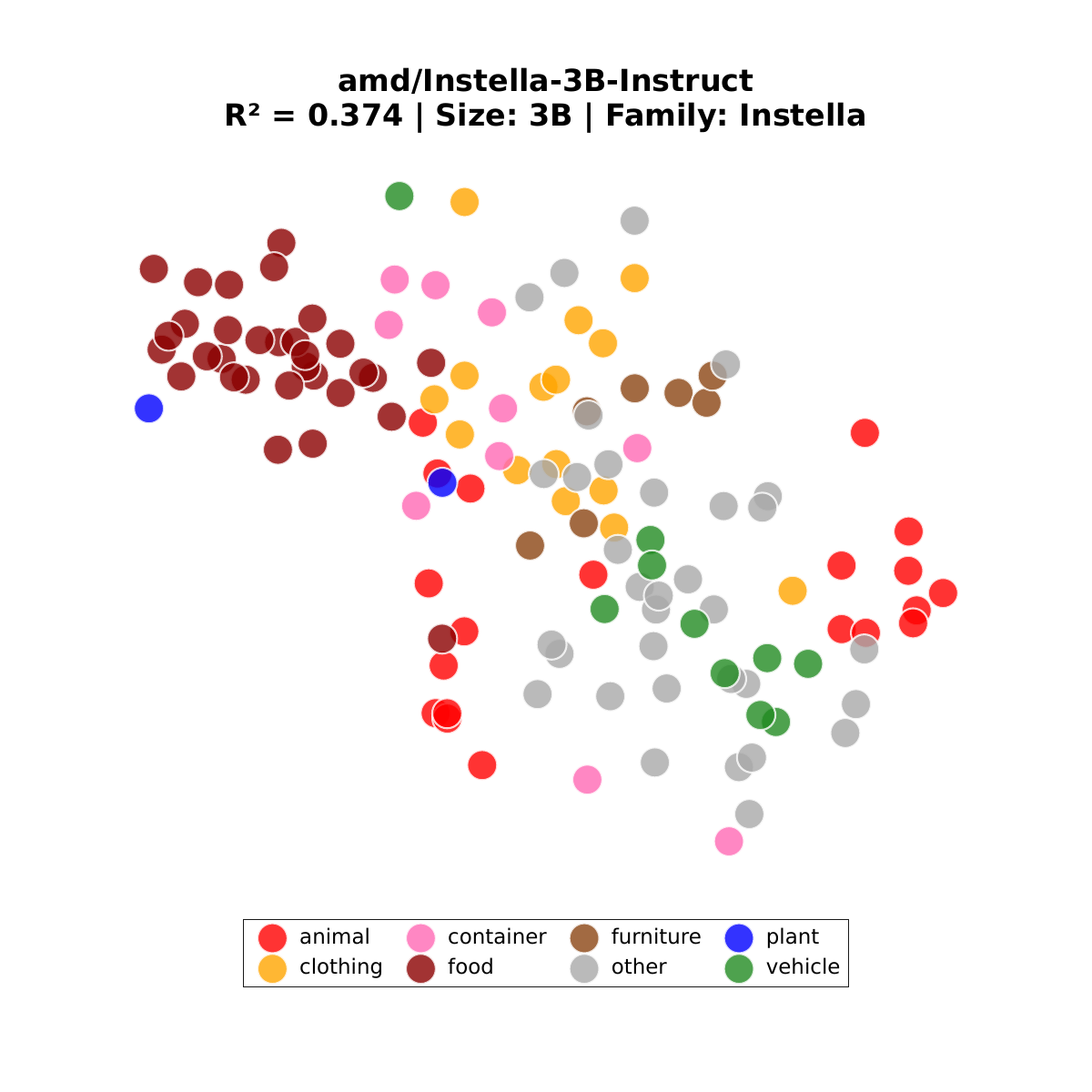} &
        \includegraphics[width=0.31\textwidth]{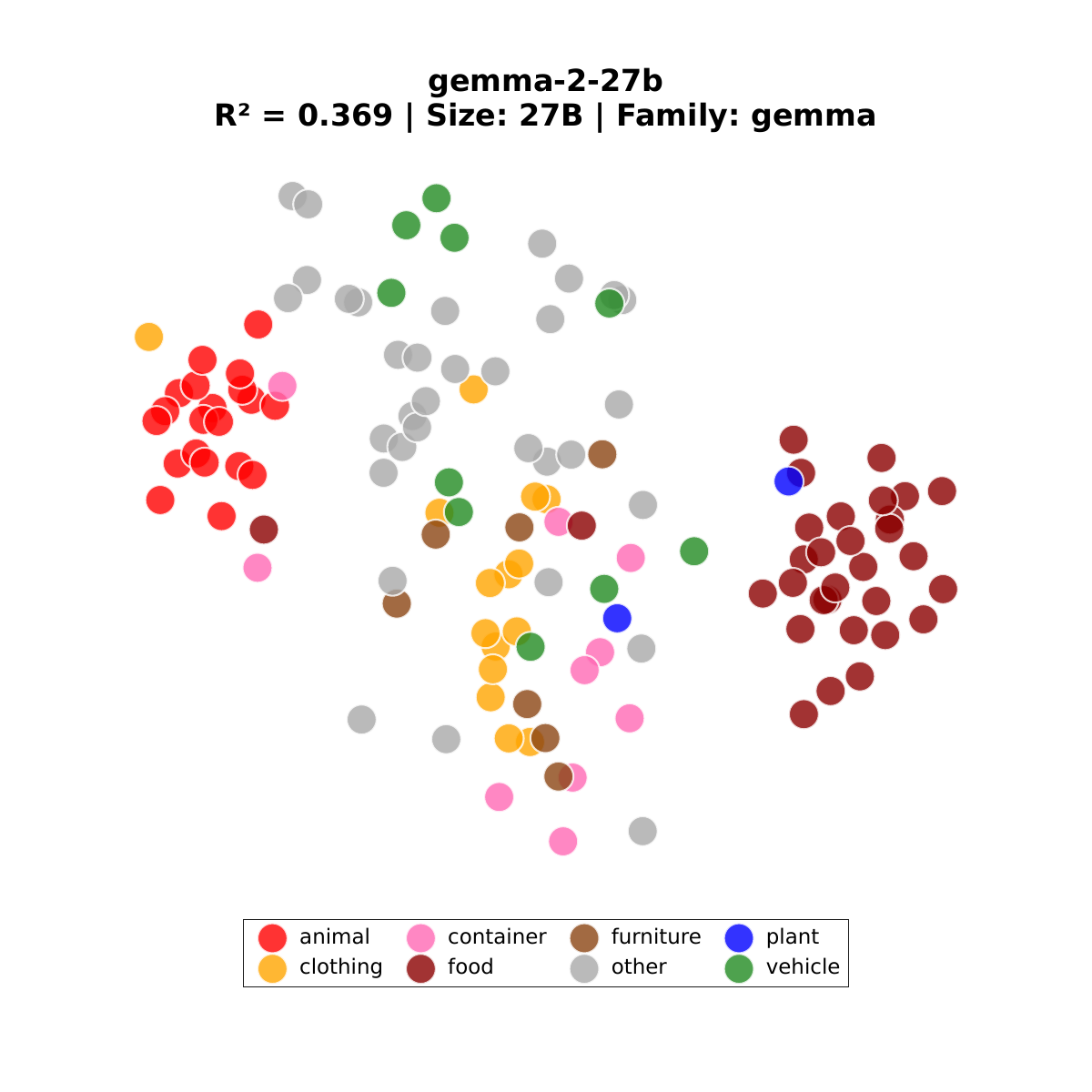} \\
        {\small (22) Qwen3-8B (0.388)} & {\small (23) Instella-3B-Inst (0.374)} & {\small (24) Gemma-2-27B (0.369)} \\
    \end{tabular}
\end{figure}

\begin{figure}[htbp]
    \centering
    \caption{t-SNE visualizations of model embeddings (Part 3). R² scores are in parentheses.}
    \label{fig:tsne_models_part3}
    \begin{tabular}{@{}c@{\hspace{2mm}}c@{\hspace{2mm}}c@{}}
        \includegraphics[width=0.31\textwidth]{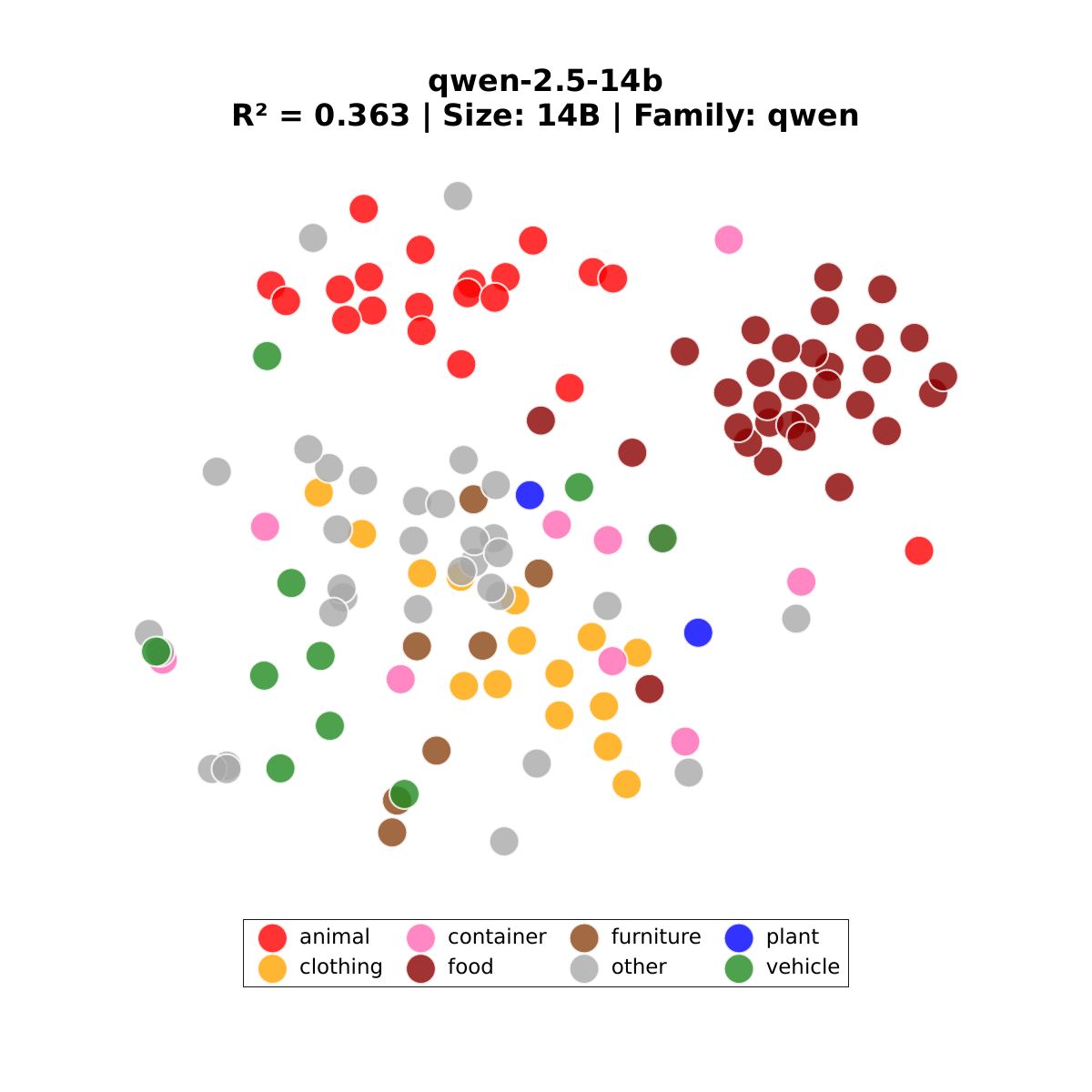} &
        \includegraphics[width=0.31\textwidth]{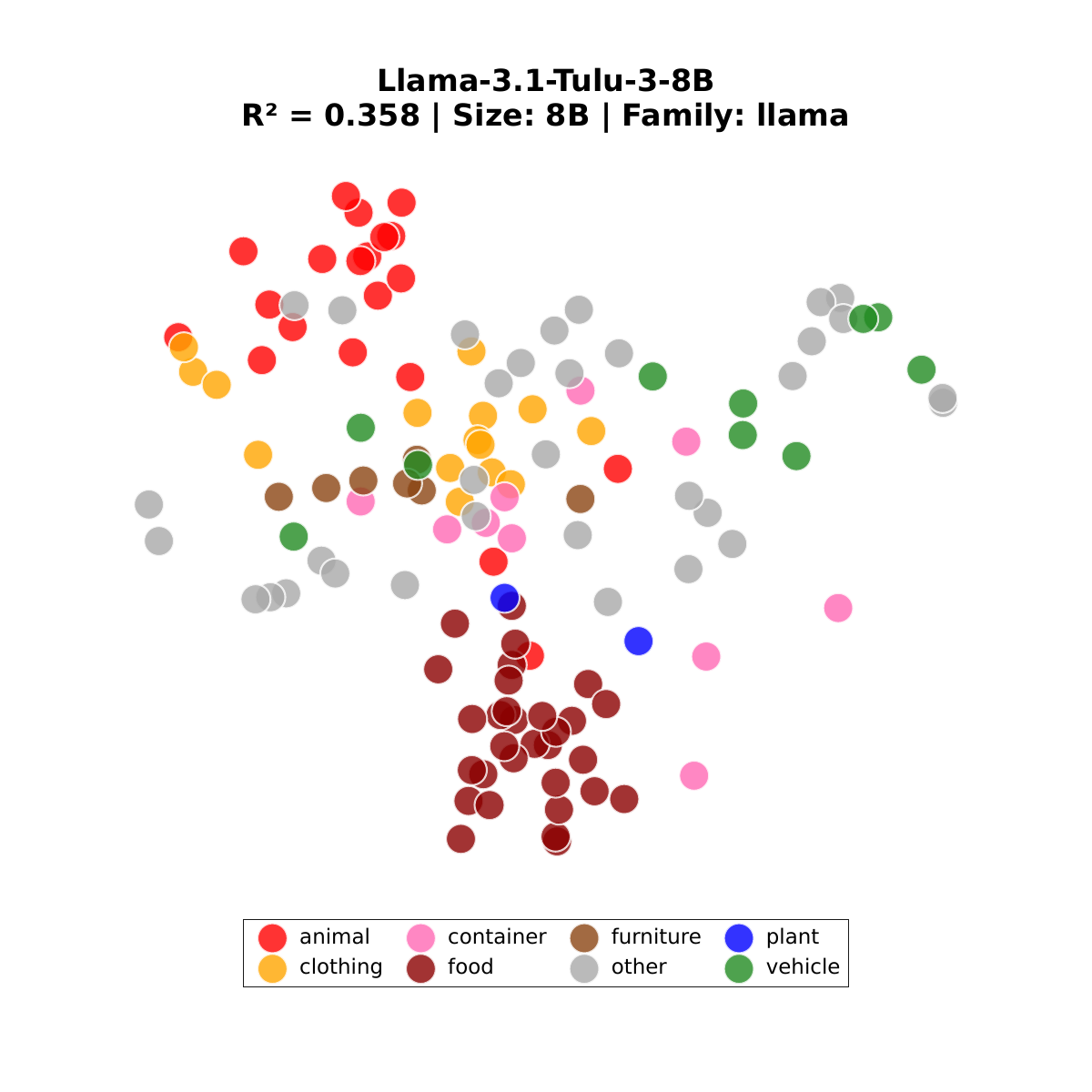} &
        \includegraphics[width=0.31\textwidth]{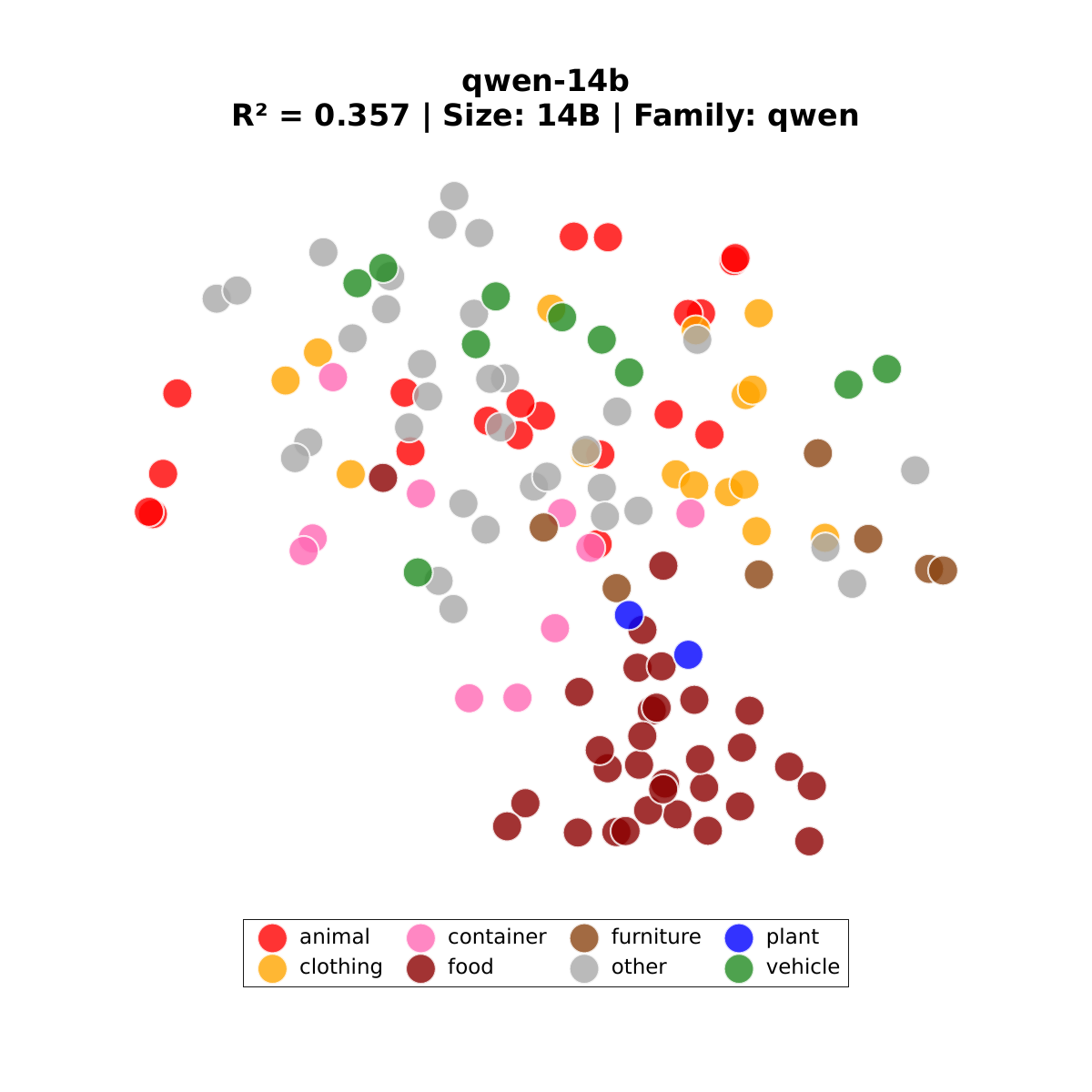} \\
        {\small (25) Qwen2.5-14B (0.363)} & {\small (26) Llama-3.1-Tulu (0.358)} & {\small (27) Qwen-14B (0.357)} \\[3mm]
        
        \includegraphics[width=0.31\textwidth]{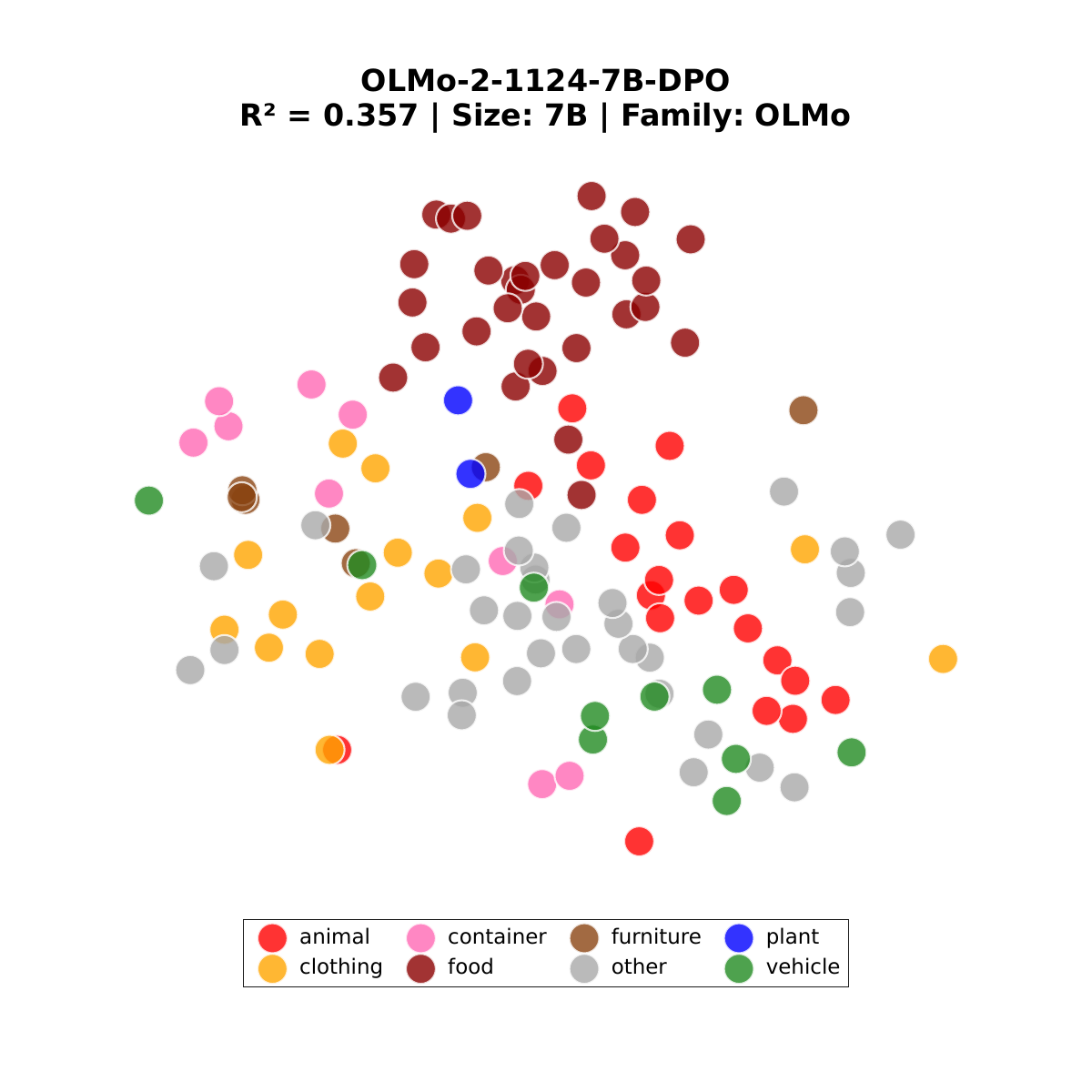} &
        \includegraphics[width=0.31\textwidth]{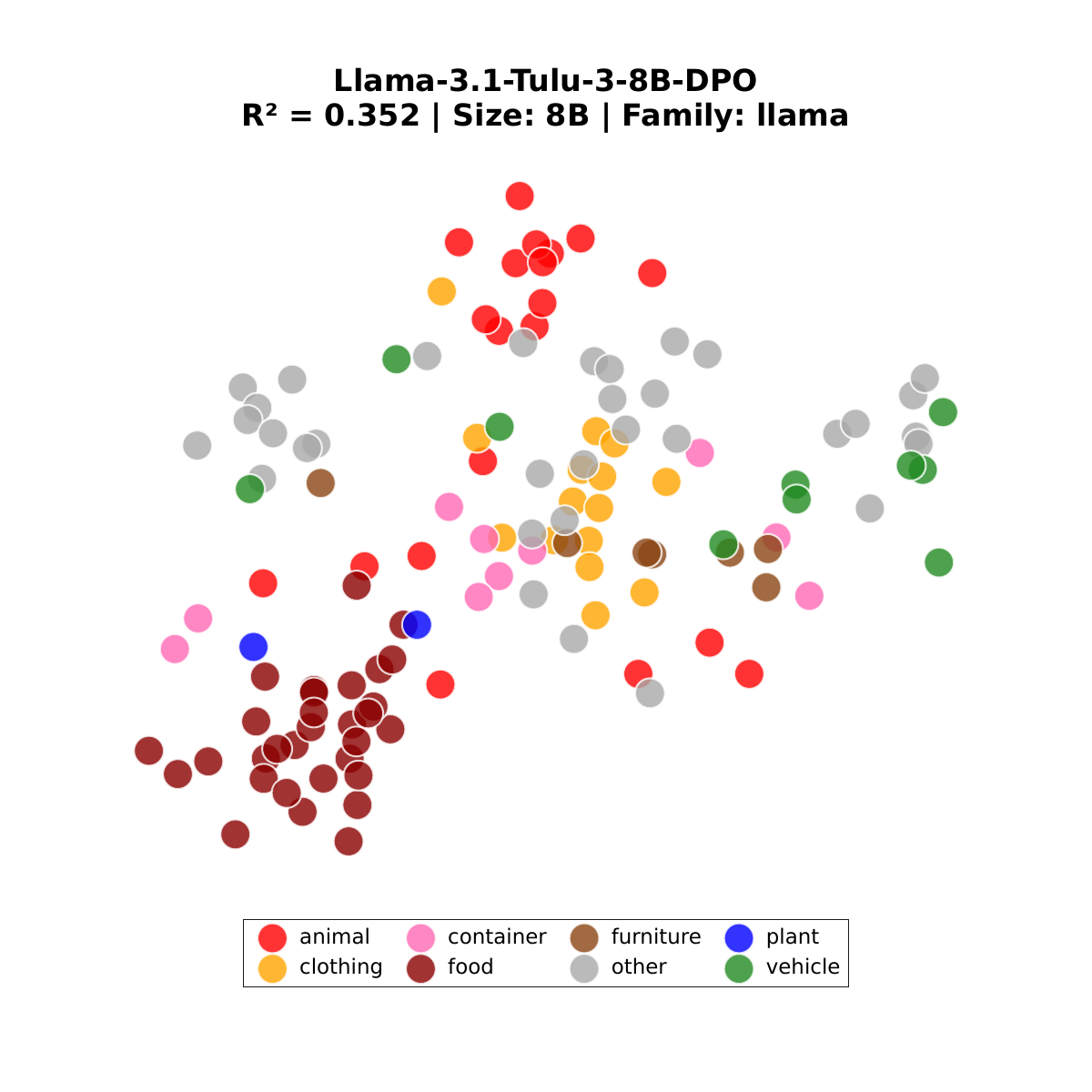} &
        \includegraphics[width=0.31\textwidth]{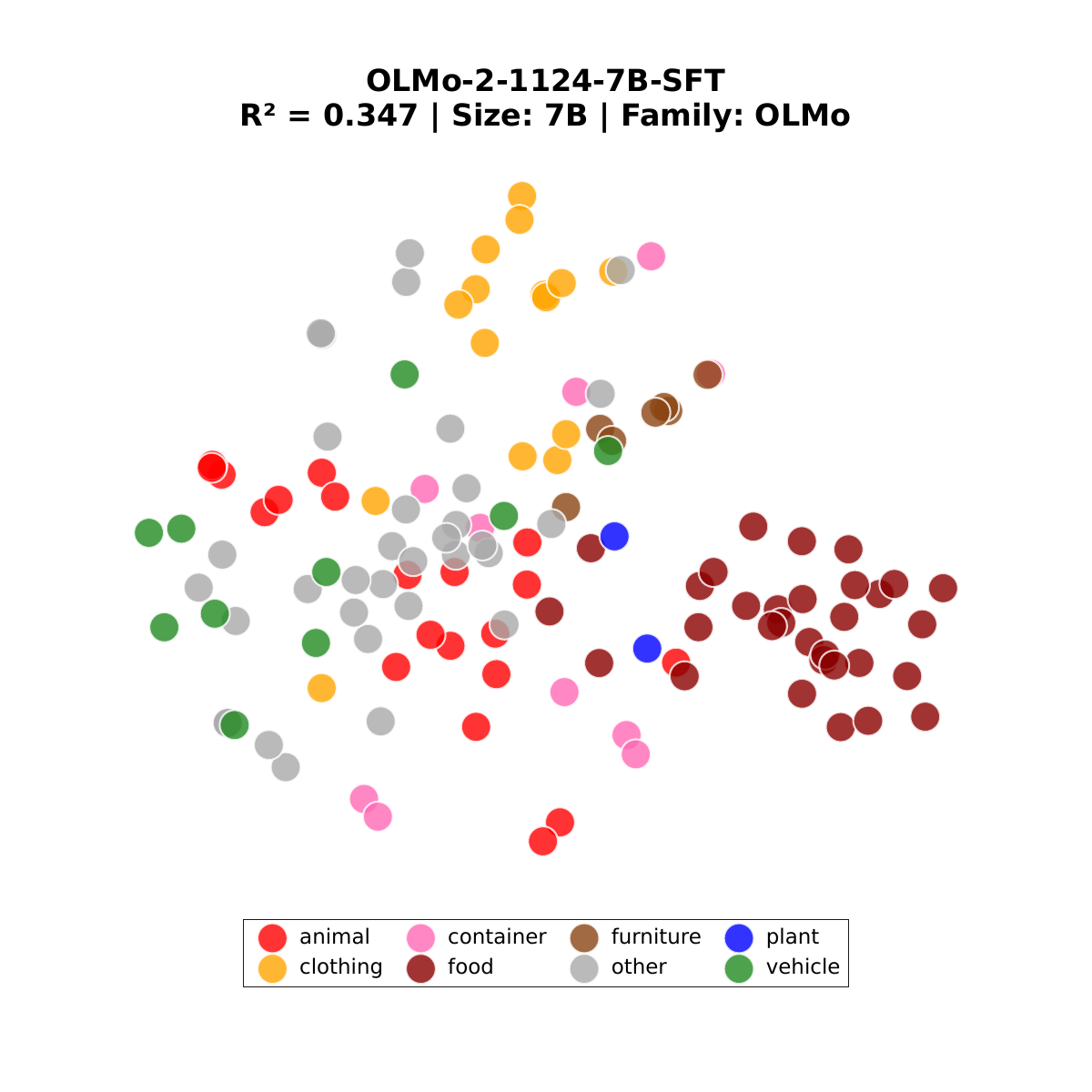} \\
        {\small (28) OLMo-2-7B-DPO (0.357)} & {\small (29) Llama-3.1-Tulu-DPO (0.352)} & {\small (30) OLMo-2-7B-SFT (0.347)} \\[3mm]
        
        \includegraphics[width=0.31\textwidth]{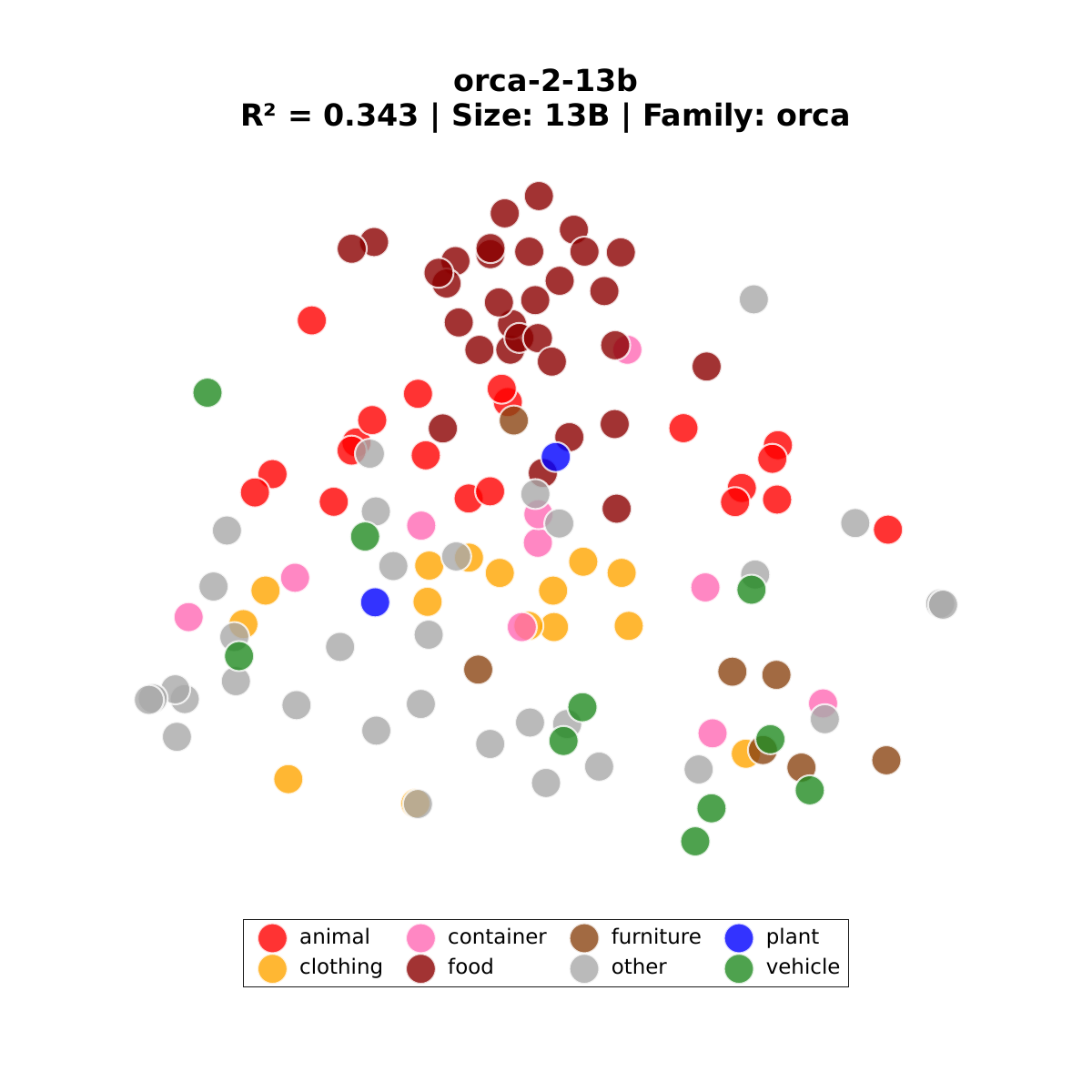} &
        \includegraphics[width=0.31\textwidth]{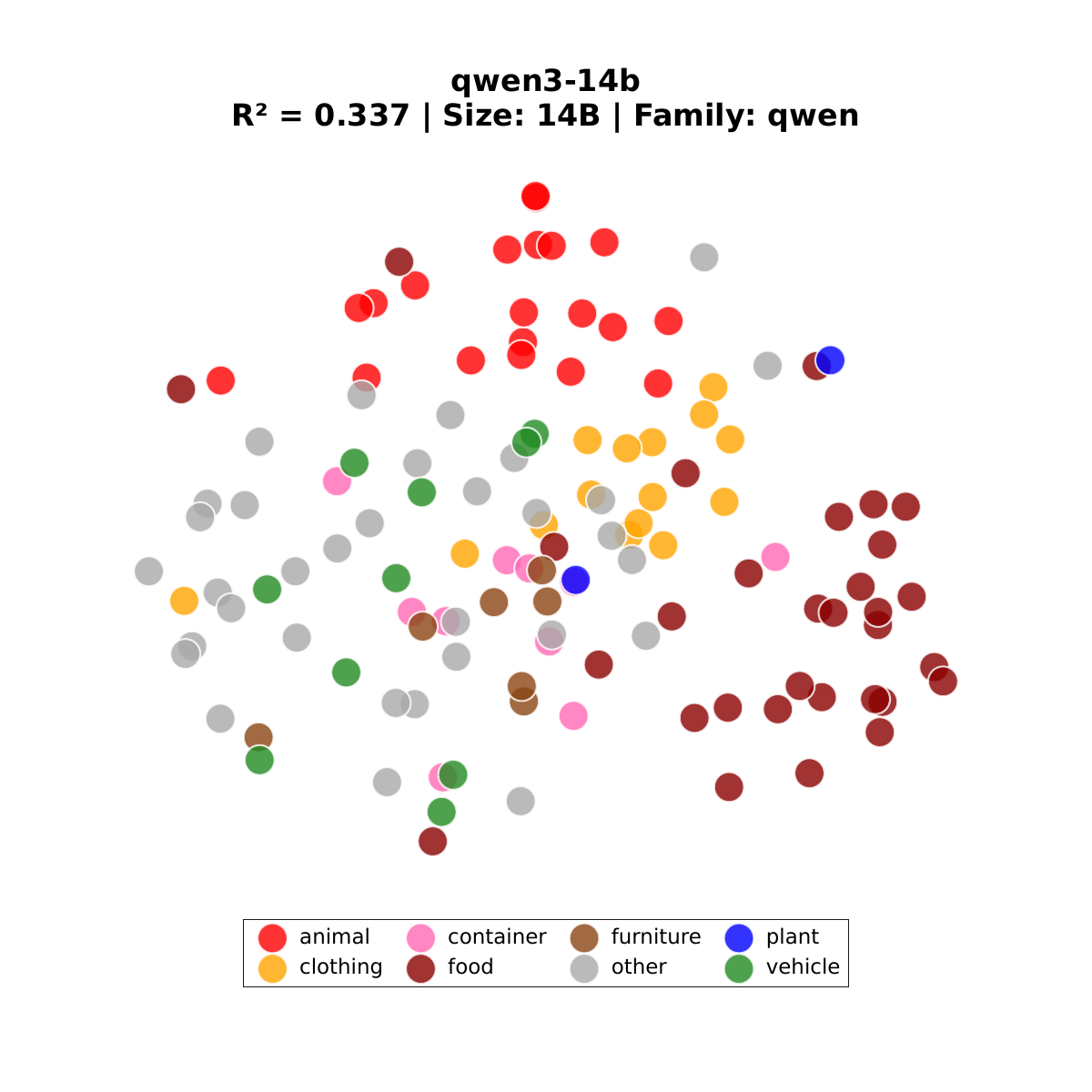} &
        \includegraphics[width=0.31\textwidth]{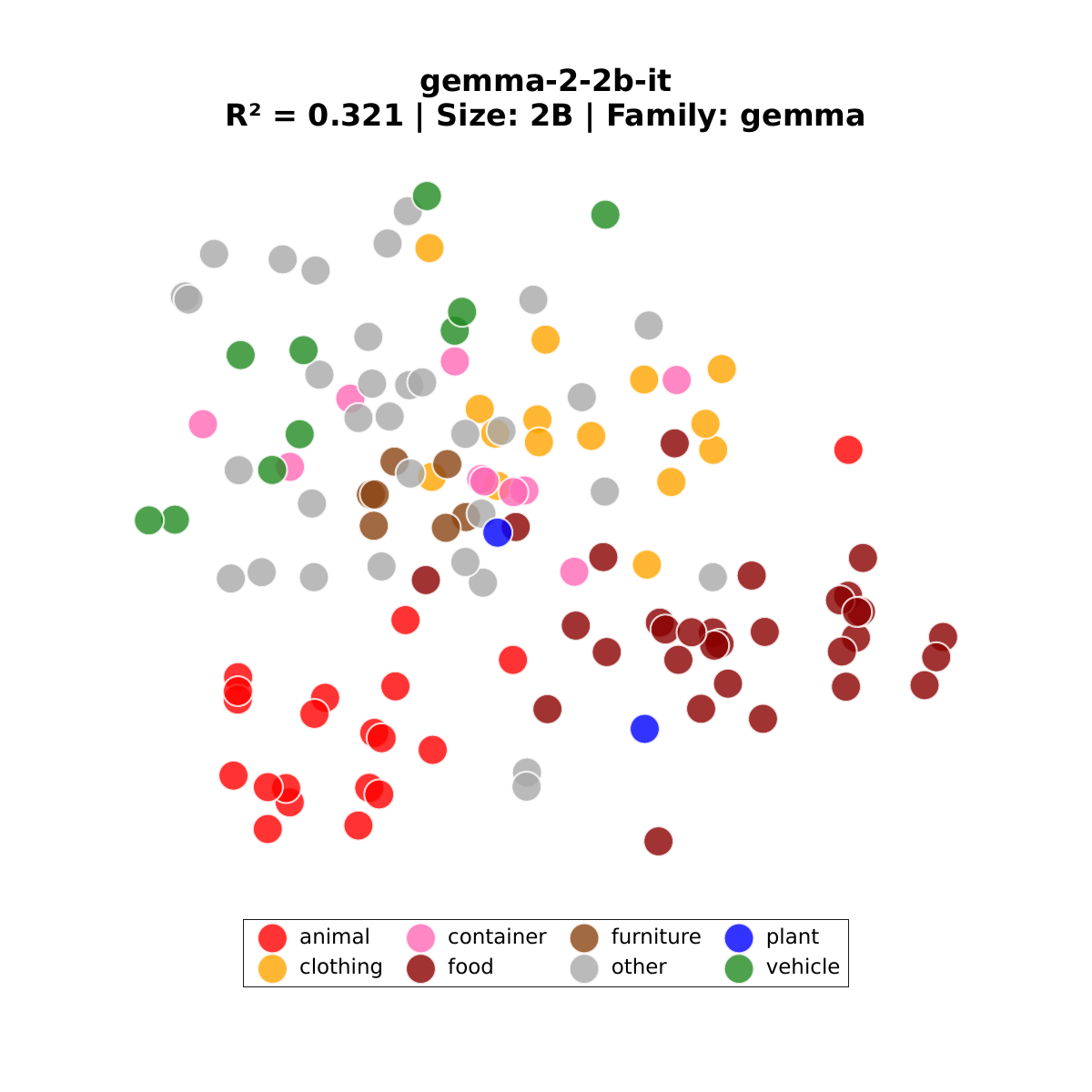} \\
        {\small (31) Orca-2-13B (0.343)} & {\small (32) Qwen3-14B (0.337)} & {\small (33) Gemma-2-2B-IT (0.321)} \\[3mm]
        
        \includegraphics[width=0.31\textwidth]{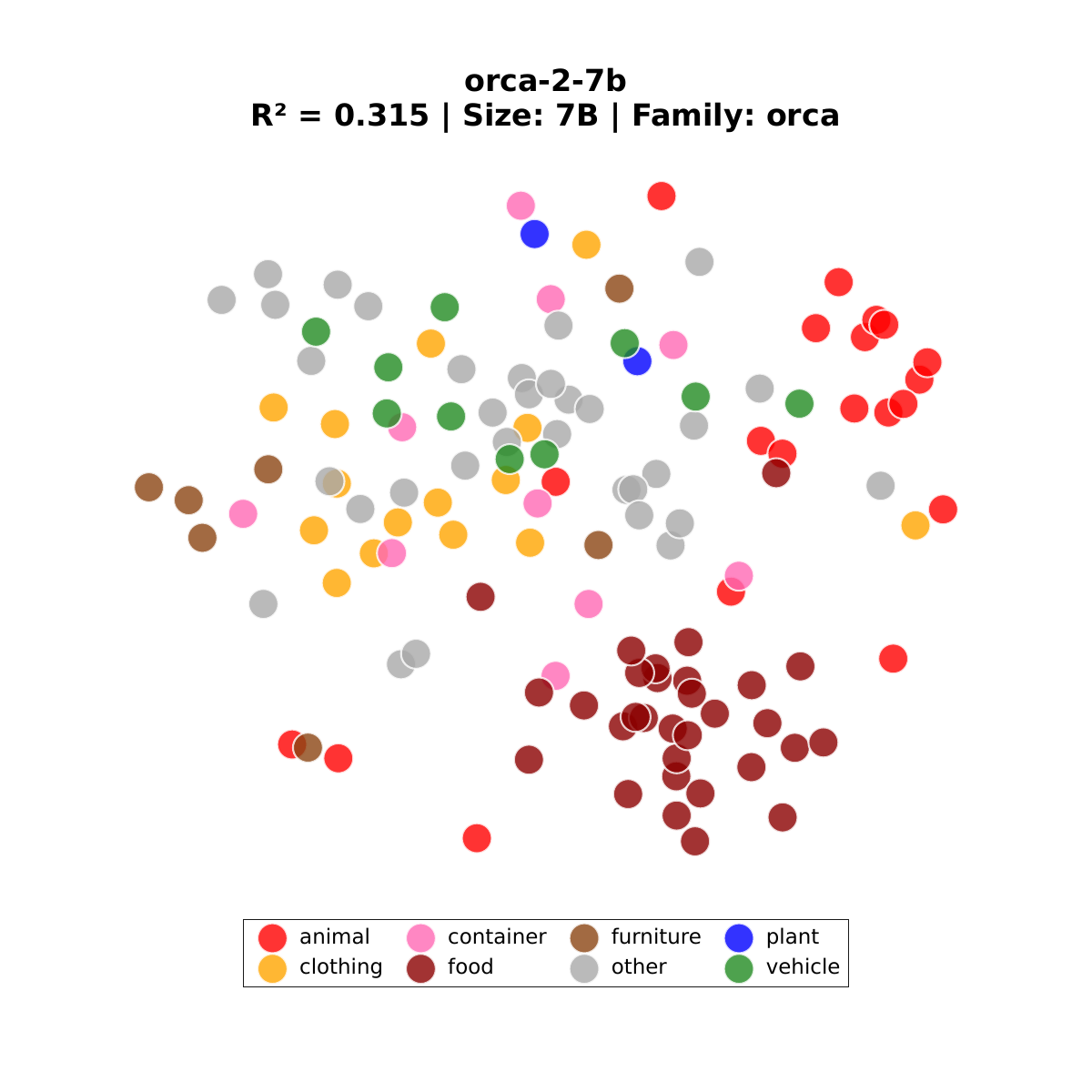} &
        \includegraphics[width=0.31\textwidth]{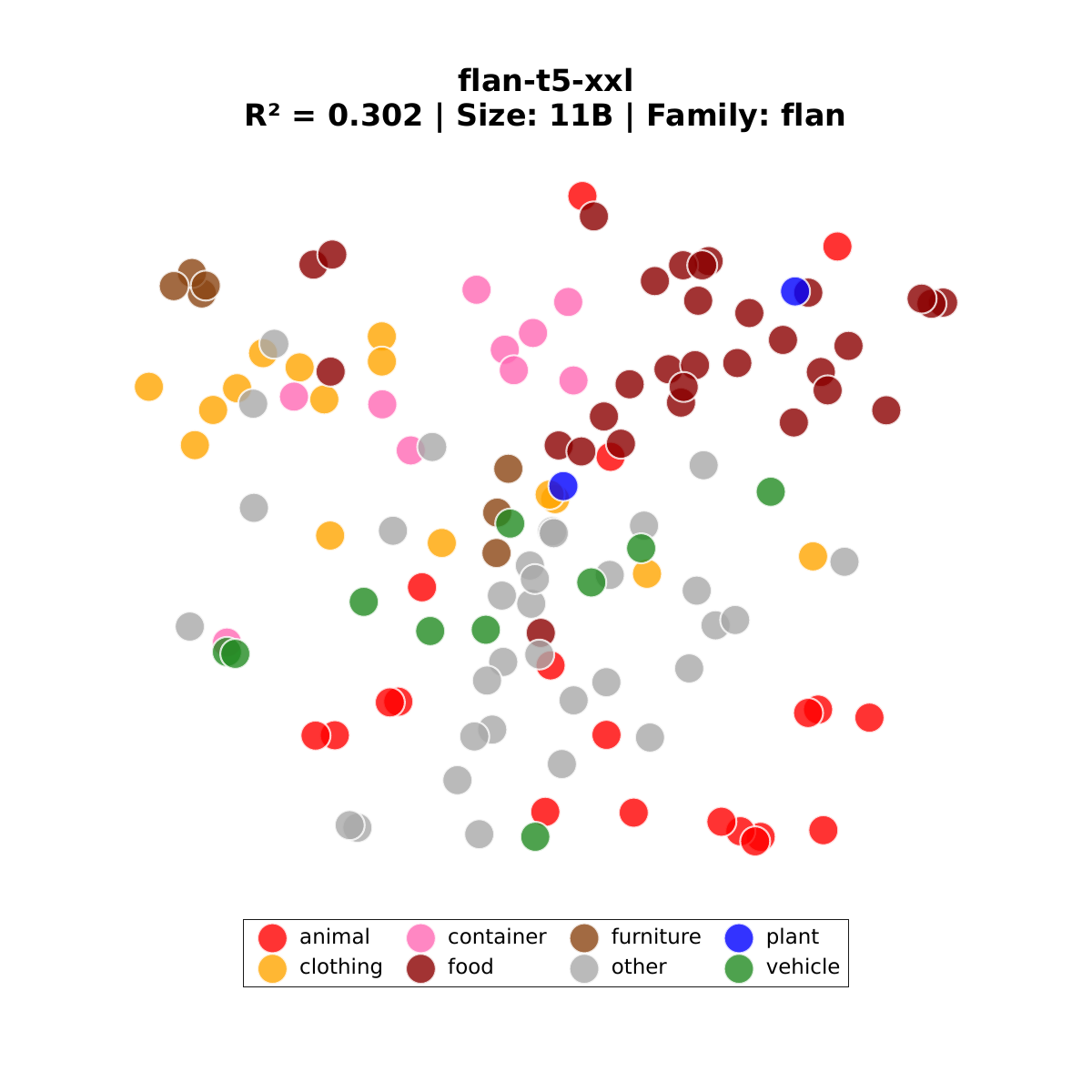} &
        \includegraphics[width=0.31\textwidth]{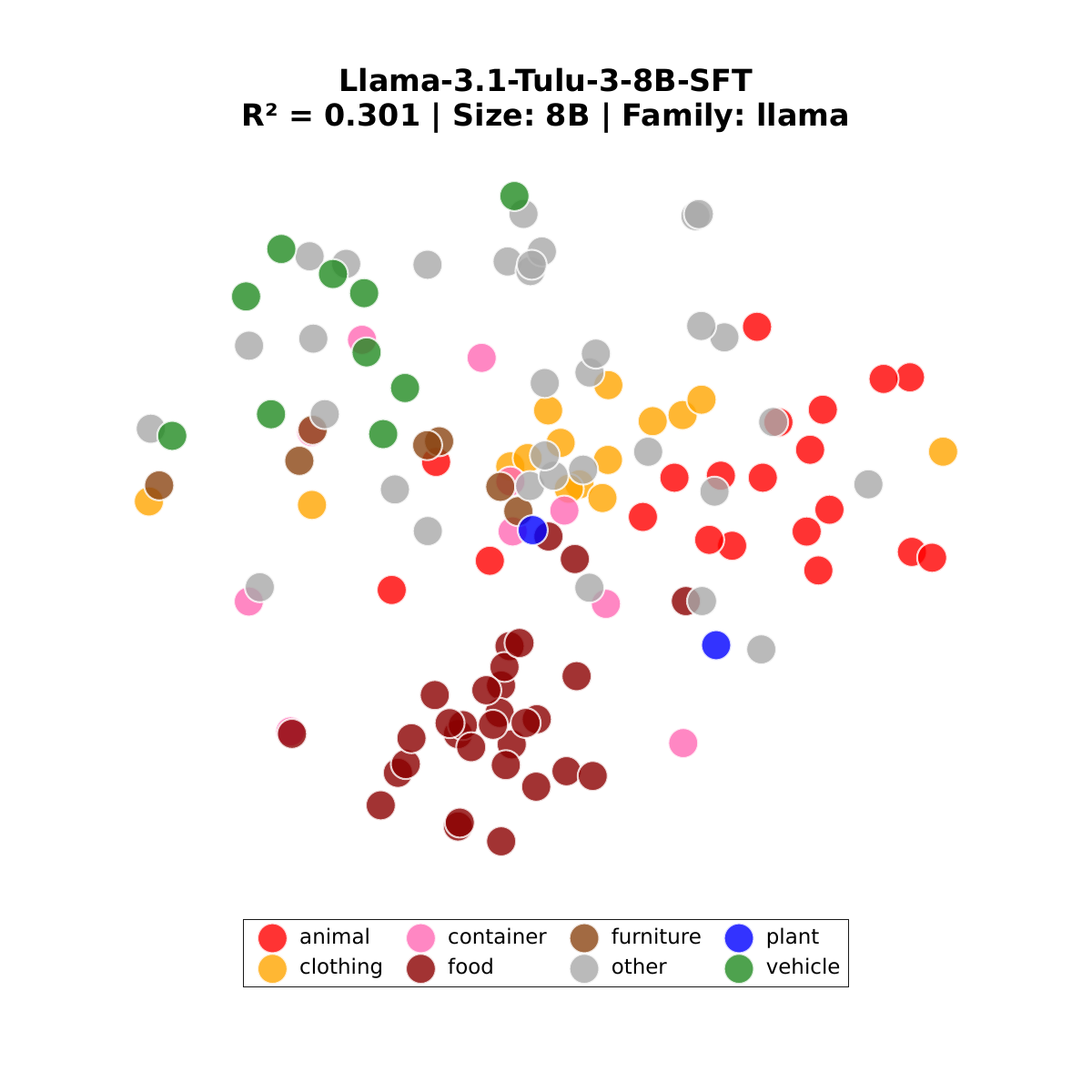} \\
        {\small (34) Orca-2-7B (0.315)} & {\small (35) FLAN-T5-XXL (0.302)} & {\small (36) Llama-3.1-Tulu-SFT (0.301)} \\
    \end{tabular}
\end{figure}

\begin{figure}[htbp]
    \centering
    \caption{t-SNE visualizations of model embeddings (Part 4). R² scores are in parentheses.}
    \label{fig:tsne_models_part4}
    \begin{tabular}{@{}c@{\hspace{2mm}}c@{\hspace{2mm}}c@{}}
        \includegraphics[width=0.31\textwidth]{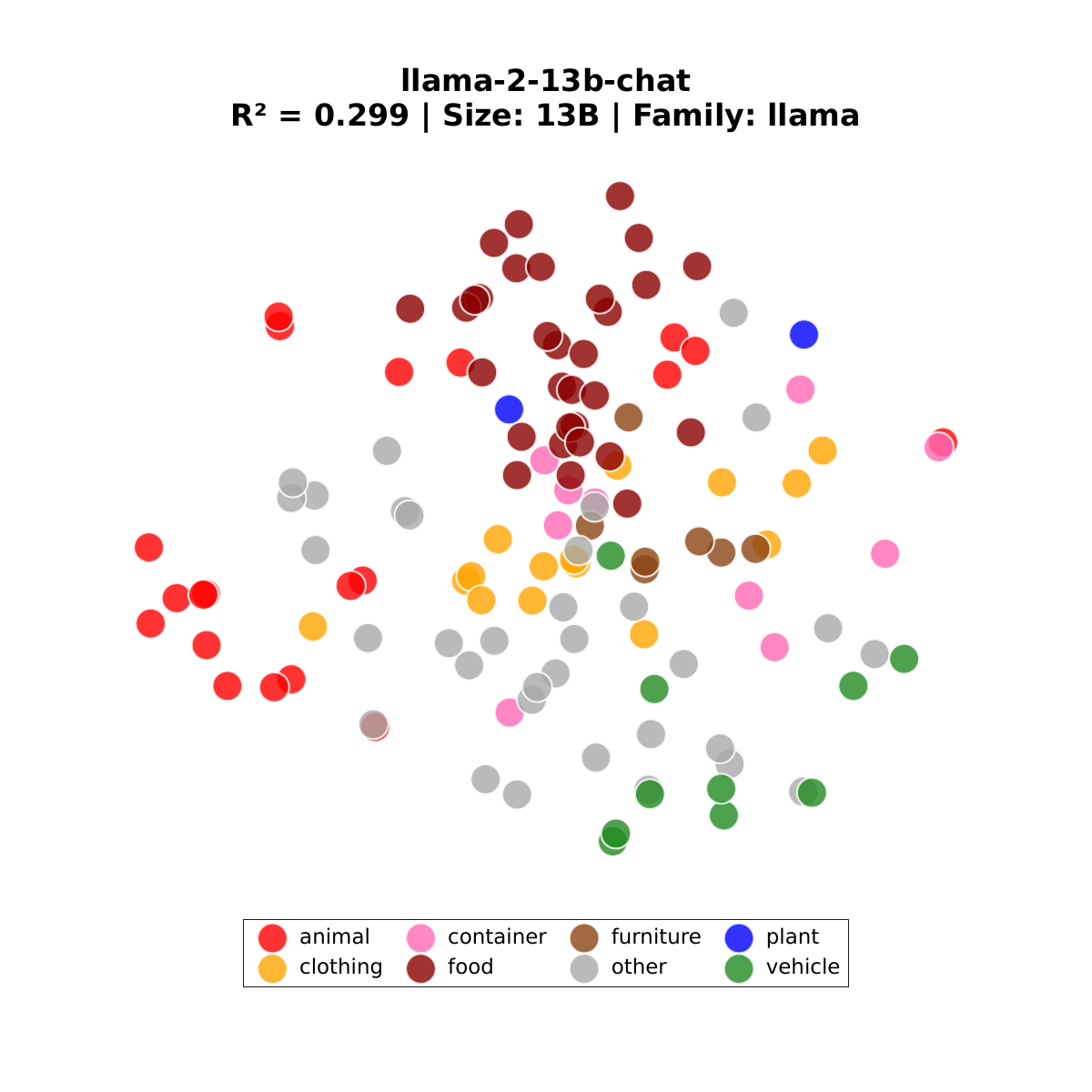} &
        \includegraphics[width=0.31\textwidth]{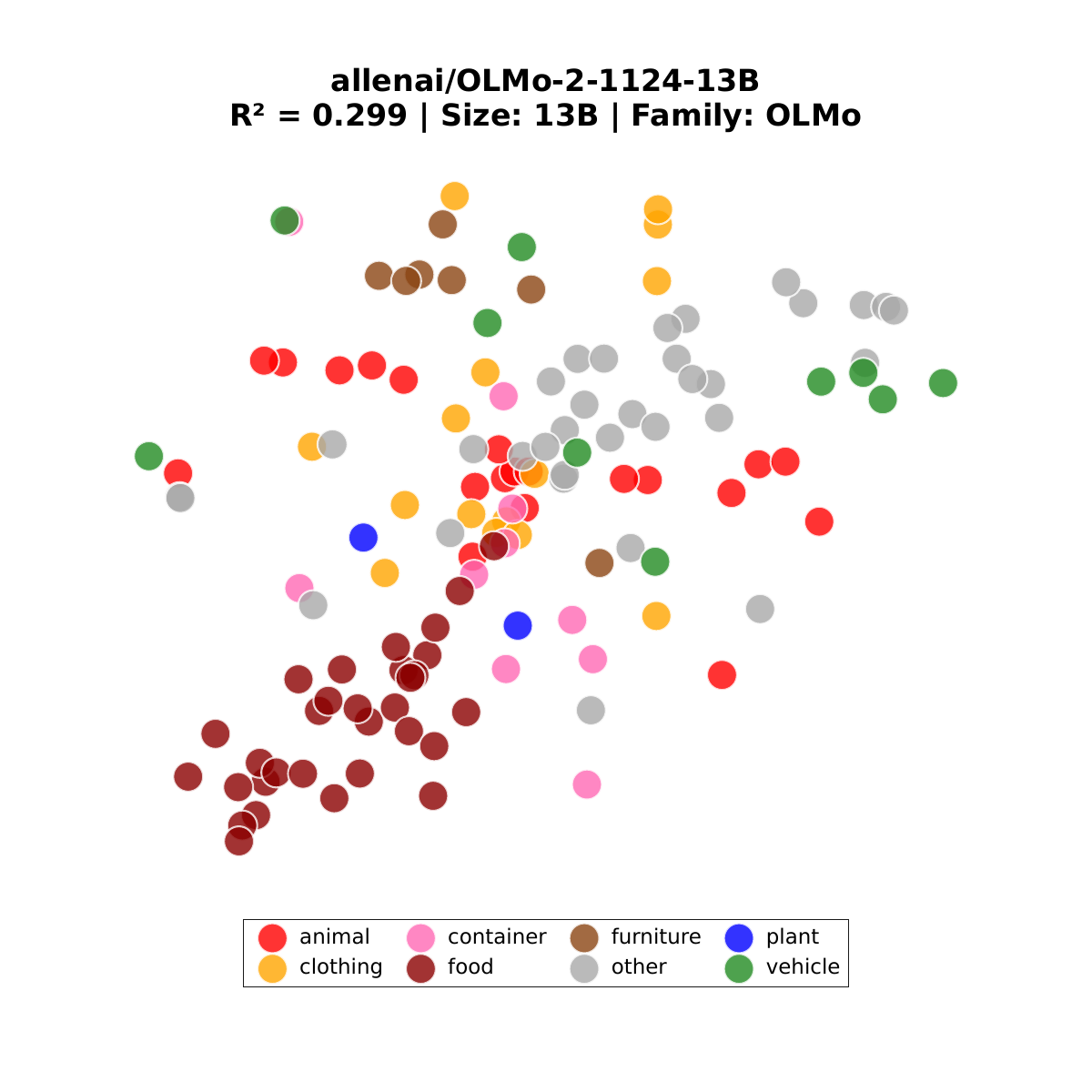} &
        \includegraphics[width=0.31\textwidth]{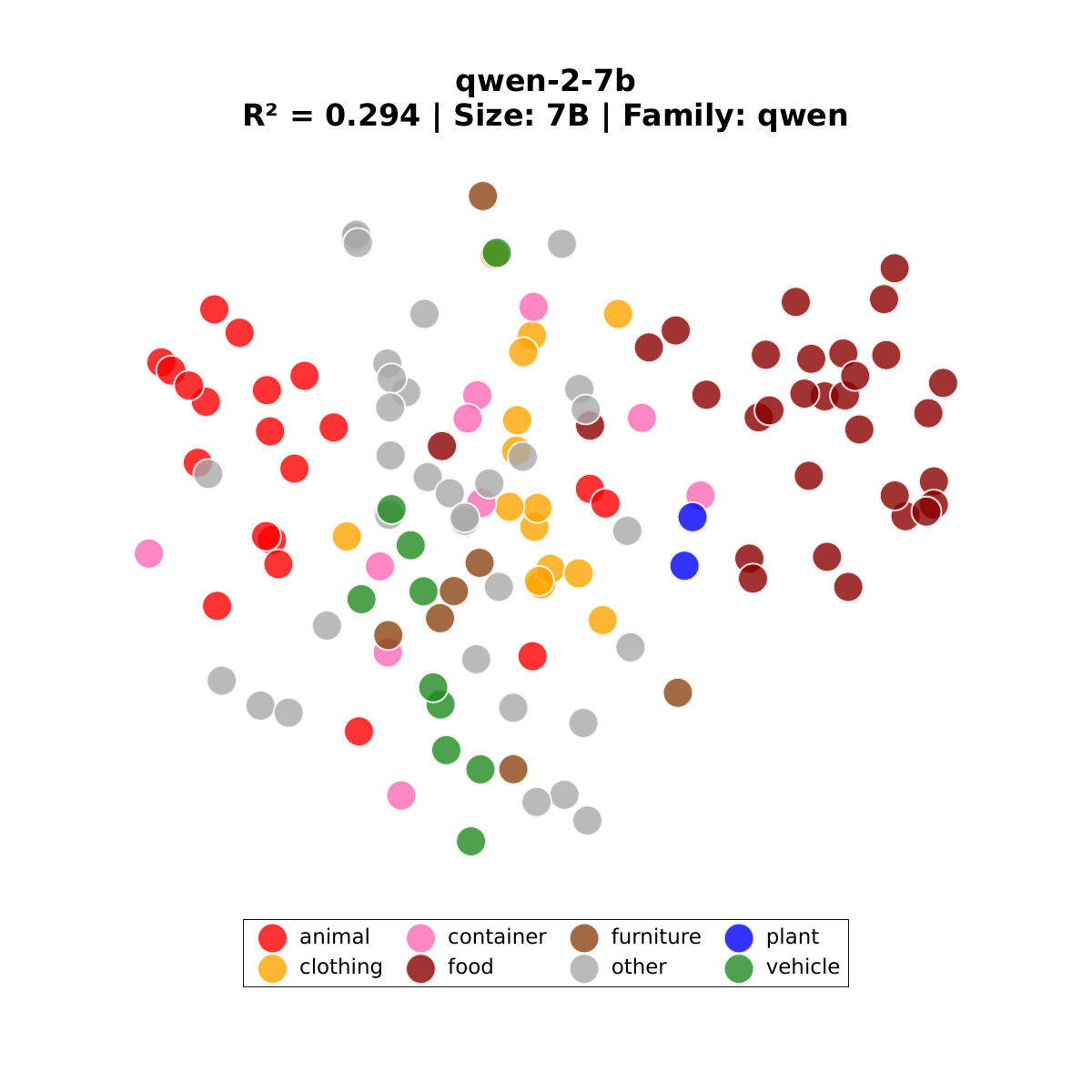} \\
        {\small (37) Llama2-13B-Chat (0.299)} & {\small (38) OLMo-2-13B (0.299)} & {\small (39) Qwen2-7B (0.294)} \\[3mm]
        
        \includegraphics[width=0.31\textwidth]{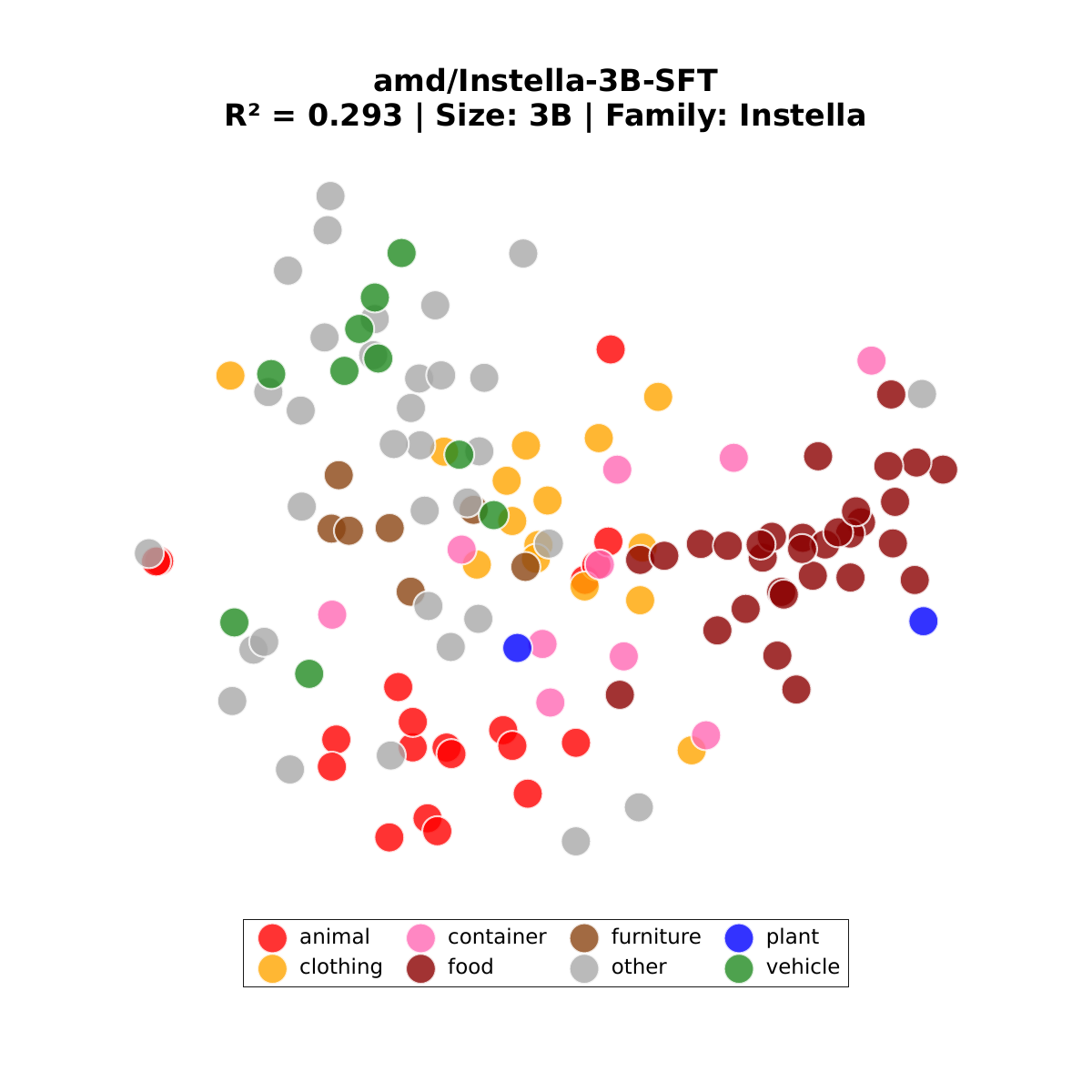} &
        \includegraphics[width=0.31\textwidth]{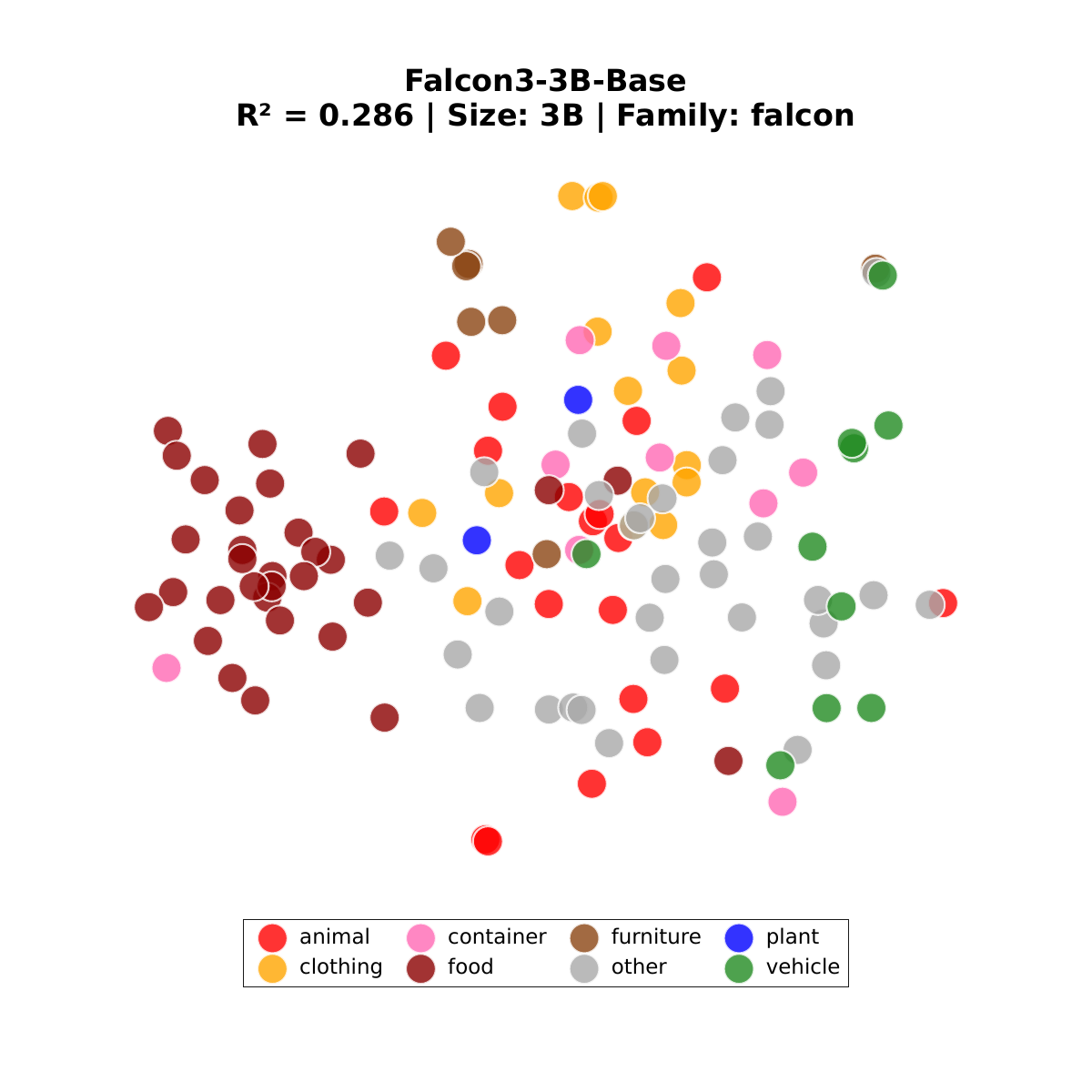} &
        \includegraphics[width=0.31\textwidth]{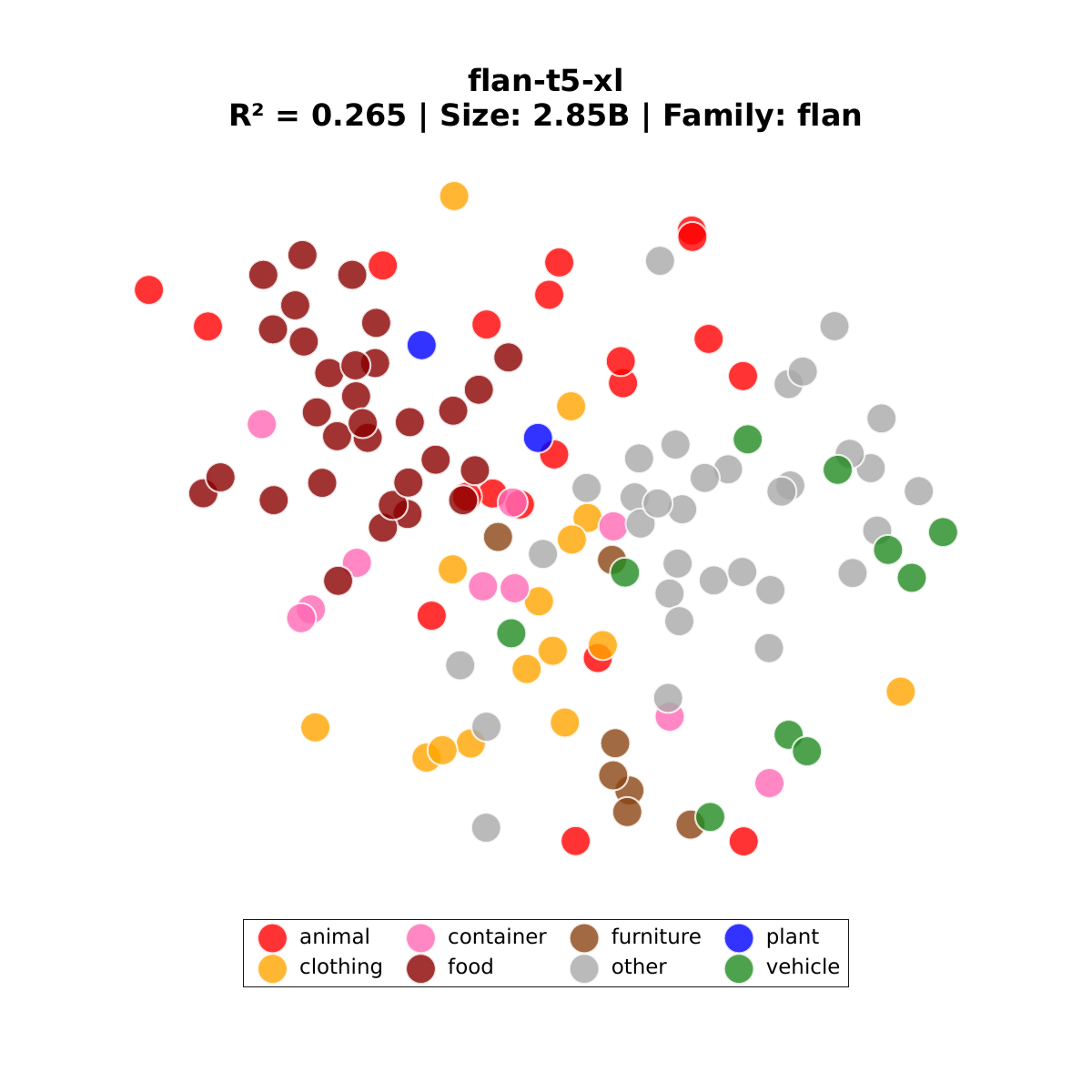} \\
        {\small (40) Instella-3B-SFT (0.293)} & {\small (41) Falcon3-3B (0.286)} & {\small (42) FLAN-T5-XL (0.265)} \\[3mm]
        
        \includegraphics[width=0.31\textwidth]{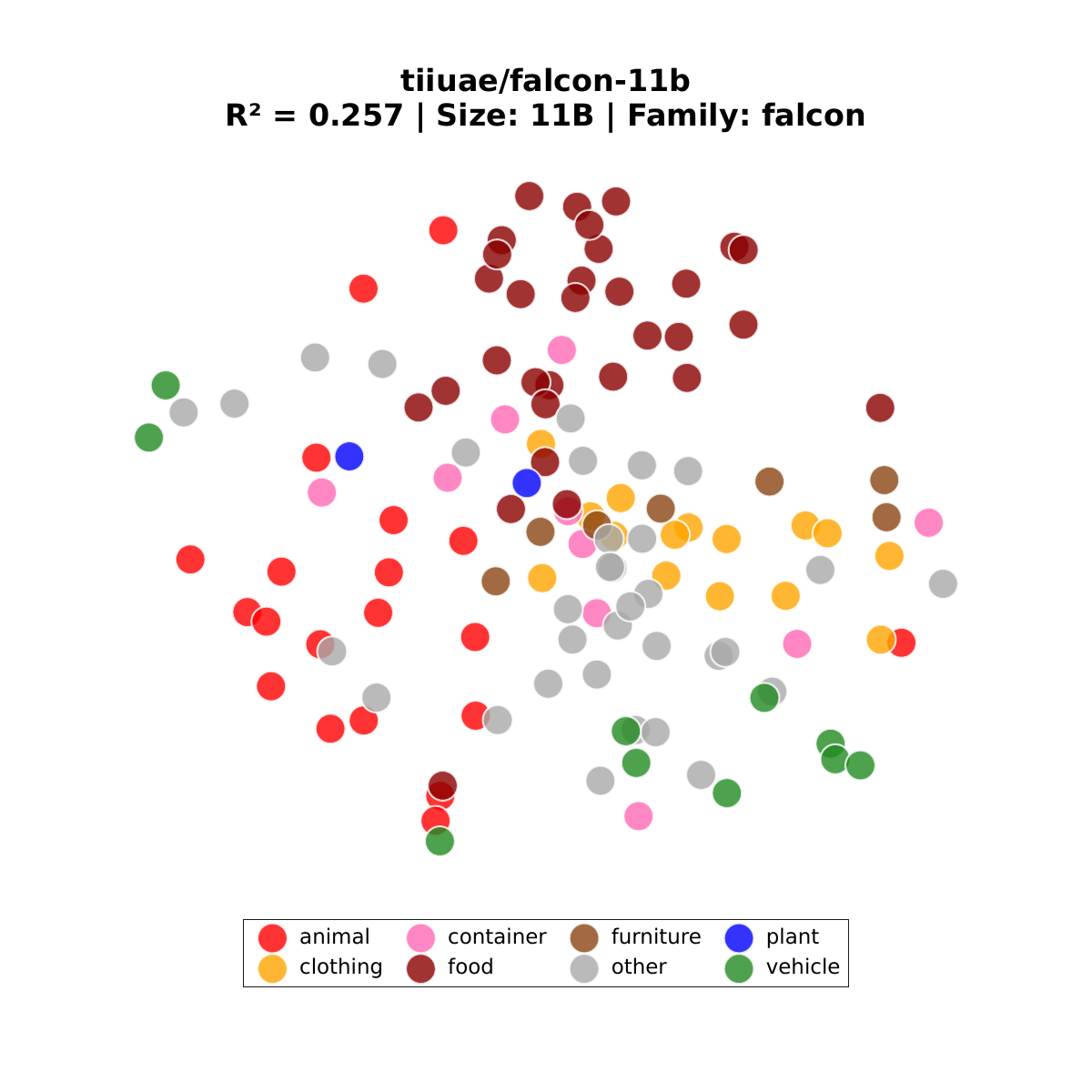} &
        \includegraphics[width=0.31\textwidth]{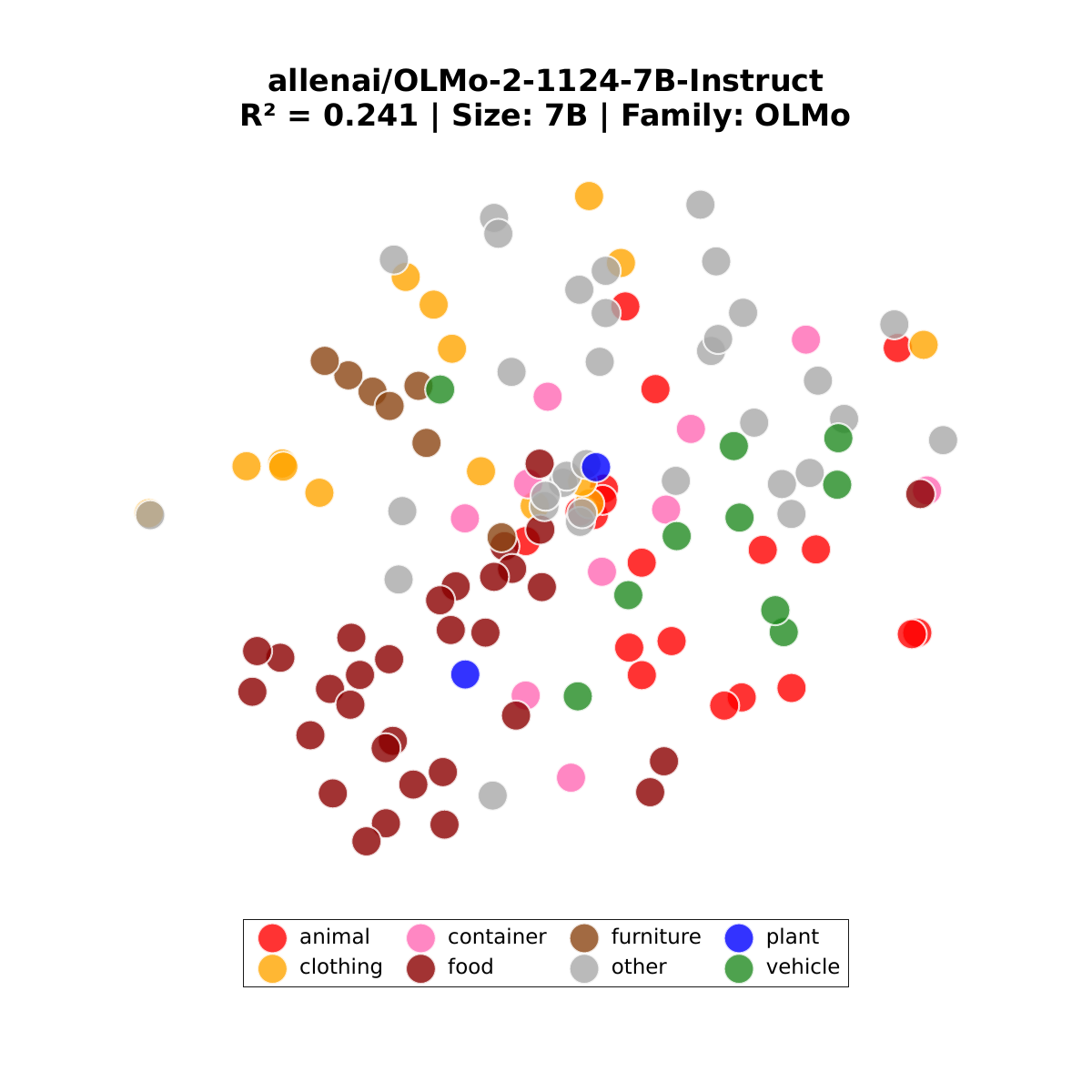} &
        \includegraphics[width=0.31\textwidth]{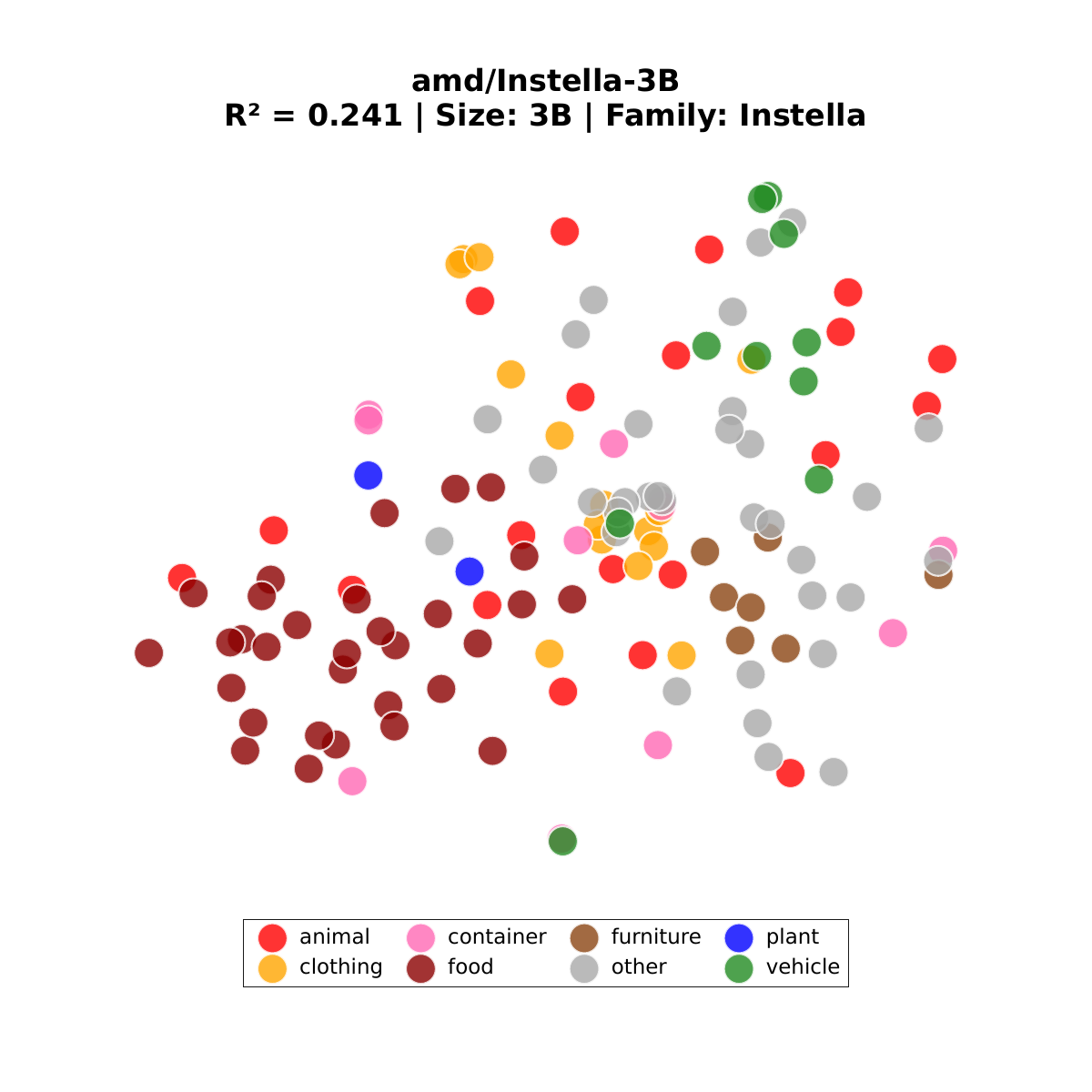} \\
        {\small (43) Falcon-11B (0.257)} & {\small (44) OLMo-2-7B-Inst (0.241)} & {\small (45) Instella-3B (0.241)} \\[3mm]
        
        \includegraphics[width=0.31\textwidth]{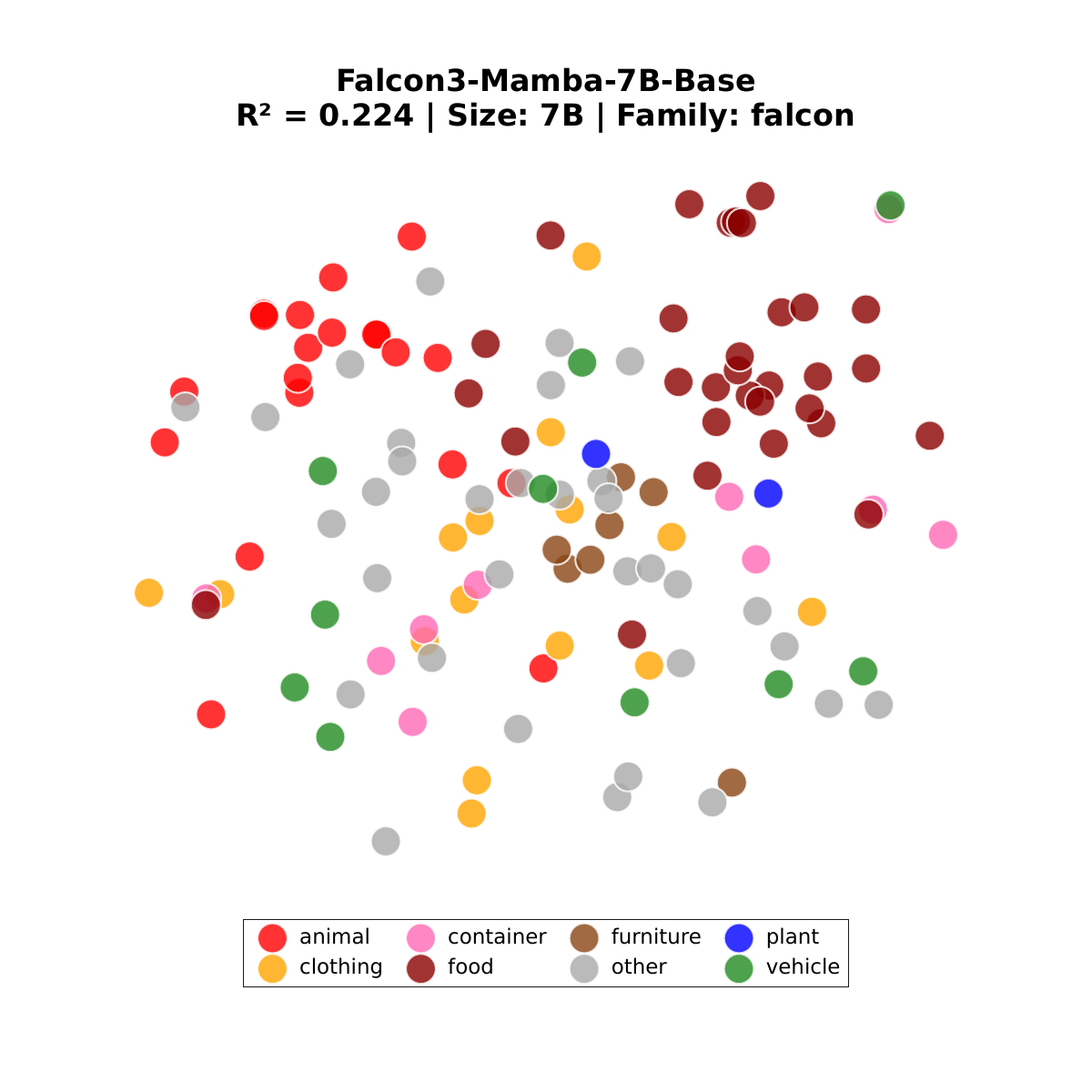} &
        \includegraphics[width=0.31\textwidth]{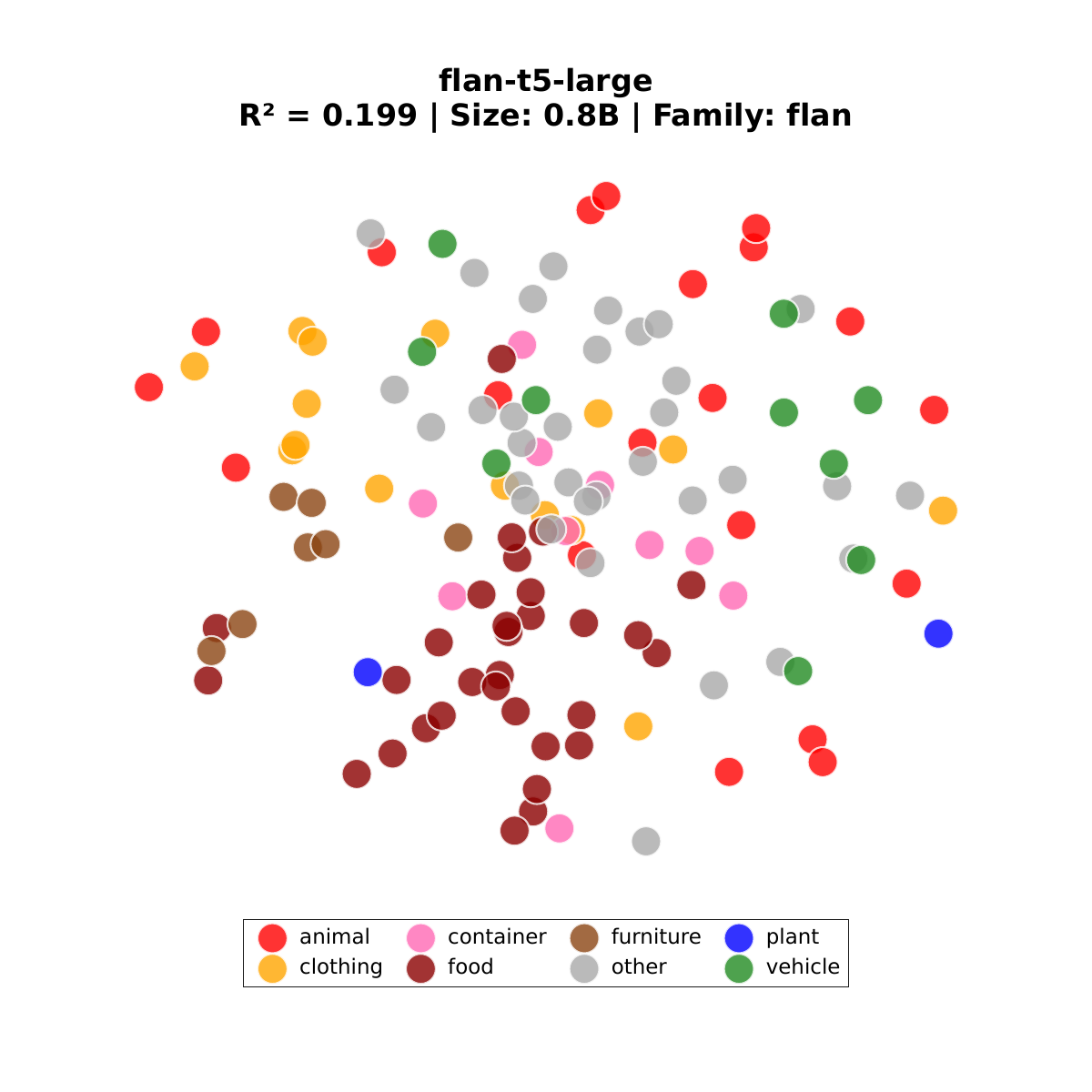} &
        \includegraphics[width=0.31\textwidth]{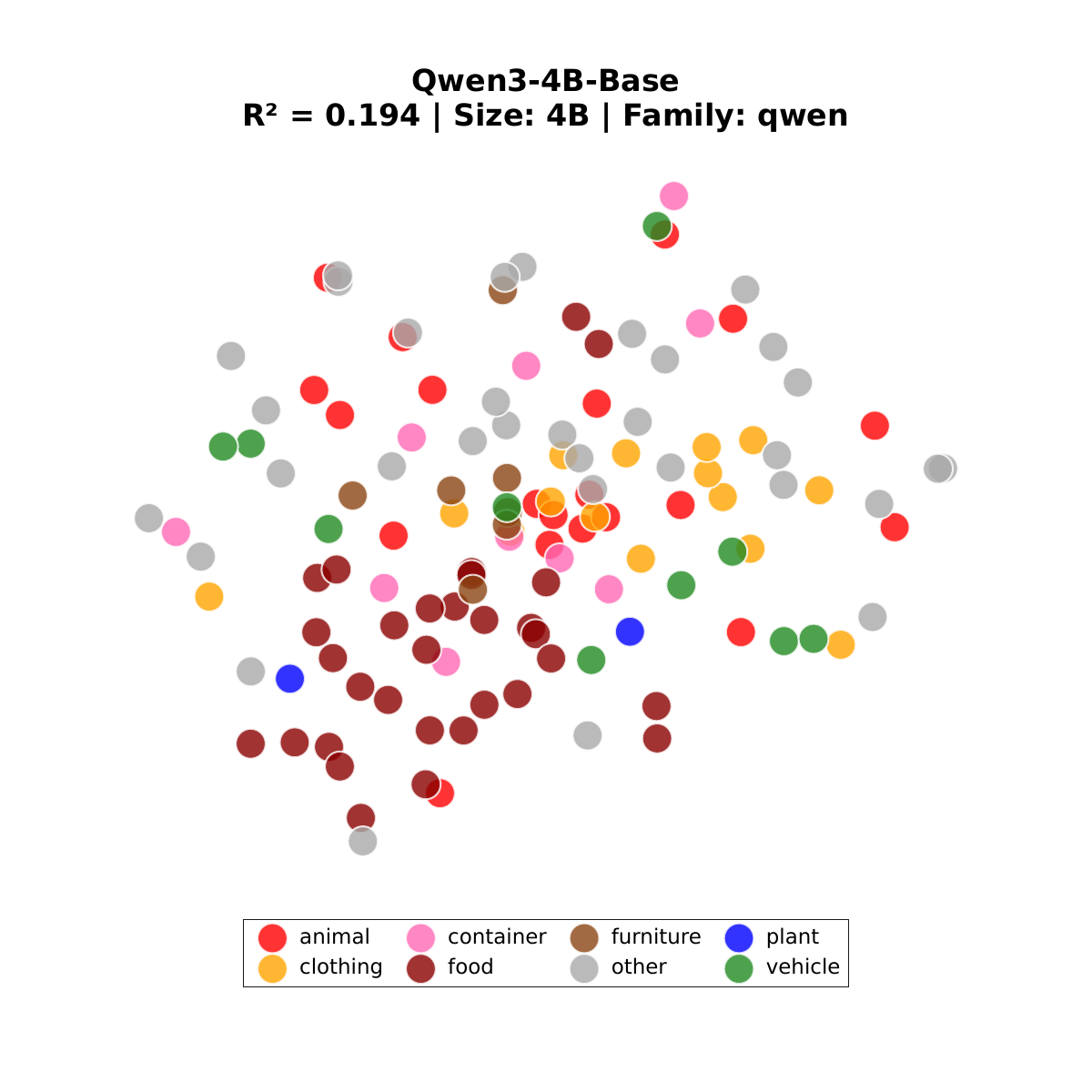} \\
        {\small (46) Falcon3-Mamba-7B (0.224)} & {\small (47) FLAN-T5-Large (0.199)} & {\small (48) Qwen3-4B (0.194)} \\
    \end{tabular}
\end{figure}

\begin{figure}[htbp]
    \centering
    \caption{t-SNE visualizations of model embeddings (Part 5). R² scores are in parentheses.}
    \label{fig:tsne_models_part5}
    \begin{tabular}{@{}c@{\hspace{2mm}}c@{\hspace{2mm}}c@{}}
        \includegraphics[width=0.31\textwidth]{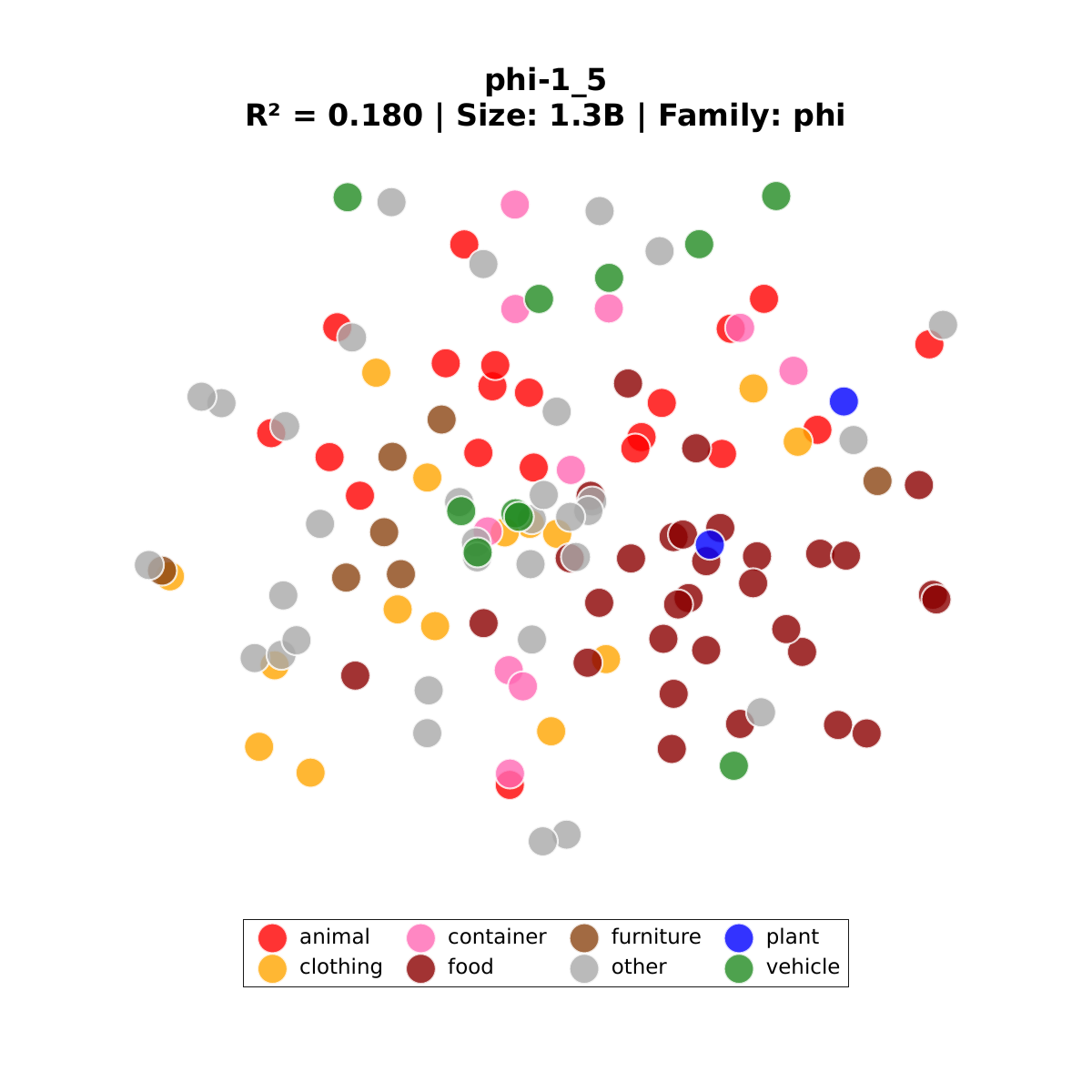} &
        \includegraphics[width=0.31\textwidth]{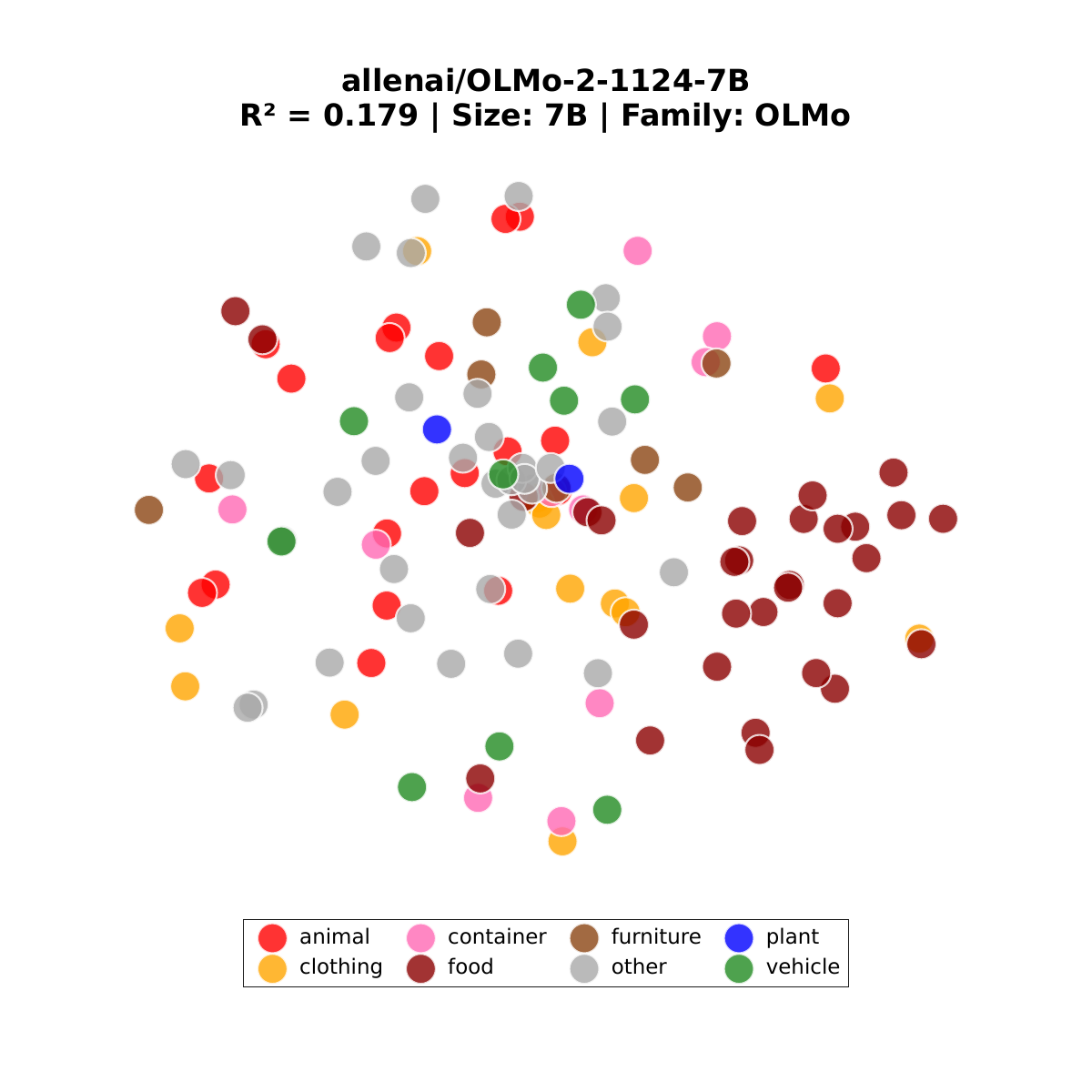} &
        \includegraphics[width=0.31\textwidth]{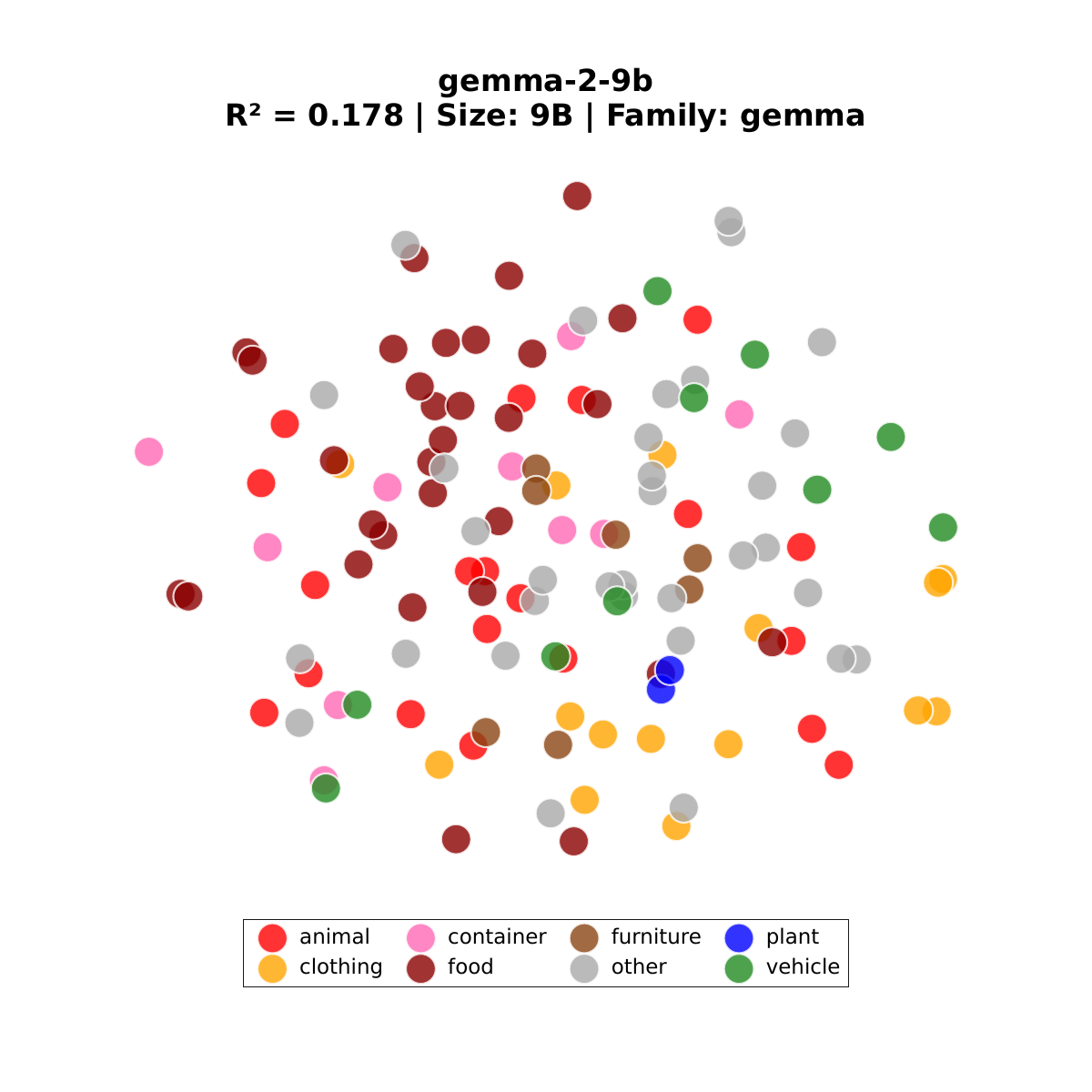} \\
        {\small (49) Phi-1.5 (0.180)} & {\small (50) OLMo-2-7B (0.179)} & {\small (51) Gemma-2-9B (0.178)} \\[3mm]
        
        \includegraphics[width=0.31\textwidth]{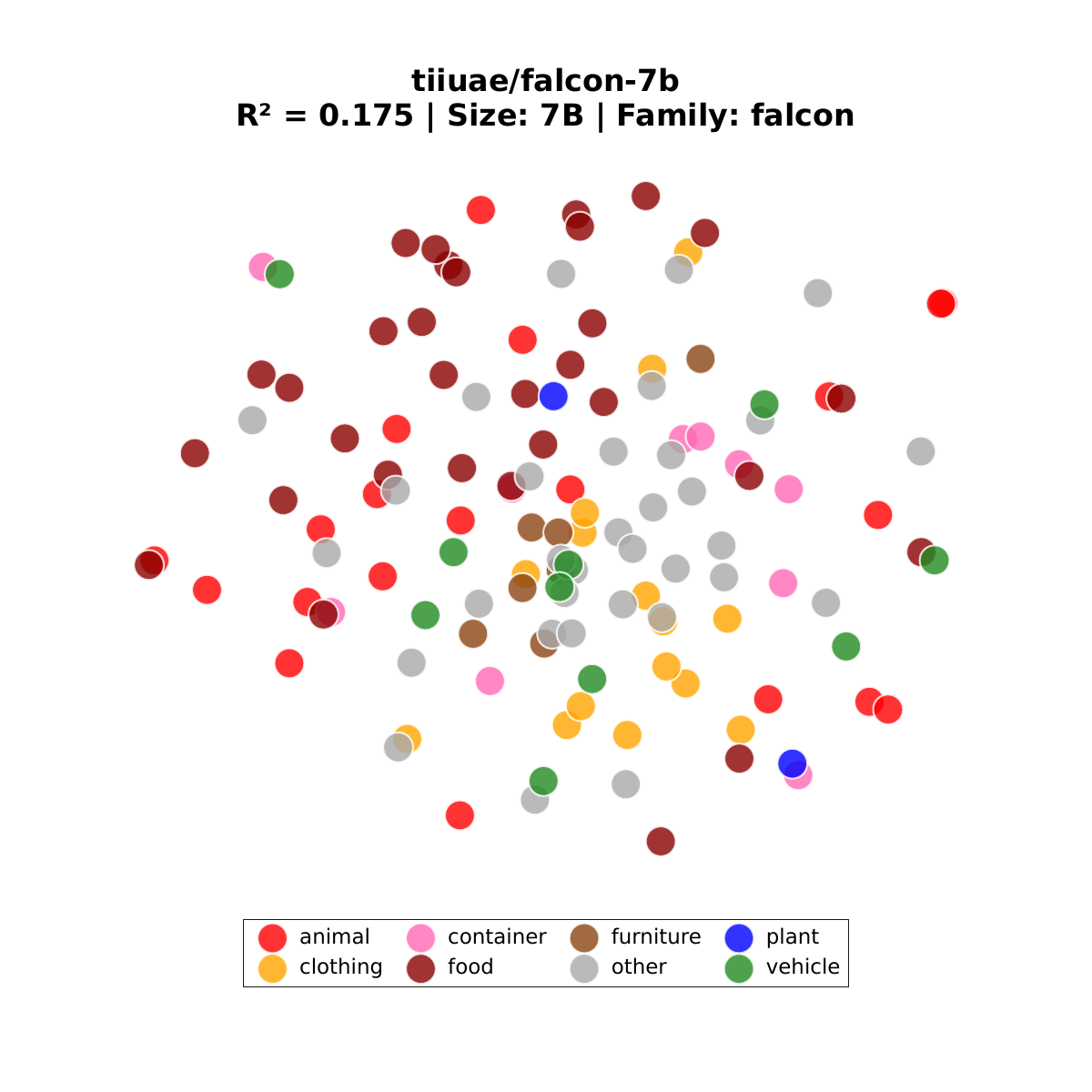} &
        \includegraphics[width=0.31\textwidth]{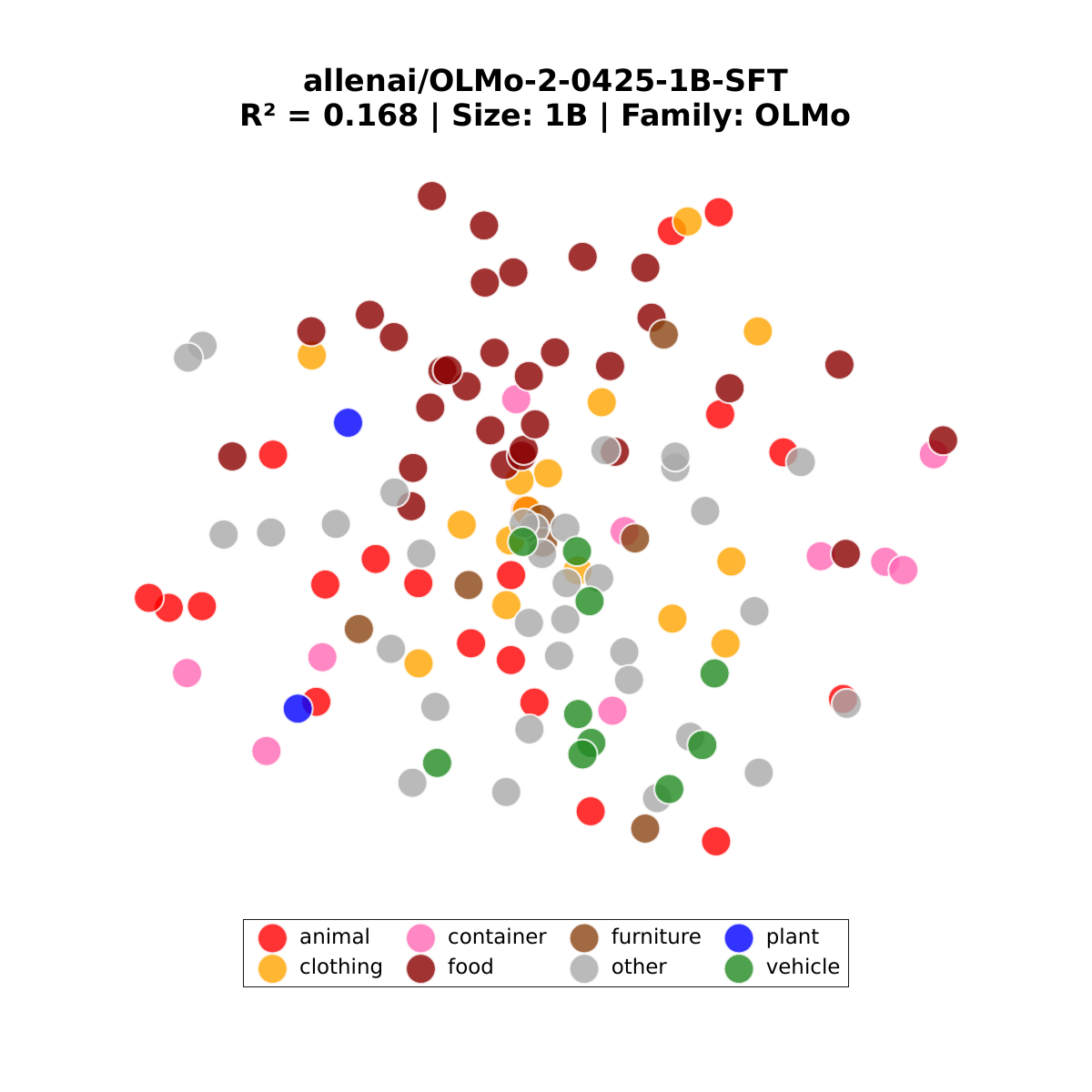} &
        \includegraphics[width=0.31\textwidth]{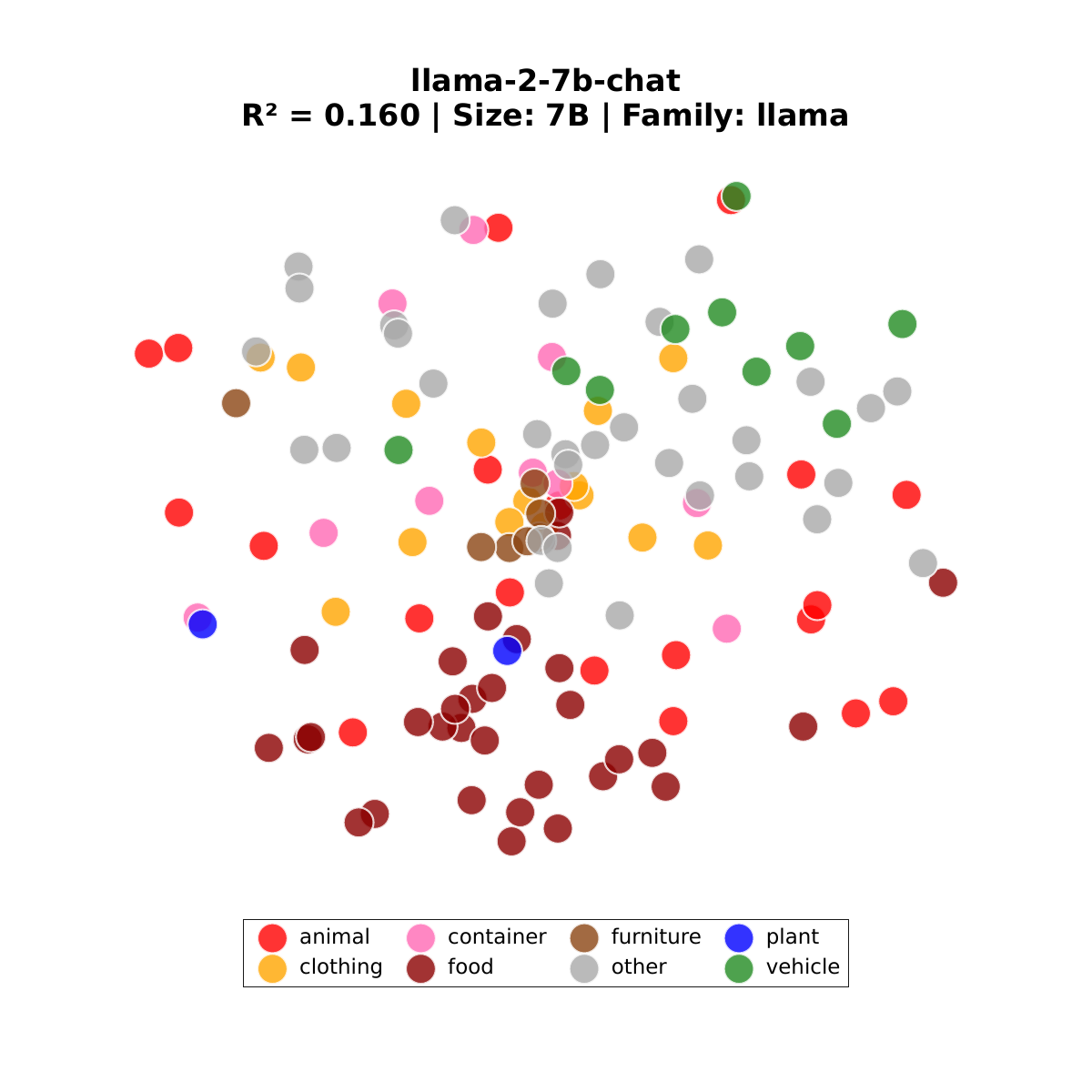} \\
        {\small (52) Falcon-7B (0.175)} & {\small (53) OLMo-2-1B-SFT (0.168)} & {\small (54) Llama2-7B-Chat (0.160)} \\[3mm]
        
        \includegraphics[width=0.31\textwidth]{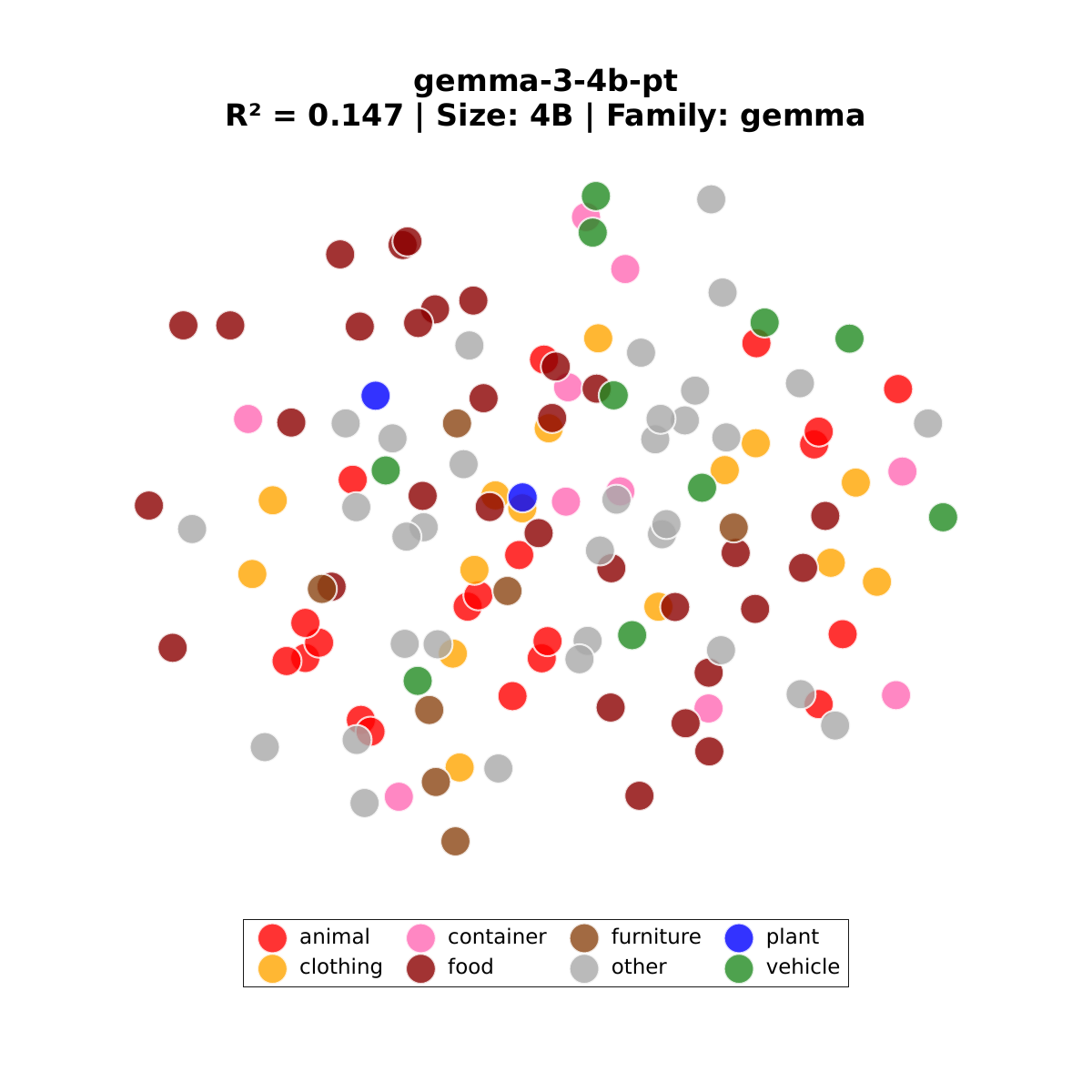} &
        \includegraphics[width=0.31\textwidth]{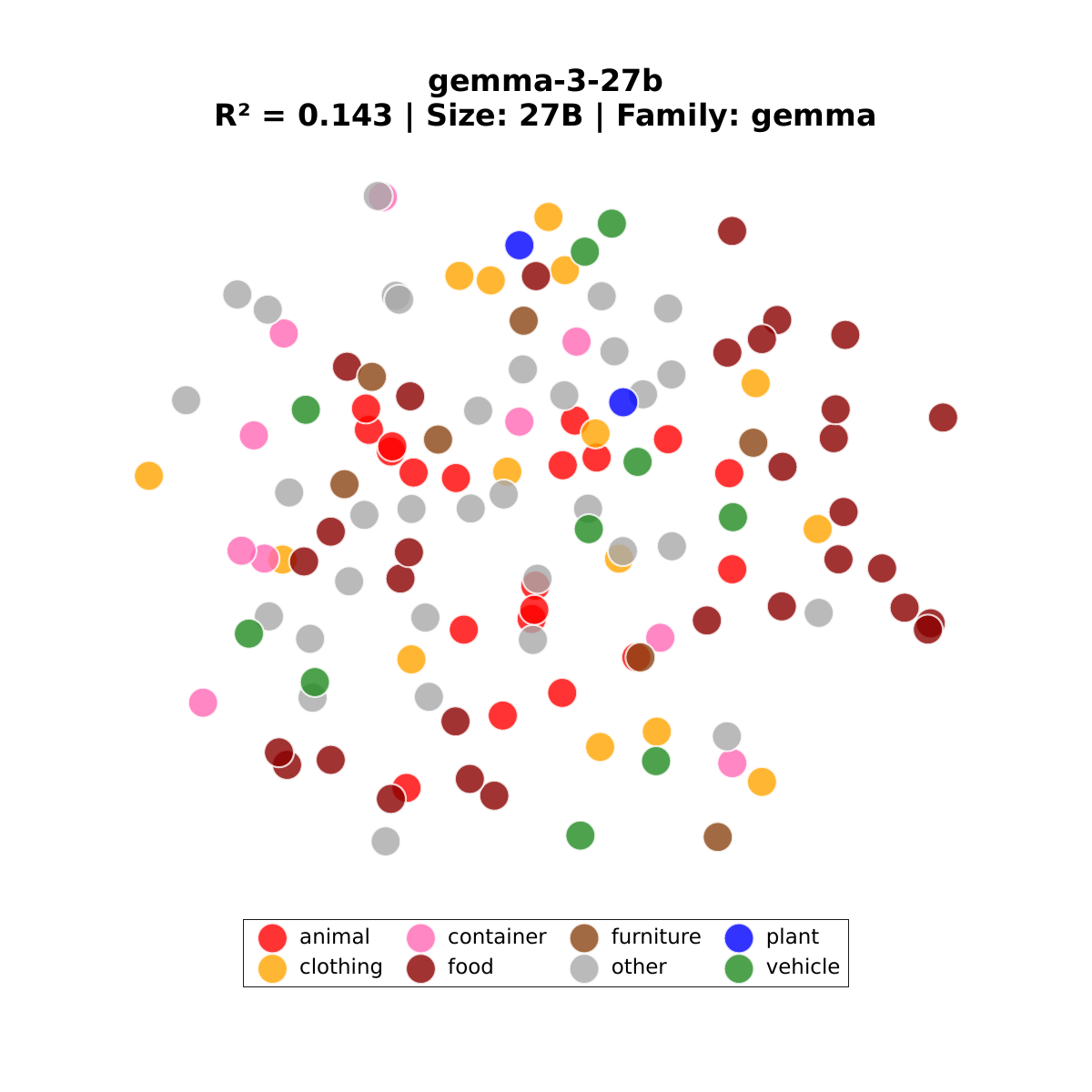} &
        \includegraphics[width=0.31\textwidth]{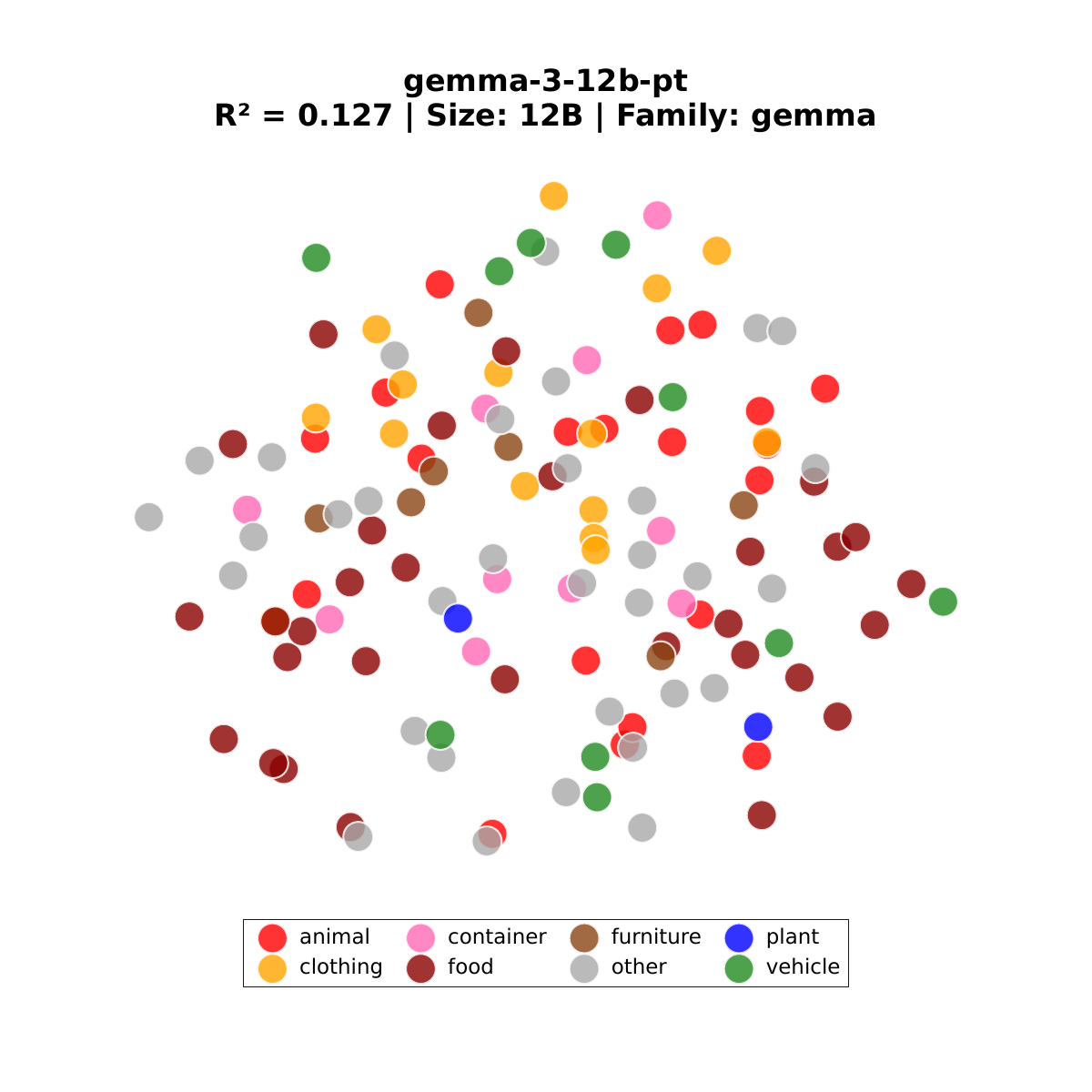} \\
        {\small (55) Gemma-3-4B-PT (0.147)} & {\small (56) Gemma-3-27B-PT (0.143)} & {\small (57) Gemma-3-12B-PT (0.127)} \\[3mm]
        
        \includegraphics[width=0.31\textwidth]{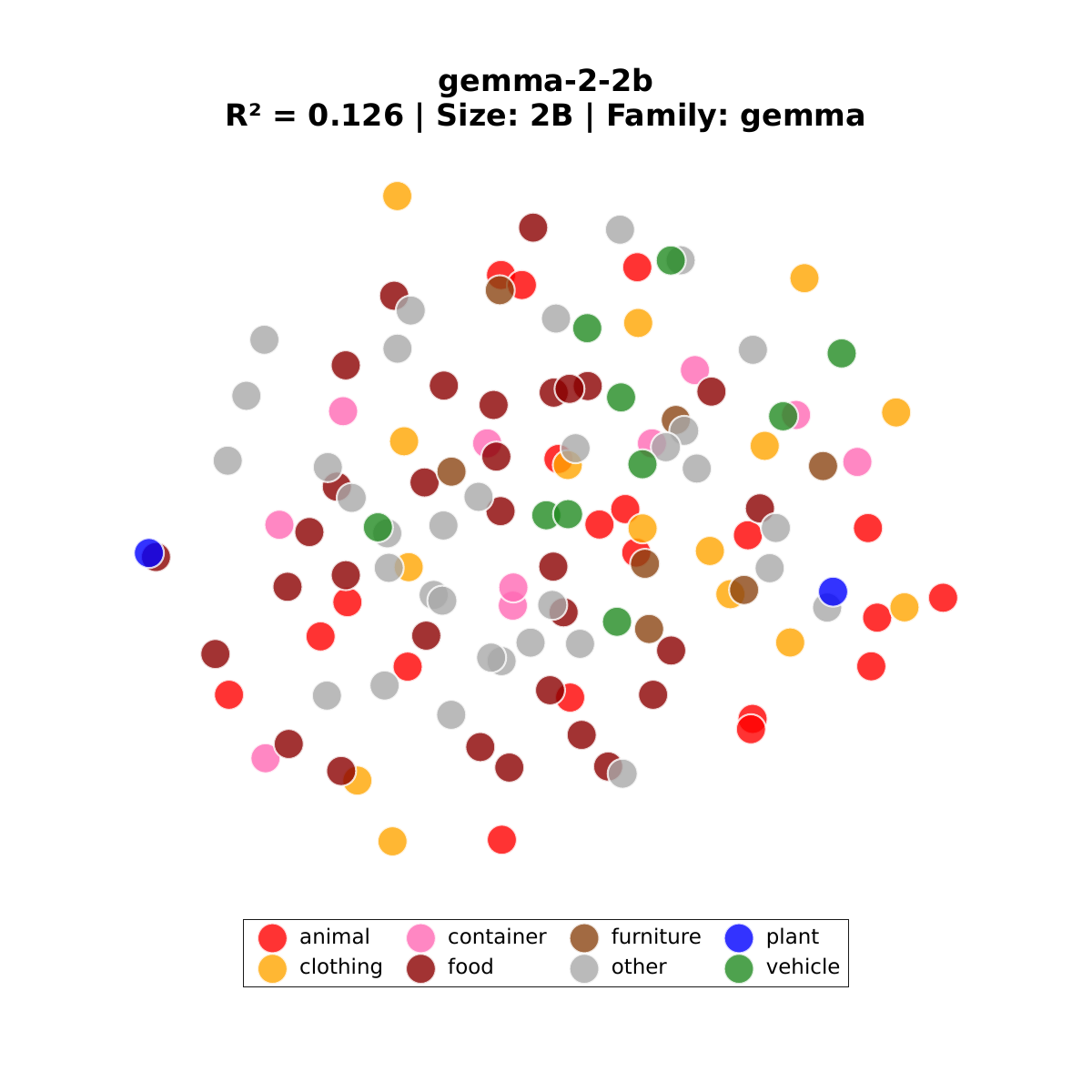} &
        \includegraphics[width=0.31\textwidth]{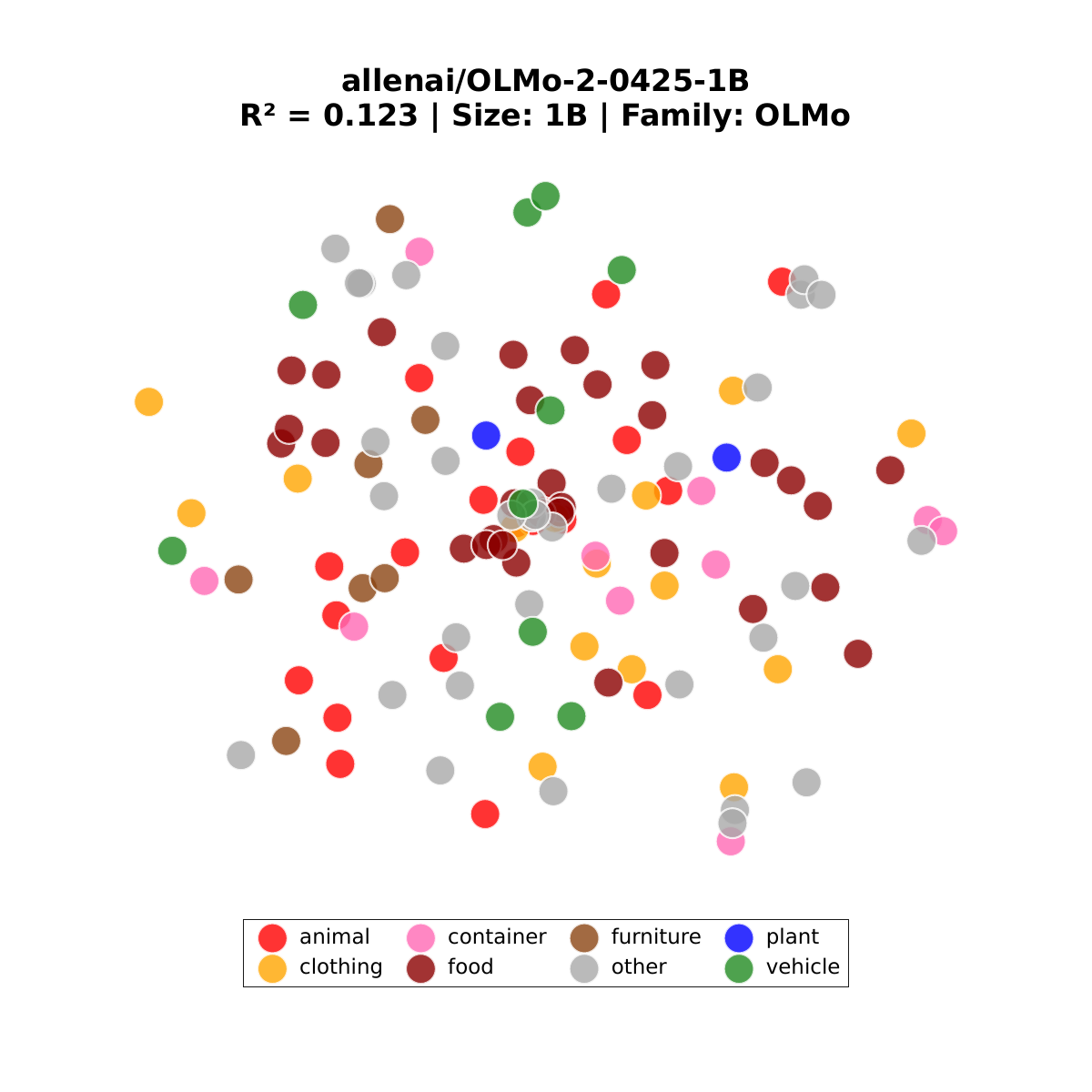} &
        \includegraphics[width=0.31\textwidth]{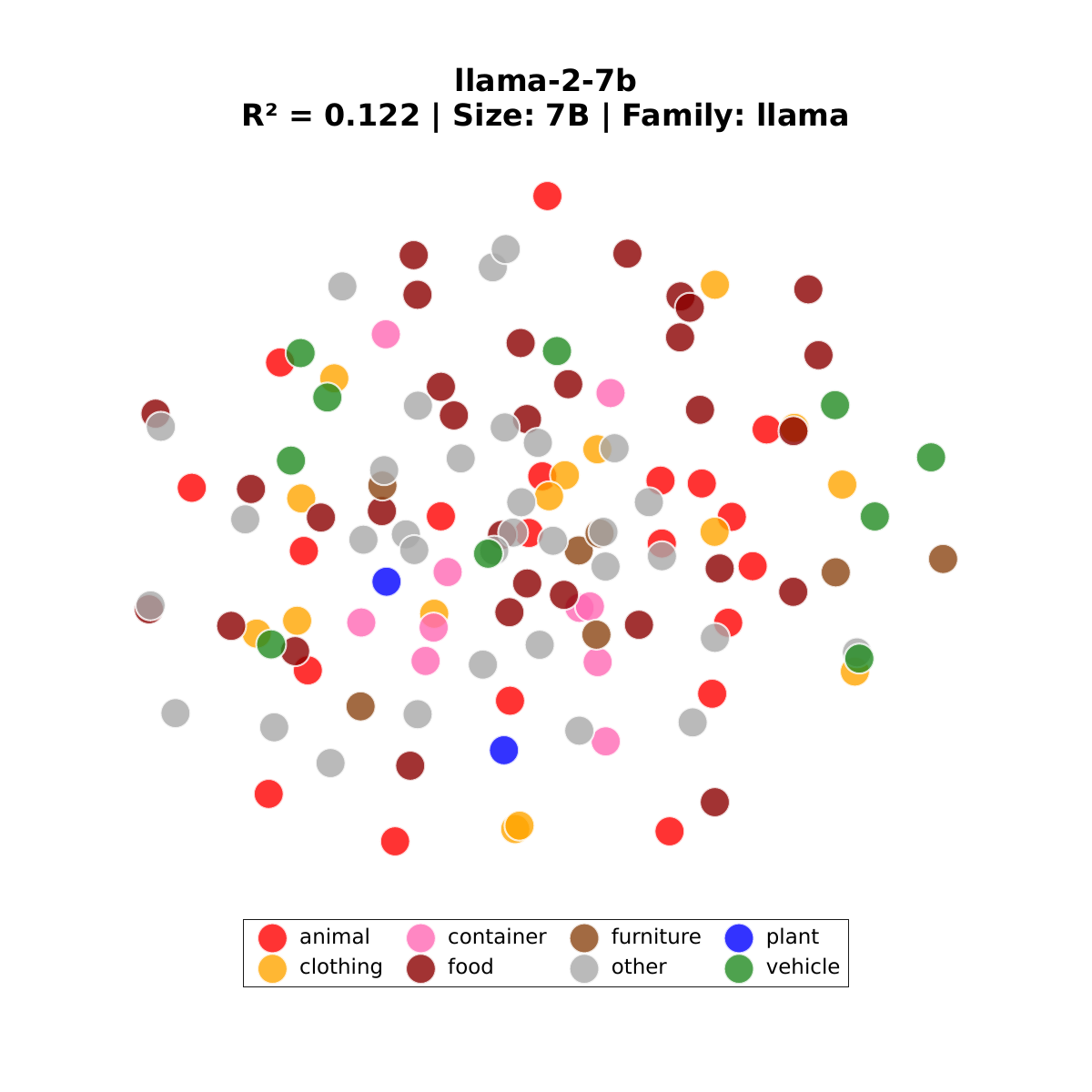} \\
        {\small (58) Gemma-2-2B (0.126)} & {\small (59) OLMo-2-1B (0.123)} & {\small (60) Llama2-7B (0.122)} \\
    \end{tabular}
\end{figure}

\begin{figure}[t!]
    \centering
    \caption{t-SNE visualizations of model embeddings (Part 6). R² scores are in parentheses.}
    \label{fig:tsne_models_part6}
    \begin{tabular}{@{}c@{\hspace{2mm}}c@{\hspace{2mm}}c@{}}
        \includegraphics[width=0.31\textwidth]{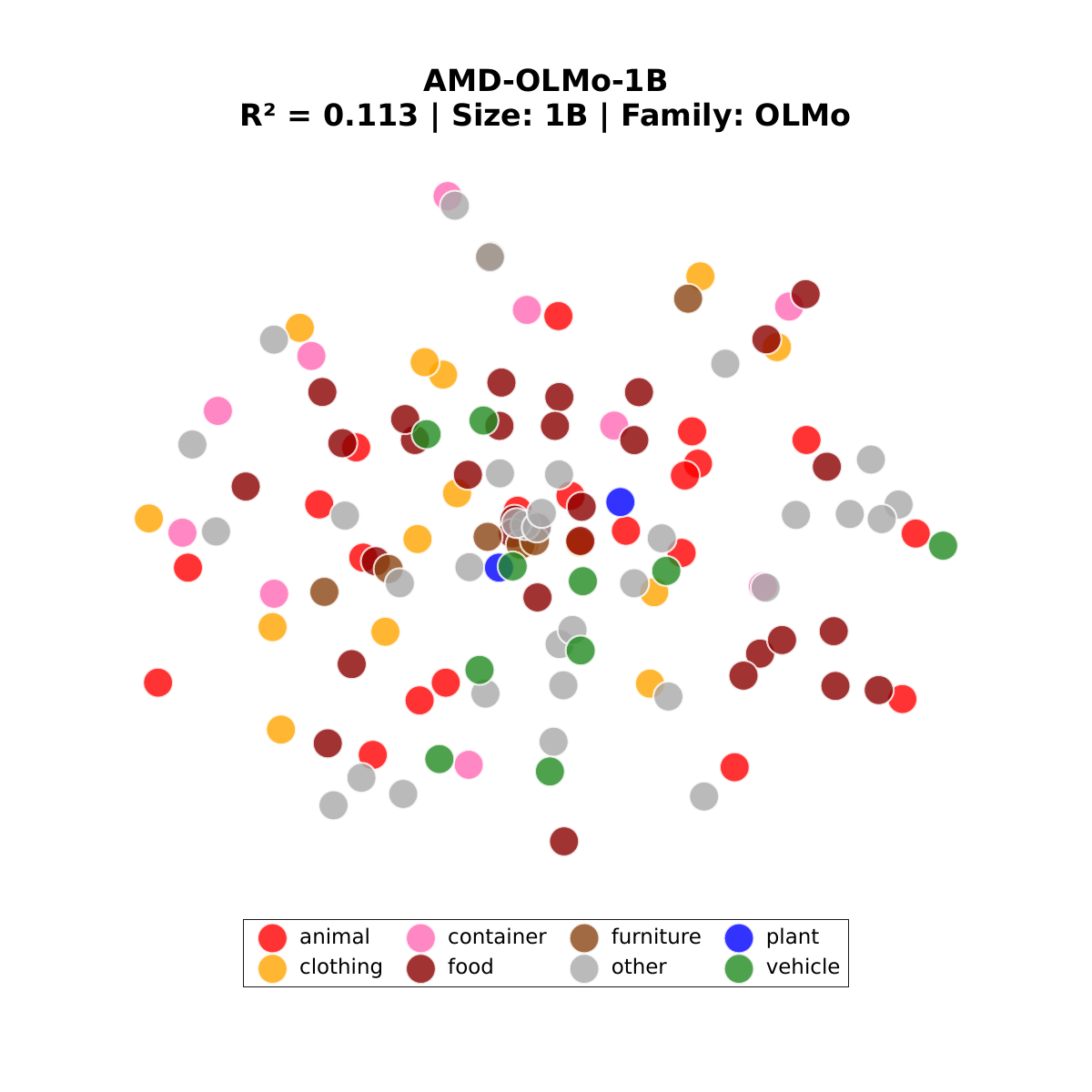} &
        \includegraphics[width=0.31\textwidth]{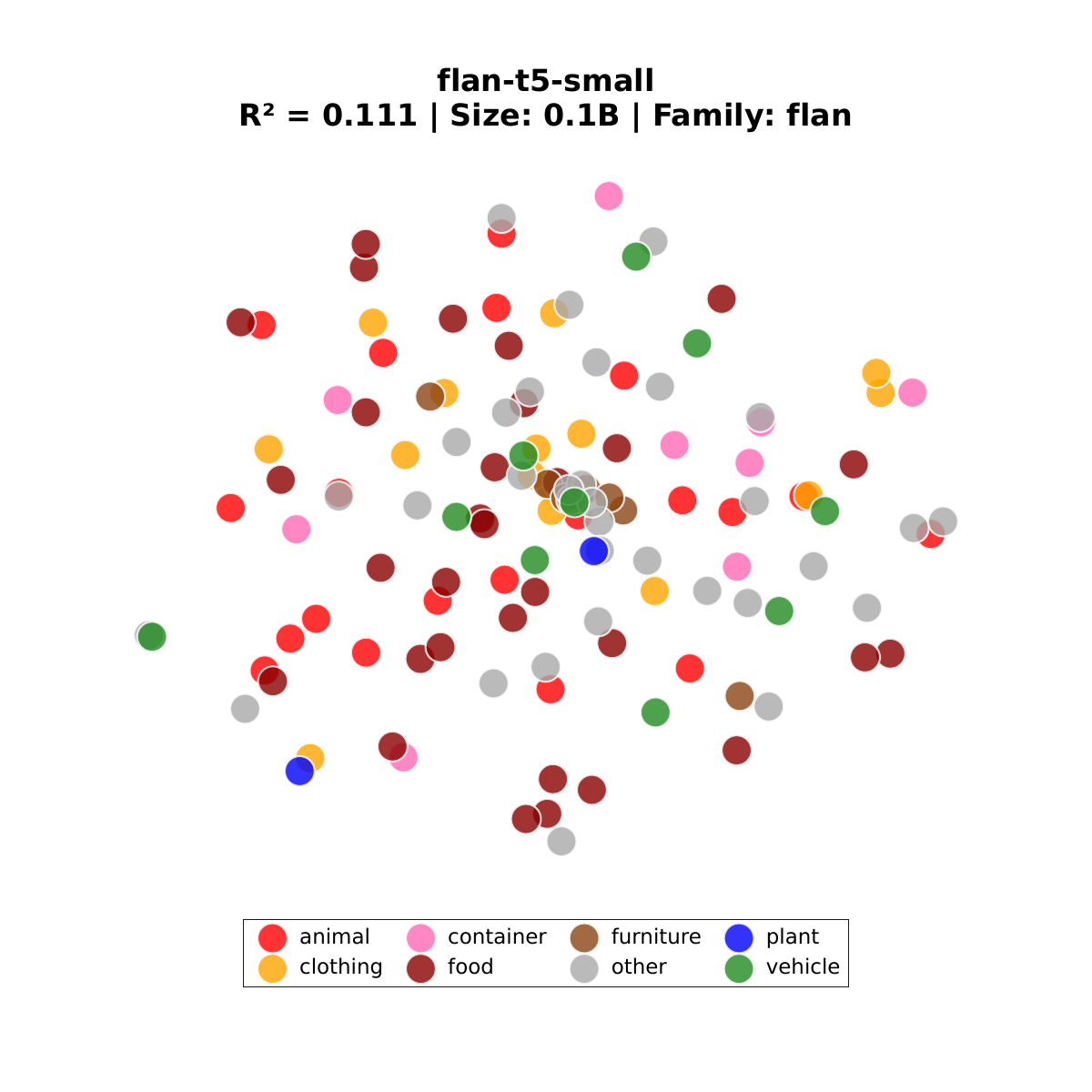} &
        \includegraphics[width=0.31\textwidth]{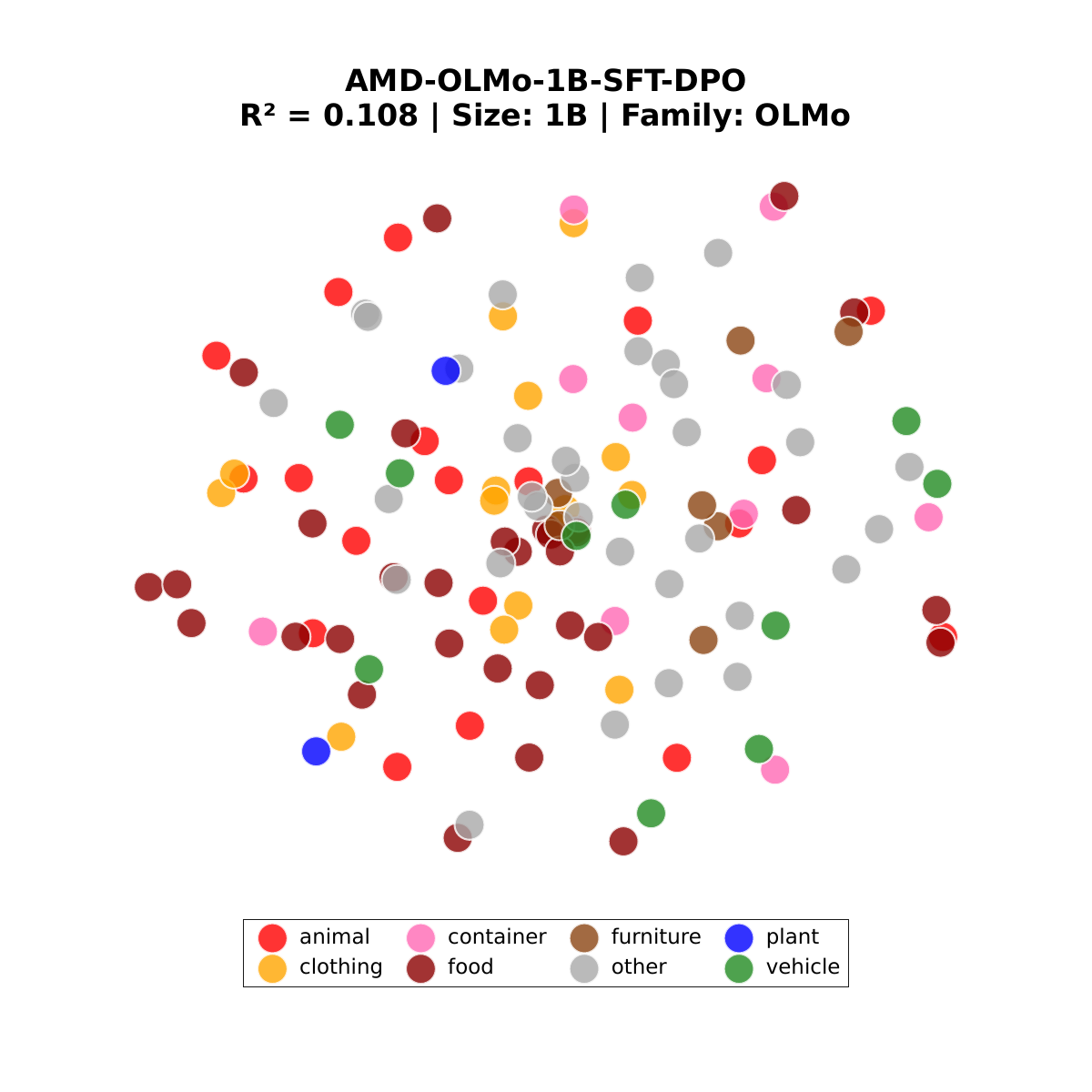} \\
        {\small (61) AMD-OLMo-1B (0.113)} & {\small (62) FLAN-T5-Small (0.111)} & {\small (63) AMD-OLMo-1B-DPO (0.108)} \\[3mm]
        
        \includegraphics[width=0.31\textwidth]{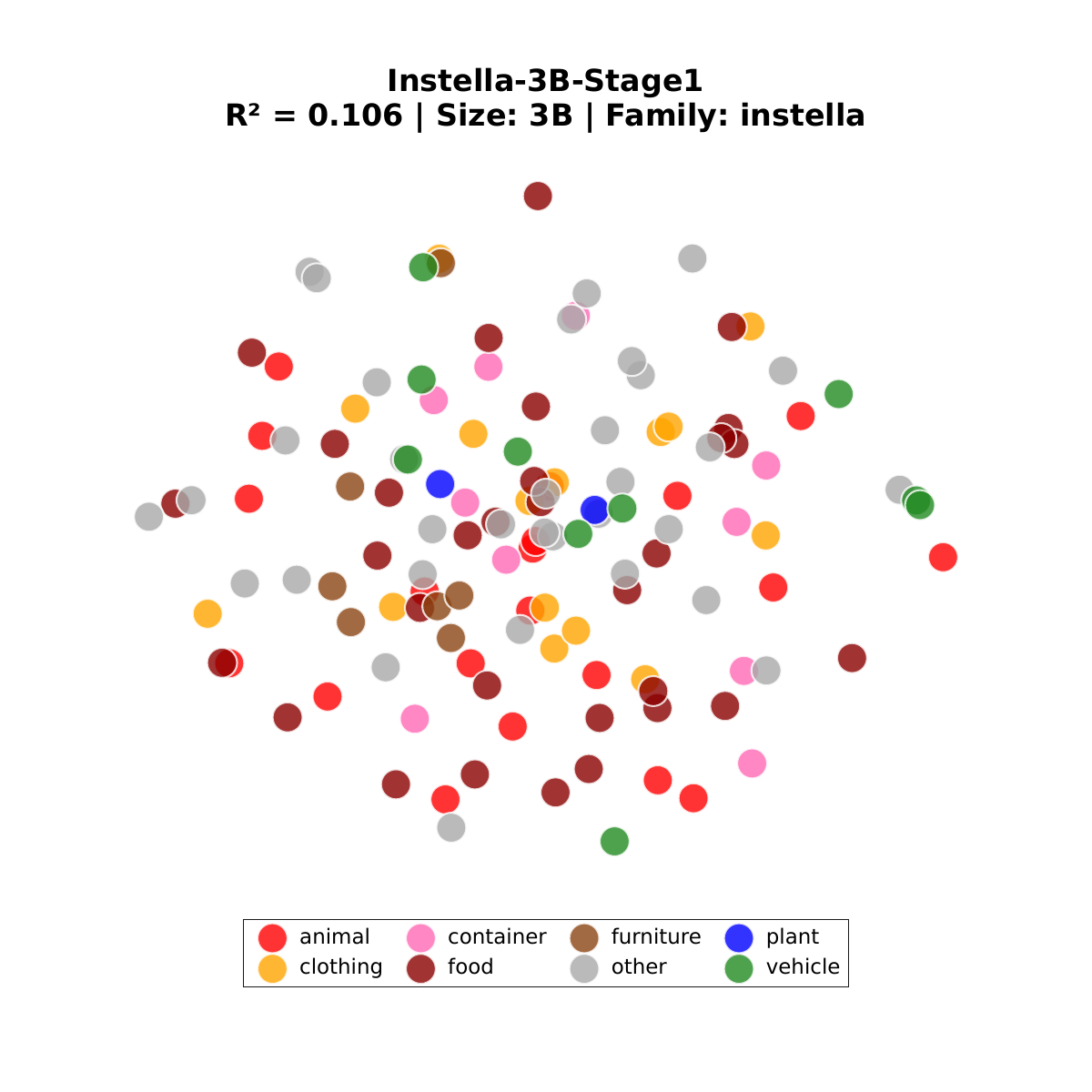} &
        \includegraphics[width=0.31\textwidth]{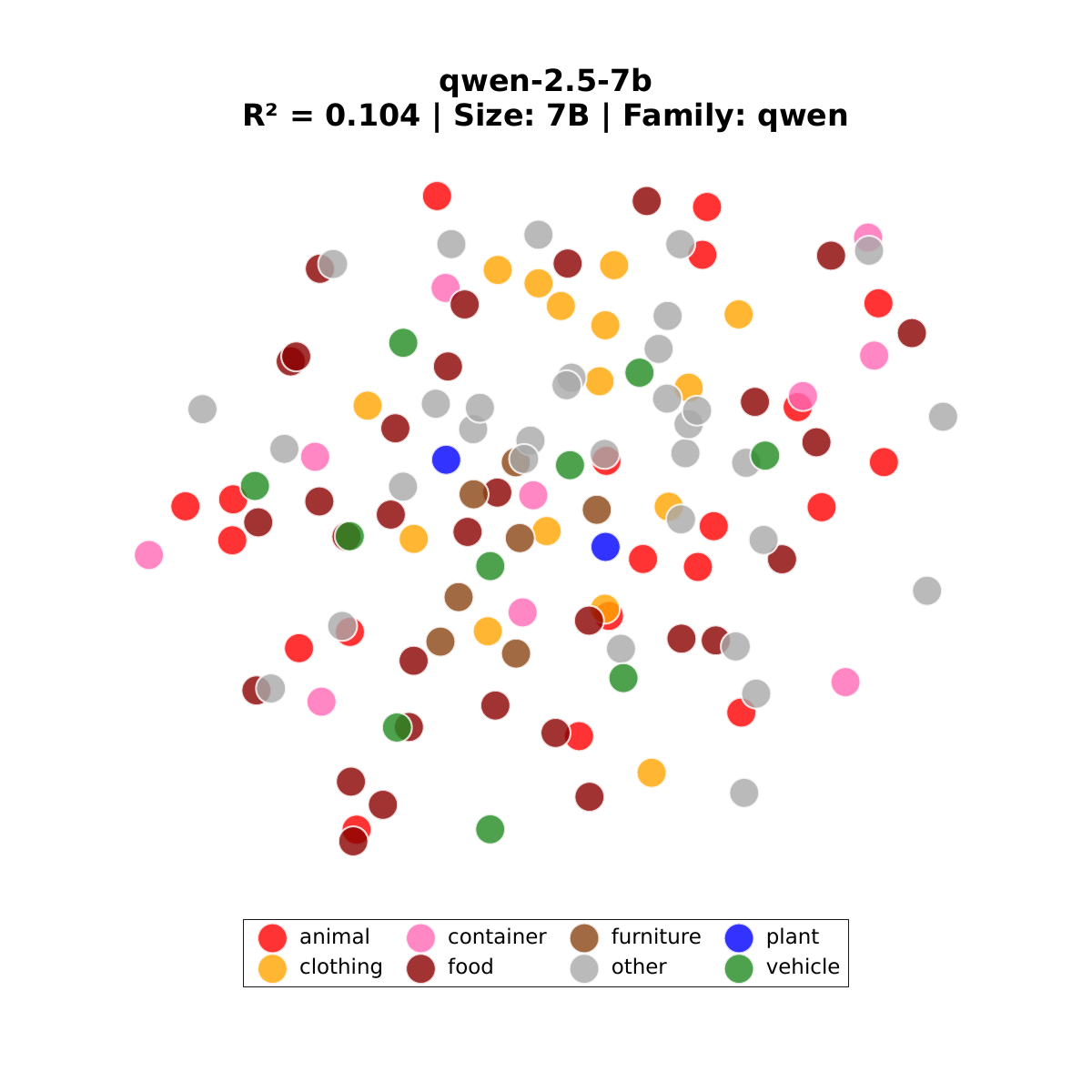} &
        \includegraphics[width=0.31\textwidth]{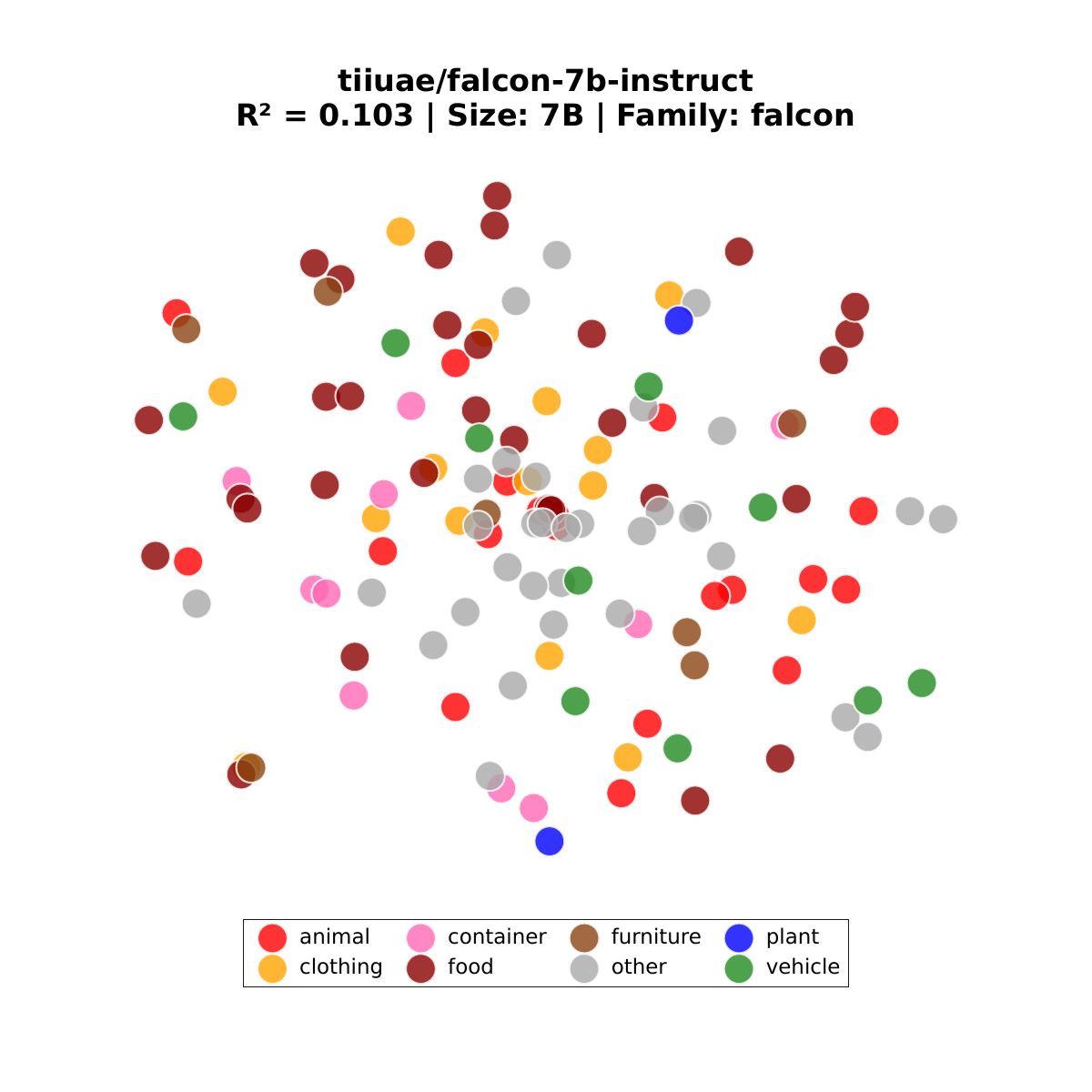} \\
        {\small (64) Instella-3B-Stage1 (0.106)} & {\small (65) Qwen2.5-7B (0.104)} & {\small (66) Falcon-7B-Inst (0.103)} \\[3mm]
        
        \includegraphics[width=0.31\textwidth]{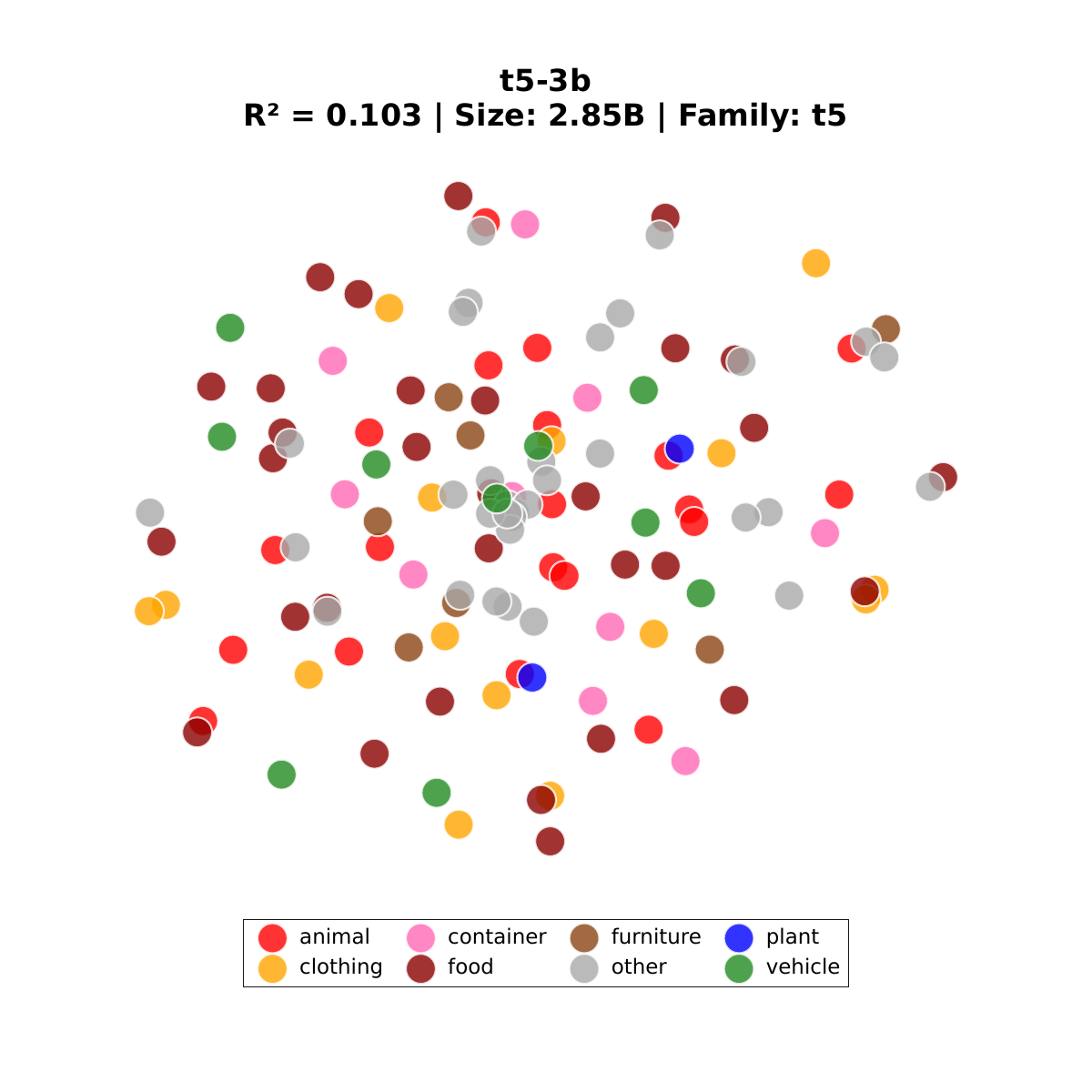} &
        \includegraphics[width=0.31\textwidth]{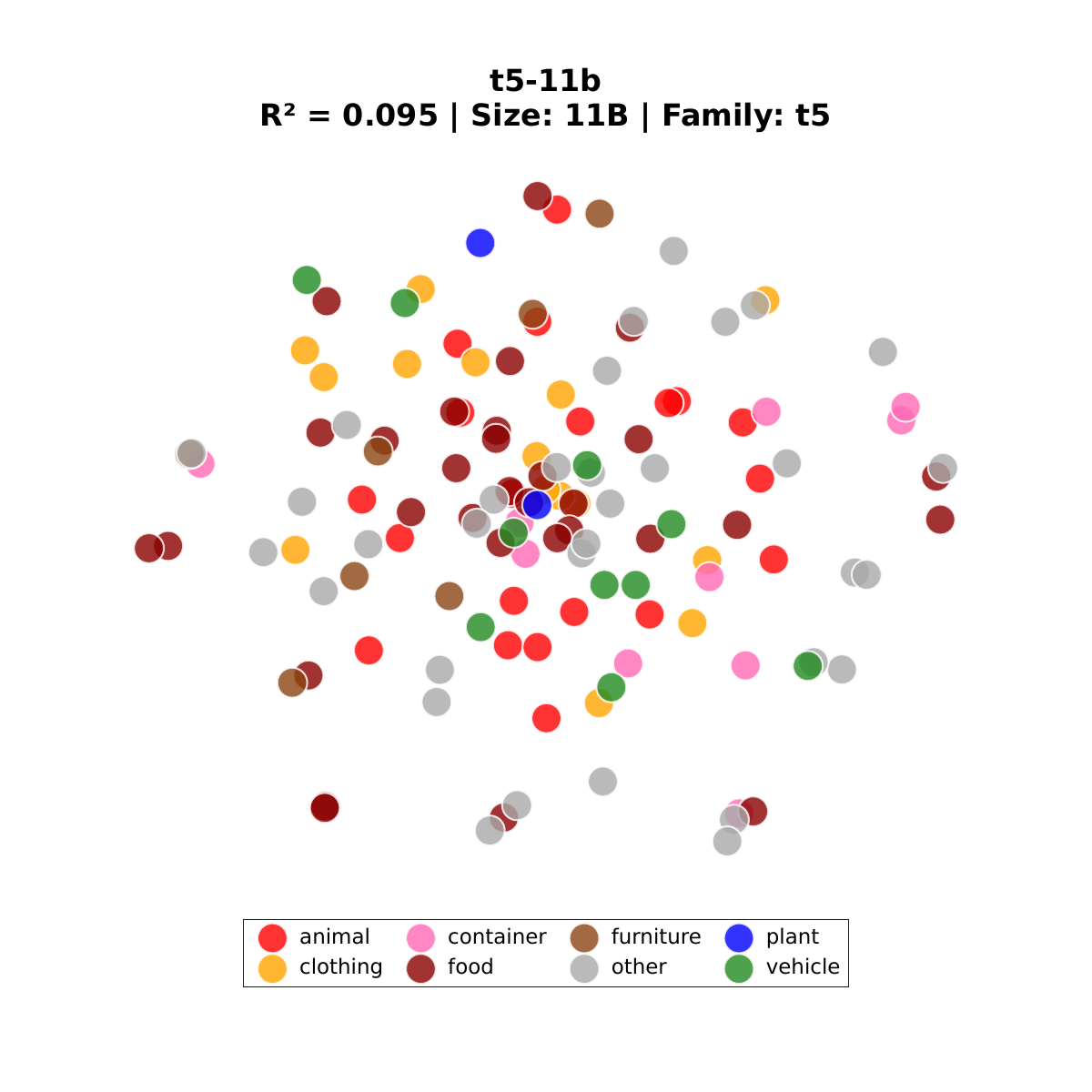} &
        \includegraphics[width=0.31\textwidth]{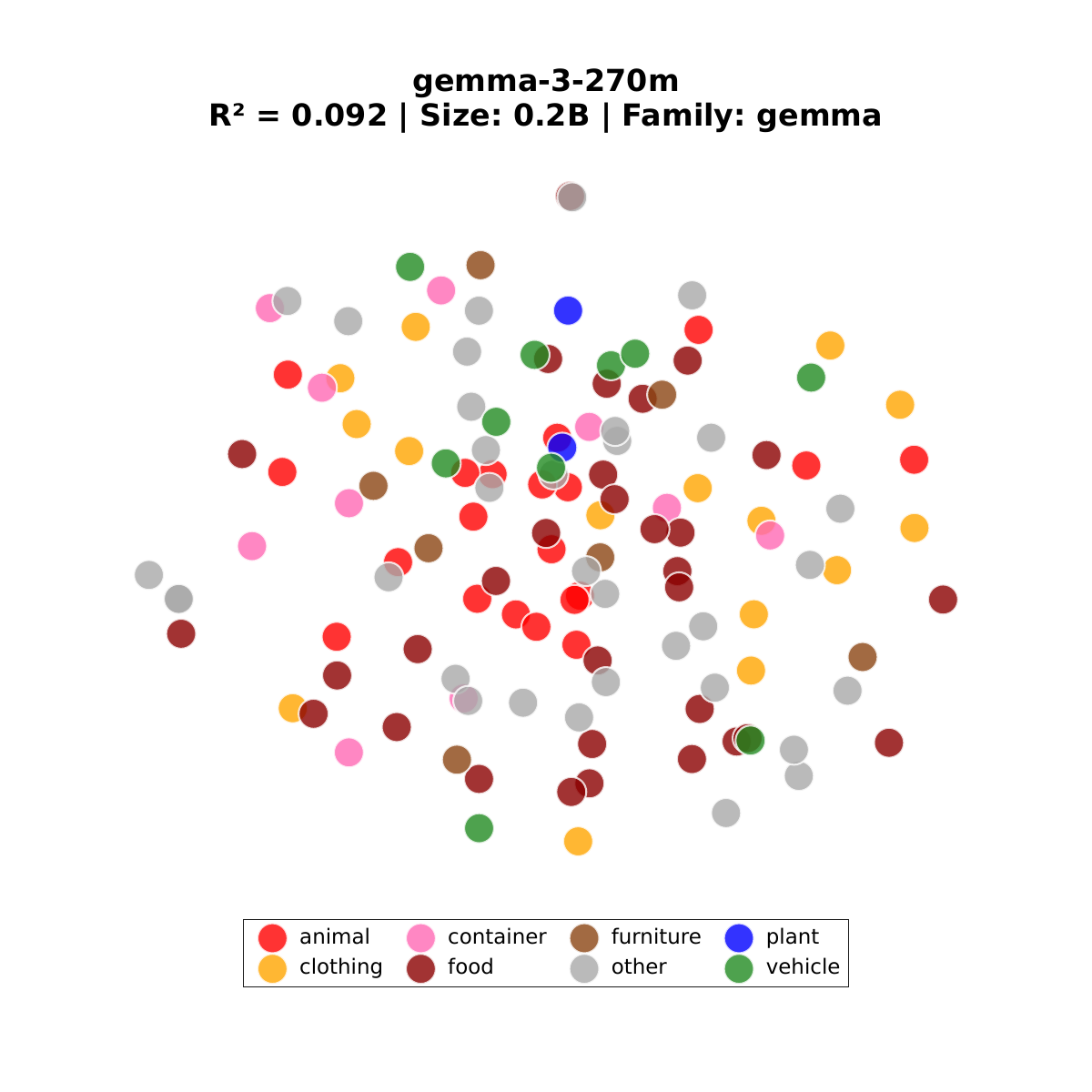} \\
        {\small (67) T5-3B (0.103)} & {\small (68) T5-11B (0.095)} & {\small (69) Gemma-3-270M (0.092)} \\[3mm]
        
        \includegraphics[width=0.31\textwidth]{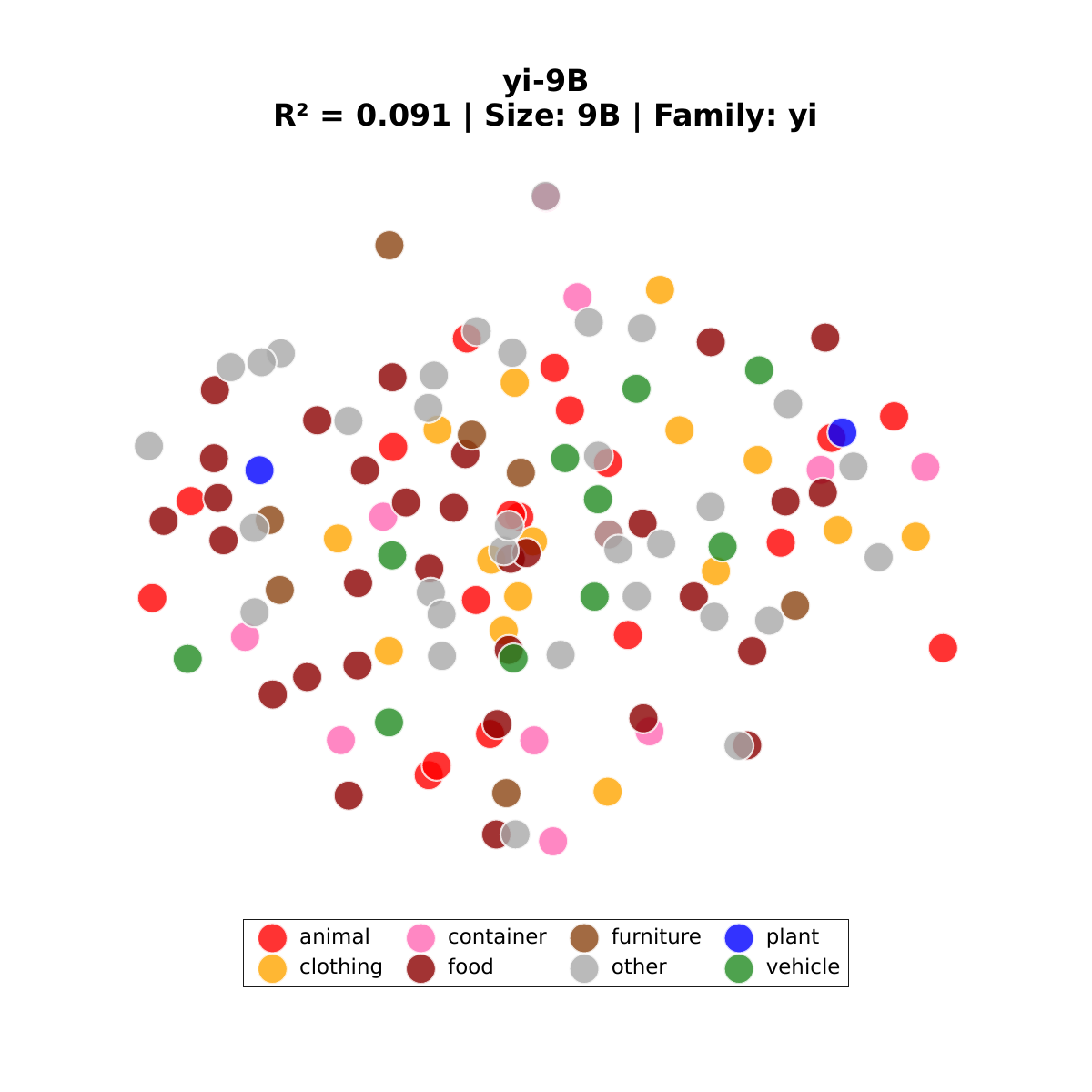} &
        \includegraphics[width=0.31\textwidth]{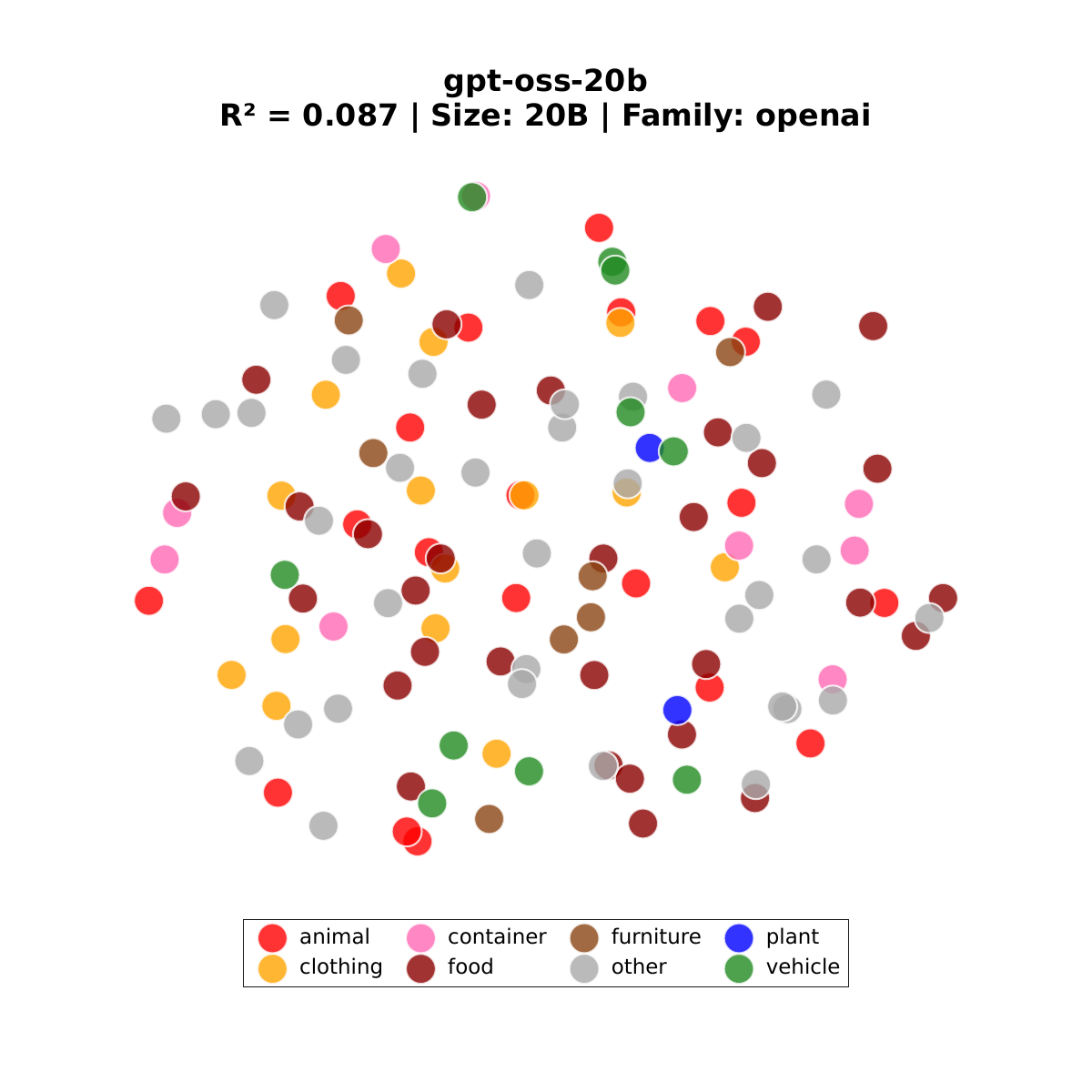} &
        \includegraphics[width=0.31\textwidth]{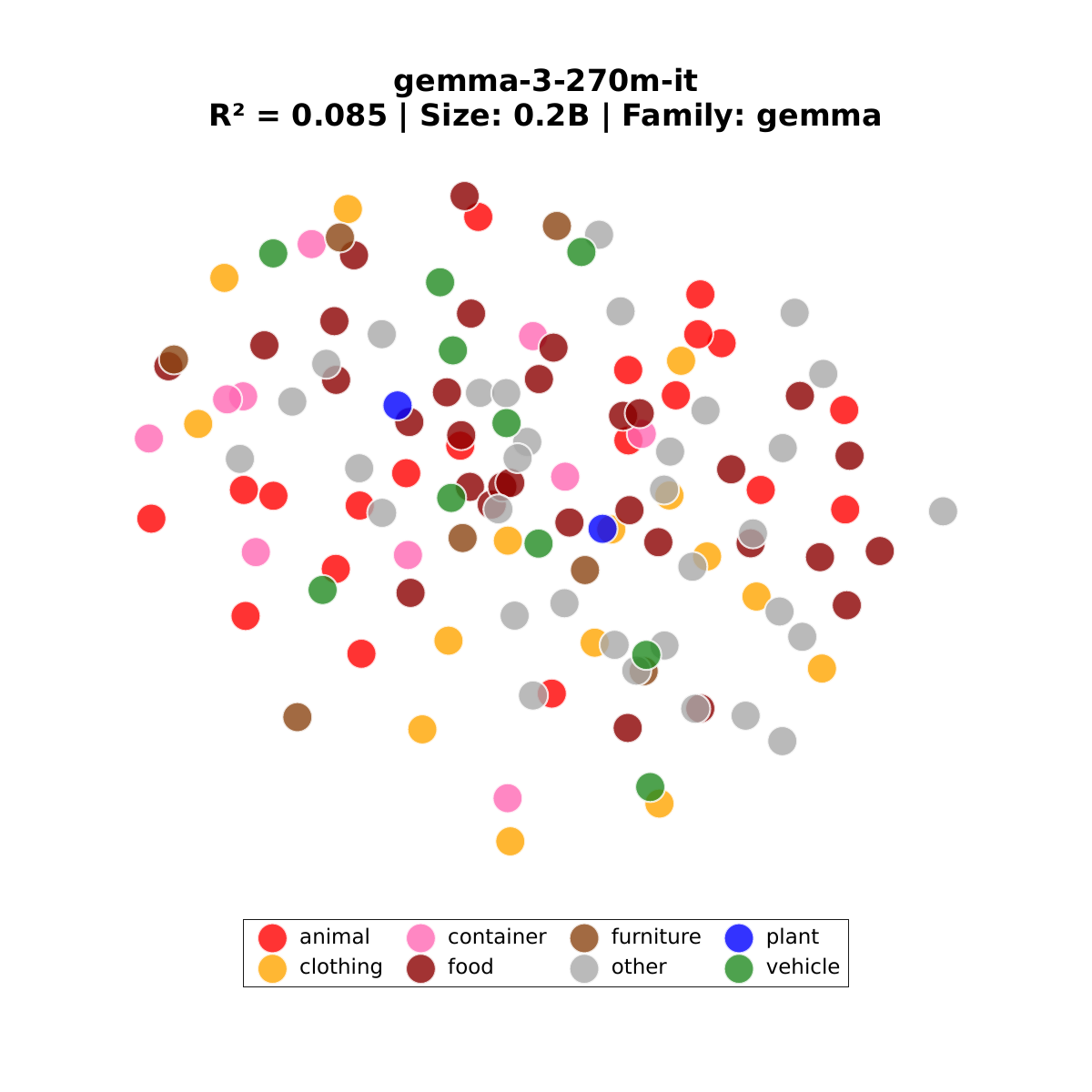} \\
        {\small (70) Yi-9B (0.091)} & {\small (71) GPT-OSS-20B (0.087)} & {\small (72) Gemma-3-270M-IT (0.085)} \\
    \end{tabular}
\end{figure}
\newpage
\begin{figure}[t!]
    \centering
    \caption{t-SNE visualizations of model embeddings (Part 7). R² scores are in parentheses.}
    \label{fig:tsne_models_part7}
    \begin{tabular}{@{}c@{\hspace{2mm}}c@{\hspace{2mm}}c@{}}
        \includegraphics[width=0.31\textwidth]{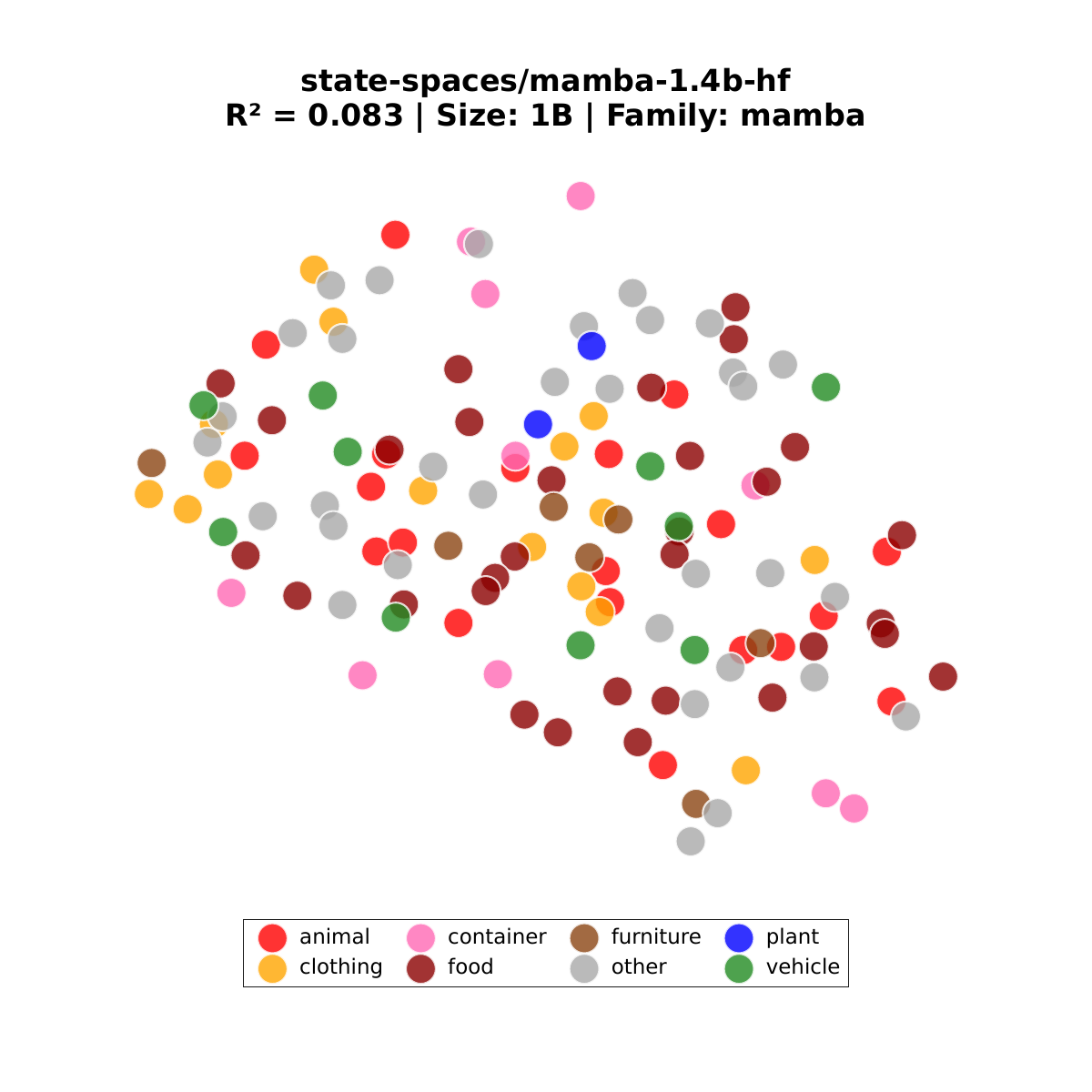} &
        \includegraphics[width=0.31\textwidth]{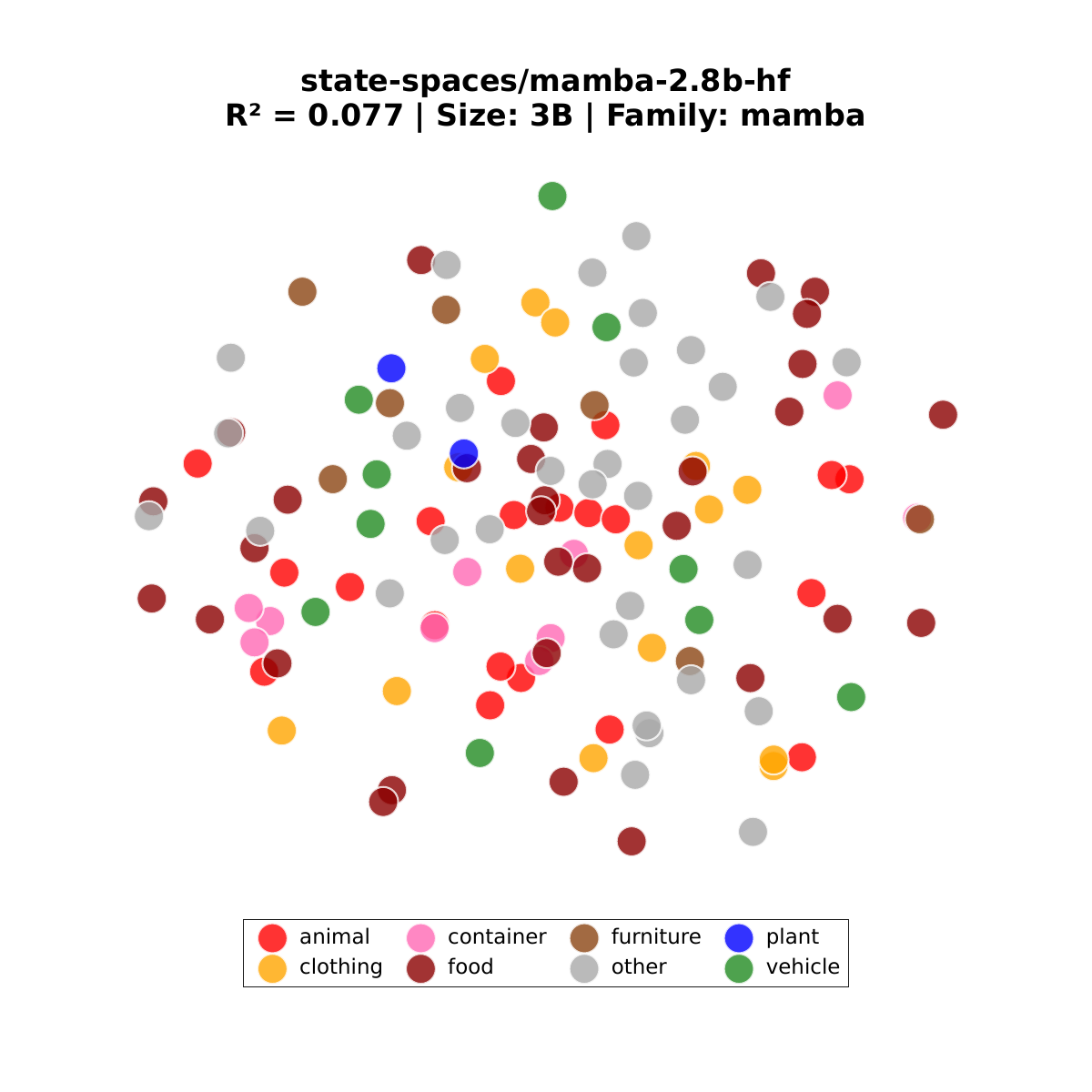} &
        \includegraphics[width=0.31\textwidth]{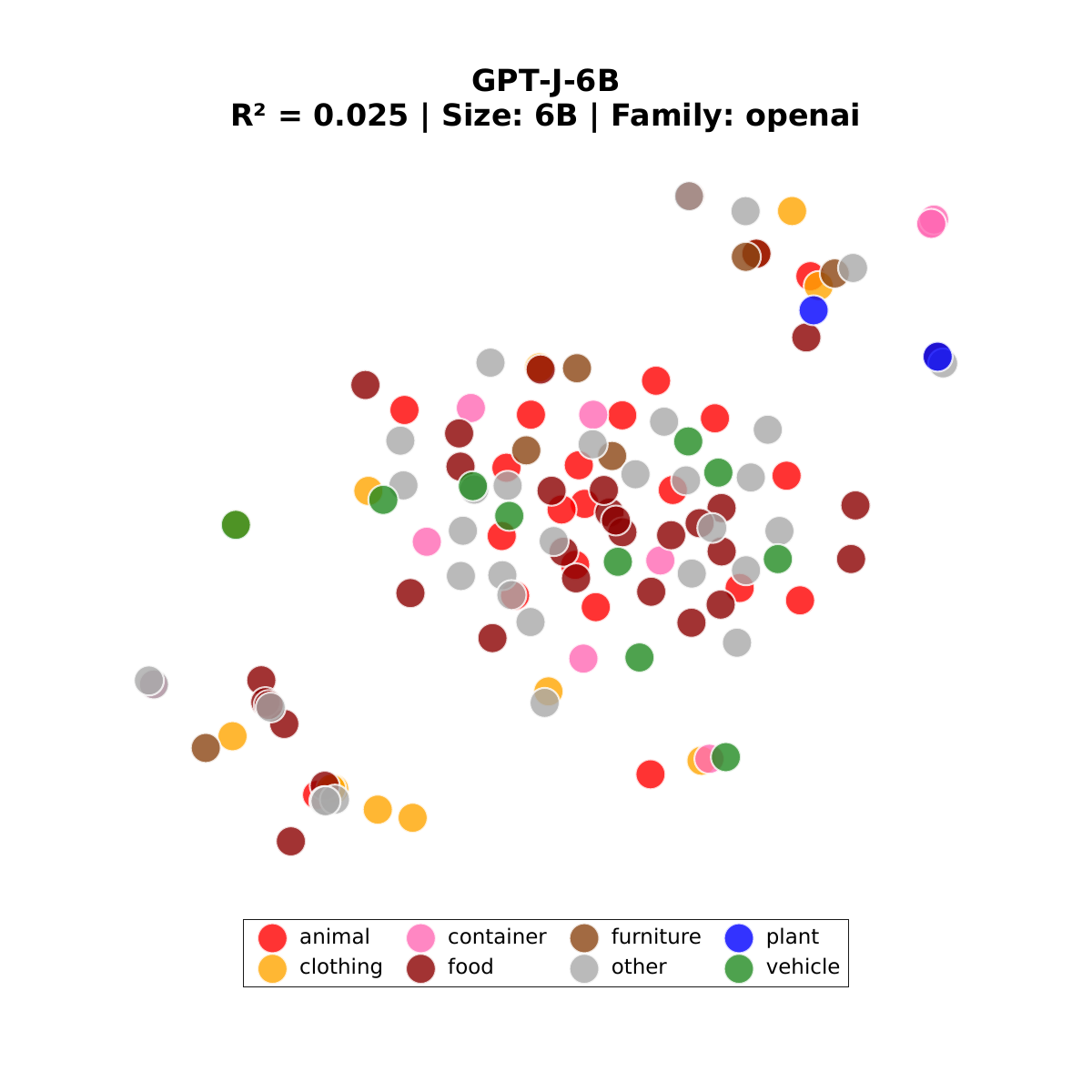} \\
        {\small (73) Mamba-1.4B (0.083)} & {\small (74) Mamba-2.8B (0.077)} & {\small (75) GPT-J-6B (0.025)} \\
    \end{tabular}
\end{figure}



\end{document}